\documentclass[12pt,oneside]{book}

\newcommand{\mytitle}{Simultaneous Dimensionality Reduction for Extracting Useful Representations of Large Empirical Multimodal Datasets}
\newcommand{\myname}{Eslam Abdelaleem}
\newcommand{\mypreviousdegrees}{
	B.Sc., University of Science and Technology at Zewail City, Giza, Egypt, 2019
 \\ 	M.Sc., Emory University, Atlanta, Georgia, 2023
}
\newcommand{\mydegree}{Doctor of Philosophy} % your degree is either Doctor of Philosophy or Master of Arts/Science/Development Practice or another area
\newcommand{\mydoctype}{dissertation} % "dissertation" for doctoral degree and "thesis" for master's degree
\newcommand{\mydepartment}{Physics Department}
\newcommand{\advisor}{Ilya Nemenman, Ph.D.}
\newcommand{\committeeOne}{Christopher J. Rozell, Ph.D.}
\newcommand{\committeeTwo}
{Daniel M. Sussman, Ph.D.}
\newcommand{\committeeThree}{Gordon J. Berman, Ph.D.}
\newcommand{\committeeFour}{Justin C. Burton, Ph.D.}
\newcommand{\mydean}{Kimberly Jacob Arriola, Ph.D.}
\newcommand{\myyear}{2024}

\usepackage{formatting/emory}
\usepackage{amsmath}
\usepackage{amsfonts}
\usepackage{amssymb}
\usepackage{amsthm}
\usepackage{mathtools} % for multlined environment
\usepackage{amscd}
\usepackage[hidelinks]{hyperref}
\usepackage{siunitx}
\usepackage{graphicx}
\usepackage{booktabs}
\usepackage[ruled,linesnumbered]{algorithm2e}
\usepackage[numbers]{natbib}
\usepackage{wrapfig}
\usepackage{tikz}
\usepackage{multirow}
\usepackage{microtype}      % microtypography
\usepackage{xcolor}         % colors
\usetikzlibrary{positioning, quotes}
\usepackage{enumerate}
\usepackage{adjustbox}
\usepackage{array} 
\usetikzlibrary{shapes, arrows.meta}
\usepackage{float}
\usepackage{bm}
\usepackage{subcaption}

\usepackage{lipsum} % for demo purposes only

\theoremstyle{definition}

\usepackage{arabtex}
\usepackage{utf8}
\usepackage{setspace}
\usepackage{tikz}
\usepackage{neuralnetwork}
\usepackage{caption}

\begin{document}
\setcode{utf8}
% Set the depth for numbering subsections, subsubsections, and subsubsubsections
\setcounter{secnumdepth}{4}

% Define a command for subsubsubsections
\newcommand{\subsubsubsection}[1]{\paragraph{#1}\mbox{}\\}
\pagestyle{empty} % no page numbers on special pages
% Distribution Agreement

\vspace*{\fill}

\noindent\textbf{Distribution Agreement}

\bigskip 
\bigskip

\begin{singlespace}
\noindent In presenting this thesis or dissertation as a partial fulfillment of the requirements for an advanced degree from Emory University, I hereby grant to Emory University and its agents the non-exclusive license to archive, make accessible, and display my thesis or dissertation in whole or in part in all forms of media, now or hereafter known, including display on the world wide web. I understand that I may select some access restrictions as part of the online submission of this thesis or dissertation. I retain all ownership rights to the copyright of the thesis or dissertation. I also retain the right to use in future works (such as articles or books) all or part of this thesis or dissertation.
\end{singlespace}

\bigskip
\bigskip

\noindent Signature:

\smallskip

\noindent 
\begin{tabular}{@{}llc}
\underline{\hspace{3in}} & \hspace{0.75in} & \underline{\hspace{1in}} \\
\myname & & Date
\end{tabular}

\vfill
%\newpage

% Approval Sheet

\vspace*{\fill}

\begin{singlespace}
    \begin{center}
        \mytitle

        \bigskip
        
        By
        
        \bigskip
        
        \myname\\
        \mydegree
        
        \bigskip
        
        \mydepartment
        
        \vspace{0.75in}
        
        \underline{\hspace{4in}}\\
        \advisor\\
        Advisor
        
        \vspace{0.2in}
        
        \underline{\hspace{4in}}\\
        \committeeOne\\
        Committee Member
        
        \vspace{0.2in}
        
        \underline{\hspace{4in}}\\
        \committeeTwo\\
        Committee Member
        
        \vspace{0.2in}
        \underline{\hspace{4in}}\\
        \committeeThree\\
        Committee Member
        
        \vspace{0.2in}

        \underline{\hspace{4in}}\\
        \committeeFour\\
        Committee Member
        
        \vspace{0.2in}
        Accepted:
        
        \vspace{0.5in}
        
        \underline{\hspace{4in}}\\
        \mydean\\
        Dean of the James T. Laney School of Graduate Studies 
        
        \vspace{0.5in}
        
        \underline{\hspace{1.5in}}\\
        Date
    \end{center}
\end{singlespace}

\vfill
\newpage

% Abstract Cover Page

\vspace*{\fill}

\begin{singlespace}
    \begin{center}
        \mytitle
        
        \vspace{0.7in}
        
        By
        
        \vspace{0.7in}
        
        \myname\\
        \mypreviousdegrees
        
        \vspace{0.75in}
        
        Advisor: \advisor
        
        \vspace{2.5in}
        
        An abstract of\\
        A \mydoctype~submitted to the Faculty of the\\
        James T. Laney School of Graduate Studies of Emory University\\
        in partial fulfillment of the requirements for the degree of \\
        \mydegree\\
        in \mydepartment\\
        \myyear
    \end{center}
\end{singlespace}

\vfill
\newpage

% Abstract
% Note: the Abstract may not exceed one page, formatted according to the regular page formatting instructions (margins, spacing, font).  The text itself cannot exceed 350 words (not counting the title etc.)  The Abstract may be single-spaced.

\begin{singlespace}
    \begin{center}
        Abstract 
        
        \bigskip
        
        \mytitle \\
        By \myname
    
    \end{center}

    \bigskip
    
    % abstract text goes here
    \noindent The quest for simplification in physics drives the exploration of concise mathematical representations for complex systems. This Dissertation focuses on the concept of dimensionality reduction as a means to obtain low-dimensional descriptions from high-dimensional data, facilitating comprehension and analysis. We address the challenges posed by real-world data that defy conventional assumptions, such as the complex interactions within neural systems or high-dimensional dynamical systems. Leveraging insights from both theoretical physics and machine learning, this work unifies diverse dimensionality reduction methods under a comprehensive framework, the Deep Variational Multivariate Information Bottleneck. This framework enables the design of tailored reduction algorithms based on specific research questions and data characteristics. We explore and assert the efficacy of simultaneous dimensionality reduction approaches over their independent reduction counterparts, demonstrating their superiority in capturing covariation between multiple modalities, while requiring less data. We also introduced novel techniques, such as the Deep Variational Symmetric Information Bottleneck, for general nonlinear simultaneous dimensionality reduction. We show that the same principle of simultaneous reduction is the key to efficient and precise estimation of mutual information, a fundamental measure of statistical dependencies. We show that our new method is able to discover the coordinates of high dimensional observations of dynamical systems. Through analytical investigations and empirical validations, we shed light on the intricacies of dimensionality reduction methods, paving the way for enhanced data analysis across various domains. We underscore the potential of these methodologies to extract meaningful insights from complex datasets, driving advancements in fundamental research and applied sciences. As these methods evolve and find broader applications, they promise to deepen our understanding of complex systems and inform more effective experimental design and data analysis strategies.

\end{singlespace}
\newpage
 % place abstract here
% Cover Page

\vspace*{\fill}

\begin{singlespace}
    \begin{center}
        \mytitle
        
        \vspace{0.7in}
        
        By
        
        \vspace{0.7in}
        
        \myname\\
        \mypreviousdegrees
        
        \vspace{0.75in}
        
        Advisor: \advisor
        
        \vspace{2.5in}
        
        A \mydoctype~submitted to the Faculty of the\\
        James T. Laney School of Graduate Studies of Emory University\\
        in partial fulfillment of the requirements for the degree of \\
        \mydegree\\
        in \mydepartment\\
        \myyear
    \end{center}
\end{singlespace}

\vfill
\newpage

\begin{center}
\begin{RLtext}
بِسْمِ اللَّهِ الرَّحْمَنِ الرَّحِيمِ
\vspace{10pt}
وَفَوْقَ كُلِّ ذِي عِلْمٍ عَلِيمٌ (٦٧)
\vspace{10pt}
صَدَقَ اللهُ العظيمُ
\end{RLtext}
\end{center}

\begin{flushright}
\begin{RLtext}
    سورة يُوسُف
\end{RLtext}
\end{flushright}

\vspace{20pt}

\begin{center}
    \textit{And above every scholar, is the All-Knower (76)}
\end{center}
\begin{flushright}
    Yusuf
\end{flushright}

\newpage
\begin{center}
Acknowledgments 
\end{center}

\begin{spacing}{1.5}

Five years is a long time, and it's not only the five years of the PhD, but also the time that came before them and prepared for them. This might seem like a long list of people, but I'm truly grateful and thankful for all of them, and I want to acknowledge their efforts that led to this work, either directly or indirectly.

I'd like to thank my advisor, Dr. Ilya Nemenman. Ilya, you're one of the smartest people I've met in my life. You taught me how to be a better scientist and a better person. Thanks for the lovely journey. I'd also like to thank my committee members (in no particular order): Drs. Gordon Berman, Justin Burton, Daniel Sussman, and Chris Rozell. Each and every one of you helped me in different ways, whether through classes, research, comments, discussions, or teaching. You helped shape this dissertation, so thank you!

I'd like to thank my labmates: Dr. Ahmed Roman, you're the only elder brother figure I have had in my life. Thanks for teaching me and helping me, not just in research, but in life. Dr. K. Michael Martini, I really enjoyed our time working together. It was the most productive time I had, and I learned a lot from you. Drs. Michael Pasek and Sean Ridout, thanks for the time and discussions you contributed to this work, and thanks for your advice in general. Satya, thanks for lending me an ear when needed and all the advice you gave me in navigating infinite logistic obstacles. Ketuna, Zehui, and Arabind, thanks for all the discussions and chats about random things we had. To all the lab members, you made my time more enjoyable, and I learned a lot from every one of you.

I want to thank my cohort as well: Katie, Pablo, Tomi, Charles, Lakshmi, Tharindu, Tao, Xinyu, and Richa. You guys were my family when I came to Emory, and I miss our lounge time together.

I'd like to thank Barbara for being there from day zero, without your help with everything, I wouldn't be writing this today.

I also want to thank my brothers who resided/roomed around the 415 Richards St. residence. You took me in, and you became true brothers to me. Thanks to each and every one of you, and the food was/is/will be amazing!

Sailing across the ocean toward home, I'd like to express my utmost heartfelt gratitude to the late Dr. Ahmed Zewail, the founding father who gave me the chance that led to me writing these words. My whole life, you have always been a symbol and a role model, and I feel great pride when I connect myself to you and call myself Zewailian.

I'd like to thank my physics major friends from Zewail City. Being with you shaped me into the person I am today, and I'm thankful. I'd also like to thank my beloved friends from section 4. Talking to you every day for the last five years (and before) was sometimes the only thing that kept me hanging in there. Thanks to each and every one of you. I'd like to thank the whole community of Zewail City for its immense effect that extended beyond the four years of undergrad and up until now. May your soul, Dr. Zewail, rest in heaven.

I'd like to express my unconditional love and gratitude to my whole family, who have been supportive, caring, and loving for my whole life. I'd like to thank my mom and dad, who have been my number one supporters in all conditions, who loved me, nourished me, and nurtured me, and haven't stopped until today. I'd like to thank my brother Anas for being there; I was able to work here because I believed my family was in good hands because of you. My sisters, Salma and Malak, you always supported me in different ways, and I always had you at my back. Thanks again. I'd like to thank my wife, Salma, for being the reason that this work is concluded, our last year together was the most productive and the hardest in some sense as well, thanks for being with me through the better part and the bitter one. I'd also like to express my gratitude to my two late uncles, Fayez and Hamada. You both shaped me in different ways, and I'm grateful for the time I had with you, sad for the time I didn't, but grateful nonetheless. I want to thank the rest of my extended family as well, who gave me a feeling that I have roots and always have a base of love and encouragement.

To everyone here, I'm truly grateful and duly acknowledging your effect on me, whether directly on this dissertation or directly on the person who worked on it. This is yours as much as it is mine. Thank you!

\end{spacing}
 % optional, comment to remove

% TABLE OF CONTENTS/LIST OF FIGURES/LIST OF TABLES/LIST OF ALGORITHMS
\frontmatter
\tableofcontents
\listoffigures % required if figures are included in text
%\listoftables % required if tables are included in text
%\listofalgorithms % required if algorithms are included in text

% CHAPTERS
\pagestyle{myheadings}
\mainmatter
\chapter{Introduction: From Complexity to Sufficient Simplicity}\label{ch1}

We, physicists, are in a persistent search for simplification. We aim to describe complex systems using succinct equations or simplified representations. We are drawn to the notion that a simple equation, with a limited set of variables or ``\textit{features}'', can faithfully replicate the behavior of seemingly complicated physical systems. Consider, for instance, the modeling of spin arrangements in a lattice by their collective magnetization or the characterization of an Avogadro's number of gas particles using coarse-grained variables, such as pressure, volume, and temperature, as used in the ideal gas law. These simplified representations are usually more intuitive, useful (in the sense that they offer predictions, limits for the interesting cases, etc.), and analytically tractable (at least we try), albeit underpinned by a series of implicit assumptions. Assumptions may involve operating within the thermodynamic limit with an infinite number of homogeneous constituents (as in infinite spin lattices or different statistical ensembles), or navigating simpler few-body problems (where `few' typically refers to 1 or 2, general three-body problems lack closed-form solutions in Newtonian mechanics, for example). We note that the complexity of a description depends on its representation (the choice of the coordinates), and some coordinates are considered `natural' because they lead to particularly simple descriptions (for example solving spherically symmetrical problems in spherical coordinates versus Cartesian coordinates). We often presume the existence of these `natural' coordinates, either based on the physical observables of the problem or established from first principles, such as the natural association of pressure and temperature with measurable phenomena in the ideal gas law. However, the complexity of other real-world problems surpasses these idealizations. Take, for instance, a dataset obtained from recording the activity of a population of neurons (the information processing cells in the brain). When we try to model such systems, we are faced with multiple problems that challenge some of the common assumptions. For instance, the number of recorded state variables in these neurons is neither a few (so that we can model each neuron in biophysical detail), nor do we have an infinite number of uniform features that resemble the thermodynamic limit. Additionally, we do not have an infinite number of observations to statistically quantify all interactions of features among themselves. Moreover, the absence of locality adds more complexity to the situation (in the brain, neurons can be connected by long axons, so that there are no nearest neighbors interactions as in the Ising model, for example). Further, basic symmetries assumed in physical systems are not strictly preserved in the brain --- the brain is neither rotationally nor translationally invariant \cite{fumarola2022mechanisms}. In addition, dynamics in the brain, and even in each neuron, have multiple interacting temporal scales \cite{hari2015brain, monsa2020processing}. For example, the same neuron exhibits ms scale precision during emission neural action potentials \cite{kirst2009precise}, but synaptic plasticity takes minutes, hours, or even years \cite{abraham2003long, zucker2002short, Song2018chapter}). Simply put, there are no well-developed theoretical tools to produce succinct, interpretable mathematical descriptions of systems that violate these assumptions, commonly assumed in theoretical physics approaches. Most methods rely on the intuition of the researcher to propose a description of the system and then verify if it is useful in predicting the outcomes of new experiments. The cycle rarely converges to {\em quantitative} agreement between models and experiments, as we have grown to expect in more traditional areas of physics. Instead of relying solely on the scientist's intuition, an alternative approach is to leverage advances in statistics and machine learning to discover the optimal low-dimensional descriptions of experimental data for specific scientific questions. This is achieved through various methods of dimensionality reduction (DR), where one finds the optimal low-dimensional representations for a specific objective via optimization over certain classes of possible low-dimensional description space.

Such DR may be interpreted in two different languages, which may seem disparate in some contexts yet perfectly aligned in others. First is the language of theoretical physics, in which the Renormalization Group (RG) is the canonical example of dimensionality reduction \cite{wilson1983renormalization}. RG is an iterative coarse-graining procedure designed to derive mathematical descriptions of physics problems characterized by multiple length scales. Its primary objective is to extract relevant features---and the laws of their interactions---of a physical system for describing phenomena at large length scales. For this, RG integrates out degrees of freedom at short length scales recursively, while keeping track of the effect these removed degrees of freedom have on the interactions among the ones that remain. Through this process, the relevant operators, the \textit{important features}, gain prominence, while irrelevant operators start to have progressively smaller effects on the system's physical properties at large scales. Note that RG in physics is usually applied to models of complex systems, and applications directly to data are still uncommon \cite{meshulam2019coarse, kline2023multi, bradde2017pca}. The other language for DR comes from statistics, machine learning (ML), and computer science, where it is formalized as minimizing a loss function that promotes a succinct description of data directly. Here one proceeds as follows. First one postulates a measure of complexity of the description, which could be counting variables, measuring their variance, entropy, or a variety of other approaches. Second, one defines a measure of quality of a description of data, which is typically some measure of the ability to reconstruct or preserve the desired data features. Ultimately, most DR methods then are formulated by combining both measures into a single loss function, with opposite signs, balancing them against each other. Optimizing the loss function entails finding either the best compression with a fixed quality, or the highest quality at a fixed compression level. 

The earliest statistical algorithm formulated in this language is probably the  Principal Components Analysis (PCA) \cite{Hotelling1933}. It works by finding the rotation matrix that makes the covariance matrix of the system diagonal. In such a rotated basis, the new orthogonal features of the system would be arranged in descending order in terms of their explained variance (that is, their contribution to the quality of reconstruction), with the hope that only a few of those dimensions (strong compression) would be able to capture most of the variance of the original system. These preserved dimensions would then be called the latent features of the system. More recently with the advances in ML, neural network-based methods, such as autoencoders \cite{Salakhutdinov2006} and their variational counterparts \cite{Welling2014, Lerchner2016}, have become the state of the art in DR. Because of the neural networks' ability to approximate any continuous functions, such algorithms can search for nonlinear latent features that can minimize the loss function and describe the system better than linear methods, such as PCA. However, the simplicity and interpretability of PCA still make it a go-to method, even with all the recent advances in machine learning. Crucially, within the ML language, interpretability (or the lack of it) of latent features that emerge from nonlinear DR is a significant concern. This is especially true for neural network-based methods, where interpretability is almost nonexistent.

The similarity between the two languages is intriguing. Indeed,  what the physicists call relevant operators are in some situations what the statisticians call latent features. Correspondingly, there's a body of literature that explores those mappings \cite{mehta2014exact, bradde2017pca}, which can be exact in some scenarios. The biggest distinctions between the two are: (i) largely model-based approach of RG methods vs.\ data-driven structure of ML methods, (ii) explicit use of physical symmetries in RG approaches, which is only now being incorporated into ML methods, and (iii) deep theoretical justification, understanding, and interpretability of RG and its findings as opposed to {\em ad hoc} design and success and poor interpretability of ML methods. We hope that this dissertation will make a dent in the last point, providing better theoretical underpinnings behind some ML-based DR methods. 

By obtaining the latent features of the system and operating within this constructed low-dimensional space, we simplify our description of what initially appears to be a complex problem with many variables. This may make it easier to adequately sample and potentially model the system. The assumption that systems have latent low-dimensional descriptions is pervasive across various natural and social sciences fields, and applying DR methods to the analyzed data is sometimes the standard first step in any analysis in these fields. For example, in the analysis of neural activity \cite{Fairhall2016, Harris2021, Churchland2022, Harris2019}, behavior \cite{Harris2019, Ganguly2021, Churchland2022, Poeppel2017}, complex dynamical systems \cite{Tishby2009, Lipson2022}, or even in seemingly very different fields, such as economics \cite{Rudd2020, Tkacz2001, Gambetti2010, Zabotina2002} and systems management \cite{Lewis2013, Thompson2006}, the assumption of utilizing a low-dimensional description for a seemingly high-dimensional system is usually the default choice. Under which conditions such systems have low dimensional representations is a poorly understood --- and rarely studied --- question. This, however, typically does not stand in the way of DR methods being used, useful, and often successful in these fields. For example, it is standard to study the activity of an animal brain by first performing some form of PCA on it to obtain a low-dimensional description of the neural activity \cite{Fairhall2016}, and then to use it to predict some measured behavior (usually, the number of principal components preserved is of order 10). In some cases, latent variables can be interpreted. For example, in economics, it is common to assume the existence of common latent features among the observed prices (certainly an assumption, with weak theoretical justification \cite{jiang2017localized}). However, these latent features can then be interpreted as larger scale market indicators, such as the total market dynamics, inflation, or sector-specific indicators, etc.~\cite{Freyaldenhoven2022, Zabotina2002, Tkacz2001, Gambetti2010}.

Notably, when one talks about DR---including in the examples above---one typically reduces the dimensionality of a single large-dimensional set of variables. For example, in RG approaches, one combines microscopic spins into magnetization or the velocity of particles into the average velocity of a fluid in a mesoscopic volume. Similarly, in ML approaches to scientific data, one typically compresses data modalities one at a time. For example, in experiments on neural control of motor behavior, one separately finds a low-dimensional description of the neural activity and another low dimensional description of the behavior. However, while this might be sufficient to describe the system for some types of questions, it is not always enough. For example, in the neural control of behavior example, recording the neural activity without accounting for the resulting behavior makes it challenging to ensure that all the activity preserved during DR is truly relevant to the observed behavior, or that all the behavioral sequences identified by the DR are controlled neurally. Indeed, even a relatively simple organism like the {\em Drosophila melanogaster} fruit fly possesses hundreds of thousands of neurons across various connected brain regions \cite{zheng2018complete} and exhibits numerous stereotyped behaviors \cite{berman2014mapping}, which change depending on the environment. Consequently, the same neural activity may result in different behaviors, and vice versa. Some neural activity is just internal signal processing and is not behaviorally relevant, while some behavior is purely a mechanical response to the environment and not neurally controlled.

These types of problems, involving dimensionality reduction in more than one set of qualitatively different variables, which in the ML language are termed multiple \textit{views} or \textit{modalities}, present a unique set of additional advantages and challenges in a field known as multiview or multimodal learning. Combining multiple sources of observations in the analysis enables a more detailed and relevant description of the system, in the sense that it disentangles dependencies among the sources. In the example above, one can find patterns of neural activity that predict behavior, as well as stereotypical behaviors that seem to be neurally controlled.  However, with the benefits of multimodality, where each modality is often high-dimensional in its own right, the appropriate DR methods must necessarily become more complex, acknowledging the statistical structure of the data. For a single modality, DR is relatively straightforward: we need to preserve a quality of the description, while reducing its size.  However, even in the simplest multimodal situation, where we have only two modalities\footnote{While theoretically (and in some situations, practically \cite{Wen2015, Arora2017, Zeng2021, Hu2021}) one can consider more than two modalities (as demonstrated in~\ref{App:MultiViewLosses}), there are several limitations associated with doing so. For example, in the case of two modalities, one estimates covariance matrices, which have the dimensionality of the first modality times the second modality, and one needs to estimate elements of such matrix with good accuracy. However, for three modalities, one has to deal with a three-dimensional covariance tensor. This increases the number of elements needed to be estimated to guide the DR, and there is typically not enough data to do this well.  Additionally, visualizing and interpreting three-way relationships becomes more challenging. Moreover, in many practical scenarios, two modalities are sufficient for addressing many relevant questions (cf.~\ref{previous work} and \ref{observations in the litrature}). Thus, from now on in this Dissertation, and unless specified otherwise, when we talk about multimodal datasets, we analyze them by considering two modalities at a time.}, it is not immediately clear how to measure both the quality of the compression and the strength of the compression during DR. That is, one does not know which statistics of the two variables one needs to preserve, and the outcome of DR certainly depends on this choice. One prevalent approach in the literature boils down the multivariate approach to two univariate ones: one independently reduces the dimensionality of both modalities and subsequently seeks correlations between the low-dimensional descriptions. The rationale behind this approach is that, naively, it may require fewer samples\footnote{Naively, estimating the variance within each modality might seem simpler than estimating the covariance between two modalities, suggesting that fewer samples would be required. However, as we will demonstrate in Chapter 2, this assumption does not always hold true, and there are caveats to consider for this argument.} and is easier to implement\footnote{PCA, one of the simplest and most cited (about 6,910,000 papers mention it on Google Scholar at the time of writing) DR methods representing this approach, is essentially the eigenvalue problem \ref{pca}, with multiple efficient implementations available in almost any programming language and statistical software.}, with the hope that the most relevant dimensions for each modality might also be relevant for the other. In the language of linear models, reducing the multivariate problem to two single variate ones assumes that high variance directions within a modality correspond to high covariance directions across modalities. The validity of this assumption is unclear {\em a priori}. 

An alternative, albeit less commonly used, approach to multivariate DR problems is to jointly and {\em simultaneously} reduce both modalities to yield joint low-dimensional descriptions, so that the low-dimensional description of one modality is maximally informative about the low-dimensional description of the other, and vice versa. In linear terms, this entails keeping the features that explain the highest covariance between both modalities, rather than their variances. Some sporadic research suggests that this approach is more favorable in terms of sample efficiency (requiring smaller datasets to achieve similar -- or even better -- accuracy than independently reducing each modality), and that it provides more concise representations (keeping fewer dimensions) \cite{Shanechi2021, Newsted1999, Sarstedt2011}. Moreover, as we will demonstrate, in certain scenarios, independent reduction can lead to undesired outcomes by failing to capture the relevant features between modalities. This discrepancy can be intuitively understood, as there is no guarantee that maximum variation within one modality contributes to maximum covariation across them. For example, when recording a moving object of interest against a fast-changing background using two cameras, reducing the frames of each camera independently is likely to preserve details about the background more than preserving details about the object itself.

{\bf This Dissertation argues that simultaneous DR (SDR) approaches should almost always be preferred to independent DR (IDR) methods when the goal is to identify co-variation between two statistically related modalities.} Such methods require less data and produce more interpretable descriptions than independent single reduction approaches. Further, they are quite easy to implement numerically, certainly not qualitatively more difficult than their single modality counterparts. In the three subsequent Chapters of this Dissertation, we provide quantitative arguments for this assertion by investigating specific important questions about the design, the practice, and the interpretation of SDR methods on a wide variety of multimodal datasets. Specifically, we address the following problems
:
\begin{enumerate}
    \item In multimodal setups, how do the outcomes of the DR process depend on using IDR vs SDR methods? What outcomes (that is, quality and interpretability of latent descriptions) can we anticipate from each approach? How do standard methods employing these approaches function? and How can we leverage the theoretical understanding of the methods to improve the standard of practice?
    \item ML research has produced a multitude of DR methods, each differing in assumptions (such as whether the data is linear or not), implementation methods (linear algebra-based versus neural network-based), and more. This diversity has obvious advantages, but it also is overwhelming, limiting the ability to study and select the best of these methods for specific problems. We aspire to unify these methods within an intuitive yet mathematically tractable and rigorous framework. Ultimately, we aim to utilize this framework to develop methods that are not merely black-box solutions, but are tailored to specific research questions.
    \item We want to ascertain the usefulness of such an understanding of simultaneous reduction. Among the multitude of ML-based Mutual Information (MI) estimators (MI being a fundamental measure of statistical dependencies between two variables), we study the successful ones -- in terms of sample size requirements, consistency, and accuracy. We aim to understand them within our general framework of dimensionality reduction problem. We note that these methods effectively use a simultaneous reduction approach, which provides a low-dimensional description of the data that is maximally informative between the two variables.
\end{enumerate}

In more detail, in Chapter \ref{ch2}, I address the first of the above problems and investigate the effects of using SDR vs IDR methods in a specific multimodal setup. We explore how these different methods affect our results, under which conditions the true latent features can be discovered by different methods, and how this knowledge can be translated into a better practice. To answer these questions, we focus on an analytically and numerically tractable generative multimodal system. We employ a generative linear model with two modalities—each characterized by shared information between the modalities, self-information relevant for each one independently, and sample noise. The magnitudes of all of these features can be controlled independently.  We focus on commonly used,  powerful linear DR methods for both SDR and IDR. We explain the specific methods and why they may lead to distinct outcomes. Then we assess the quality of the low-dimensional representations obtained by applying these methods to data generated from the linear model, identifying key parameters relevant to the reduction process. Through the application of different DR methods from both approaches within this controlled setup, we gain essential insights into the DR process, shedding light on previously mentioned but poorly understood observations in the literature. Additionally, we propose a new heuristic to differentiate between self-information features unique to each modality and shared information features within multimodal datasets---a crucial step toward enhancing the interpretability and utility of DR outcomes. This Chapter serves as a motivational step, laying the groundwork for deeper analyses and methodological advancements explored in subsequent Chapters. Our aim here is to provide a comprehensive understanding of DR methods and their implications in tackling realistic, and yet tractable data.

In Chapter \ref{ch3}, I address the second of the above mentioned problems. There are many DR methods, each differing in assumptions and implementation choices. Different methods perform differently in distinct situations, but it is unclear how to choose a good method {\em a priori}. We systematize them by developing a comprehensive, mathematically rigorous, yet practical framework for unifying different DR methods, particularly the state-of-the-art deep variational ones. These methods utilize deep neural networks and variational approximations to learn robust and precise data representations, often acting as generative models for synthesizing samples from learned distributions. Additionally, using the framework, we can design new DR methods based on the needs of a practitioner: they need to specify what they want to preserve and how they believe the latent descriptions may depend on the observed data, and we automatically derive a corresponding DR algorithm and generate its neural network implementation. The framework is based on an interpretation of the information bottleneck (IB) principle. IB sets an explicit trade-off between the strength of compression and the quality of the latent description, both measured using information-theoretic quantities. More precisely, if we have a variable $X$ that has some relationship to another variable $Y$, IB works by compressing $X$ to a new low-dimensional variable $Z$ that shares the most relation with $Y$.  The quality of the low-dimensional latent space is measured using the mutual information (MI)\footnote{Mutual information between two continuous variables $X$ and $Y$ is a measure of all statistical dependencies between these two variables. It is defined formally as $I(X;Y) = \int\int p(x,y) \log \left( \frac{p(x,y)}{p(x)p(y)} \right) \,dx\,dy$, where $x$ and $y$ are particular values of the variables $X$ and $Y$, respectively. We discuss MI in depth in Chapters~\ref{ch3},~\ref{ch4}.} between $Z$ and $Y$, while the strength of the compression is measured by how much information the latent state preserves about the compressed variable, $I(X;Z)$. Overall, the IB loss function is $\mathcal{L} = I(X;Z) - \beta I(Y;Z)$. Here $\beta$ controls the quality of the compression, so that the information between $X$ and $Y$ is squeezed through a bottleneck of $Z$, giving the name to the method. In our more general framework, we trade off the information in an encoder graph, representing statistical structures used to derive the compressed variables (how the data should be encoded in a low-dimensional space), against that in a decoder graph, representing a generative model, which specifies how we want to reconstruct the variables of interest the low-dimensional compressed space. We then approximate probability distributions in these graphs via variational approximations, which can be learned with the help of neural networks. We named this framework the {\em Deep Variational Multivariate Information Bottleneck (DVMIB)}. Crucially,  we were able to retrieve multiple DR methods previously introduced in the literature as special cases of the framework. Additionally, we were able to improve and generalize some of these existing methods. Moreover, within this framework, we introduced a novel method, the {\em Deep Variational Symmetric Information Bottleneck (DVSIB)}, which is a specific variant of SDR that allows for simultaneous compression of the two modalities into distinct latent spaces that are maximally informative about one another. These latent representations are defined in two distinct spaces,  potentially with different physical units, a quality highly sought in different fields. For example, going back to the previously mentioned example of neural activity and behavior, DVSIB would reduce the original high-dimensional neural activity into a low-dimensional one, and similarly for the behavior, such that the new low-dimensional descriptions are the ones that are maximally informative, and yet the neural and the behavioral latent spaces only use the neural activity and the behavioral recordings, respectively, in their definition. Another example that has been used extensively recently is within the multiview learning framework, where we learn a joint low-dimensional space from two (or more) modalities, such as images and their corresponding descriptive text. The goal is to have two encoders (one for each modality) that separately map the image and text into a shared latent space, allowing it to perform various tasks such as classification within that space. Additionally, two decoders are trained and used for tasks such as image retrieval and text-to-image generation. In such a scenario, the state-of-the-art architecture—Contrastive Language-Image Pre-Training (CLIP) \cite{radford2021learning}—is indeed a simultaneous reduction approach. This can be encapsulated (and potentially improved) within our new method, DVSIB. In our tests, our new method achieves better results in terms of classification accuracy and succinct low-dimensional representations on some test problems. Although real-world applications are mostly beyond the scope of this dissertation, a discussion of different applications is included in the chapters, with an application to a real physical system in the Discussion (Chapter~\ref{ch5}). More interestingly, DVSIB verifies the previously developed intuition that the SDR paradigm is, indeed, more data-efficient than IDR, providing higher classification accuracy with a lower number of samples.

In  Chapter~\ref{ch4}, I study the problem of estimating Mutual Information (MI) as an SDR problem, aiming to provide efficient and precise guidelines for MI estimation in certain scenarios. MI naturally arises from fundamental principles in various fields like communication and probability theories to quantify statistical dependencies between variables. Despite its significance, accurately estimating MI from empirical data poses severe challenges, leading to the development of multiple estimation approaches over time, none of which work universally well. Traditional methods, such as parametric, nearest neighbor, and kernel-based methods, often struggle in high-dimensional scenarios. Conversely, recent neural network-based approaches have gained traction for their practicality and ability to handle high-dimensional data. However, these methods have mostly been tested in non-realistic, toy scenarios with effectively infinite amounts of data, which is not true of {\em any} realistic situation. The ability of these methods to work at finite sample sizes (i.e., the sample size dependencies of their biases and variances) is still poorly understood. These neural network methods, although versatile, lack inherent reporting criteria to warn users when their output should not be trusted. Since they are neural network-based methods, they are optimized via a training algorithm to minimize a certain loss function. However, the criteria for when to stop the optimization have not been established previously. Our analysis addresses these challenges by studying sample efficiency and providing accurate heuristics for determining when to terminate the training and whether to trust the output of the estimators.

In the last Chapter~\ref{ch5}, I discuss promising directions for further inquiry. For example, we can use a DVSIB-like architecture to extract compact representations of dynamical systems observed via large-dimensional measurements. In this context, the two modalities may be the past and the future recordings of the dynamical system observations, and the low-dimensional descriptions then represent the dynamics of the coarse-grained description. For instance, consider deriving the laws of motion of a simple physical pendulum, where observations are movie frames recording its motion over time. The past then consists of a time window of frames, and the future could entail subsequent time windows. Then ideally a DR algorithm would extract the low-dimensional variables that are directly related to the angle and the angular velocity of the pendulum. Indeed, I show that our method can extract low-dimensional descriptions of these movies that are these physical quantities! Crucially, the method works essentially out-of-the-box, with no need to impute a lot of domain-specific physical knowledge to result in an accurate inference. I discuss the different relevant parameters for this problem, and its potential effects on the results. Another avenue worth exploring is uncovering the shared low-dimensional structure between neural recordings and resultant behavior. We anticipate that these low-dimensional spaces, to be discovered by our method, would offer improved accuracy in decoding corresponding behavior, coupled with interpretability and modeling potential owing to their generative and low-dimensional nature. The body of literature in the neuroscience domain that considers behavior while reducing neural activity typically compresses only the recorded neural activity conditioned on the behavior \cite{Shanechi2021,hurwitz2021targeted,schneider2023learnable}, with a few exceptions that consider both \cite{gondur2023multi}. However, generally, this is not performing simultaneous reduction. As a result, we end up with a single low-dimensional space for the neural activity, while the behavior remains untouched (or processed separately), which is a limitation if the behavior is high-dimensional (such as videos or detailed motion). In the few exceptions where the behavior is also reduced, the neural activity and behavior are mixed together in a shared space, making it unclear how to interpret such a latent space. Having two distinct latent spaces for the two different domains as provided by DVSIB opens many interesting avenues worth studying.

Overall, this dissertation presents the work by me and my collaborators in advancing our understanding of DR methods through the lens of physics and information theory. By elucidating the differences between independent and simultaneous reduction techniques, I hope to have provided valuable insights into the underlying principles that can help in designing better physical models of complex systems. Moreover, the development of a unified framework has already facilitated the generalization of existing methods, while paving the way for the creation of novel approaches tailored to the specific challenges encountered in, but not limited to physics, machine learning, and life sciences research. Through the application of our methodologies, particularly in the estimation of mutual information, we have demonstrated their potential efficacy in extracting meaningful insights from data. This has implications not only for understanding fundamental physical phenomena, but also for optimizing experimental design and data analysis techniques in various fields.  As we continue to refine and apply these methods in different research areas, we can expect further advancements in our understanding of the methods, and also more demonstrations of their utility for analysis of complex natural and man-made systems.
\chapter{On the Difference Between Independent and Simultaneous Dimensionality Reduction}\label{ch2}

\section{Summary}
\footnote{This chapter presents the paper \cite{abdelaleem2024simultaneous} with the title \textit{Simultaneous Dimensionality Reduction: A Data Efficient Approach for Multimodal Representations Learning}. The work was conducted in collaboration with Ahmed Roman, K.\ Michael Martini, and Ilya Nemenman. I performed all simulations, conducted all analyses, and wrote the manuscript. Ilya conceived the model, while Ahmed and Michael contributed to discussions regarding the procedures and analyses. All authors reviewed the manuscript.}Current experiments frequently produce high-dimensional, multimodal datasets---such as those combining neural activity and animal behavior or gene expression and phenotypic profiling---with the goal of extracting useful correlations between the modalities. Often, the first step in analyzing such datasets is dimensionality reduction. We explore two primary classes of approaches to dimensionality reduction (DR): Independent Dimensionality Reduction (IDR) and Simultaneous Dimensionality Reduction (SDR). In IDR methods, of which Principal Components Analysis is a paradigmatic example, each modality is compressed independently, striving to retain as much variation within each modality as possible. In contrast, in SDR, one simultaneously compresses the modalities to maximize the covariation between the reduced descriptions while paying less attention to how much individual variation is preserved. Paradigmatic examples include Partial Least Squares and Canonical Correlations Analysis. Even though these DR methods are a staple of statistics, their relative accuracy and data set size requirements are poorly understood. We use a generative linear model to synthesize multimodal data with known variance and covariance structures to examine these questions. We assess the accuracy of the reconstruction of the covariance structures as a function of the number of samples, signal-to-noise ratio, and the number of varying and covarying signals in the data. Using numerical experiments, we demonstrate that linear SDR methods consistently outperform linear IDR methods and yield higher-quality, more succinct reduced-dimensional representations with smaller datasets. Remarkably, regularized CCA can identify low-dimensional weak covarying structures even when the number of samples is much smaller than the dimensionality of the data, which is a regime challenging for all dimensionality reduction methods. Our work corroborates and explains previous observations in the literature that SDR can be more effective in detecting covariation patterns in data. These findings strengthen the intuition that SDR should be preferred to IDR in real-world data analysis when detecting covariation is more important than preserving variation.

\section{Introduction}

Many modern experiments across various fields generate massive multimodal data sets. For instance, in neuroscience, it is common to record the activity of a large number of neurons while simultaneously recording the resulting animal behavior \citep{Harris2019, Harris2021, Churchland2022, Poeppel2017}. Other examples include measuring gene expressions of thousands of cells and their corresponding phenotypic profiles, or integrating gene expression data from different experimental platforms, such as RNA-Seq and microarray data \citep{Prior2013, Bielas2017, Teichmann2018, O'Donovan2015, forthe2018}. In economics, important variables such as inflation are often measured using combinations of macroeconomic indicators as well as indicators belonging to different economic sectors \citep{Tkacz2001, Zabotina2002, Freyaldenhoven2022, Rudd2020}. In all of these examples, an important goal is to estimate statistical correlations among the different modalities.

Analyses usually begin with dimensionality reduction (DR) into a smaller and more interpretable representation of the data. We distinguish two types of DR: {\em independent} (IDR) and {\em simultaneous} (SDR) \citep{martini2024data}. In the former, each modality is reduced independently, while aiming to preserve its variation, which we call {\em self} signal. In the latter, the modalities are compressed simultaneously, while maximizing the covariation (or the {\em shared} signal) between the reduced descriptions and paying less attention to preserving the individual variation. It is not clear if IDR techniques, such as the Principal Components Analysis (PCA) \citep{Hotelling1933}, are well-suited for extracting shared signals since they may overlook features of the data that happen to be of low variance, but of high covariance \citep{Brenner2014, Knutsson1997}. In particular, poorly sampled weak shared signals, common in high-dimensional datasets, can exacerbate this issue. SDR techniques, such as Partial Least Squares (PLS) \citep{Eriksson2001} and Canonical Correlations Analysis (CCA) \citep{Hotelling1936}, are sometimes mentioned as more accurate in detecting weak shared signal \citep{Newsted1999, Sarstedt2011, Theeramunkong2016}. However, the relative accuracy and data set size requirements for detecting the shared signals in the presence of self signals and noise remain poorly understood for both classes of methods.

In this study, we aim to assess the strengths and limitations of linear IDR, represented by PCA, and linear SDR, exemplified by PLS and CCA, in detecting weak shared signals. For this,  we use a generative linear model that captures key features of relevant examples, including noise, the self signal, and the shared signal components. Using this model, we analyze the performance of the methods in different conditions.  Our goal is to assess how well these techniques can (i) extract the relevant shared signal and (ii) identify the dimensionality of the shared and the self signals from noisy, undersampled data. We investigate how the signal-to-noise ratios, the dimensionality of the reduced variables, and the method of computing correlations combine with the sample size to determine the quality of the DR. We propose best practices for achieving high-quality reduced representations with small sample sizes using these linear methods.

\section{Model}
\subsection{Relations to Previous Work}\label{previous work}
The extraction of signals from large-dimensional data sets is a challenging task when the number of observations is comparable to or smaller than the dimensionality of the data. The undersampling problem introduces spurious correlations that may appear as signals, but are, in fact, just statistical fluctuations. This poses a challenge for DR techniques, as they may retain unnecessary dimensions or identify noise dimensions as true signals. 
Here, we focus exclusively on linear DR methods. For these, the Marchenko-Pastur (MP) distribution of eigenvalues of the covariance matrix of pure noise derived using the Random Matrix Theory (RMT) methods \citep{Pastur1967} has been used to introduce a cutoff between noise and true signal in real datasets. However, recent work \citep{Nemenman2022} has shown that, when observations are a linear combination of uncorrelated noise and latent low-dimensional self signals, then the self signals alter the distribution of eigenvalues of the sampling noise, questioning the validity of this naive approach.

Moving beyond a single modality, \cite{Potters2007} calculated the singular value spectrum of cross-correlations between two nominally uncorrelated random signals. However, it remains unknown whether the linear mixing of self signals and shared signals affects the spectra of noise, and how all of these components combine to limit the ability to detect shared signals between two modalities from data sets of realistic sizes. Filling in this gap using numerical simulations is the main goal of this work, and analytical treatment of this problem will be left for the future. 

The linear model and linear DR approaches studied here do not capture the full complexity of real-world data sets and state-of-the-art algorithms. However, if sampling issues and self signals limit the ability of linear DR methods to extract shared signals, it would be surprising for nonlinear methods to succeed in similar scaling regimes on real data. Thus extending the previous work to explicitly study the effects of linear mixtures of self signals, shared signals, and noise on limitations of DR methods is likely to endow us with intuition that is useful in more complex scenarios routinely encountered in different domains of science.

Examples of scenarios with shared and self signals include inference of dynamics of a system through a latent space \citep{Tishby2009, Lipson2022}, where shared signals correspond to latent factors that are relevant for predicting the future of the system from its past, while self signals correspond to nonpredictive variation \citep{Tishby2001}.
In economics, shared and self signals correspond to diverse macroeconomic indicators that are grouped into correlated distinct categories in structural factor models \citep{Gambetti2010, Tkacz2001, Rudd2020, Zabotina2002}. 
In neuroscience, shared signals can correspond to the latent space, by which neural activity affects behavior, while self signals encode neural activity that does not manifest in behavior and behavior that is not controlled by the part of the brain being recorded from \citep{Fairhall2015, Harris2019, Ganguly2021, Shanechi2021, Fairhall2016, Churchland2022, Poeppel2017}.

Interestingly, in the context of the neural control of behavior, it was noticed that SDR reconstructs the shared neuro-behavioral latent space more efficiently and using a smaller number of samples than IDR \citep{Shanechi2021}. Similar observations have been made in more general statistical contexts \citep{Newsted1999, Sarstedt2011, Theeramunkong2016,Maggioni2021}, though the agreement is not uniform \citep{Thompson2006, Thompson2012, Lewis2013}. Because of this, most practical recommendations for detecting shared signals are heuristic \citep{Sarstedt2021}, with widely acknowledged, but poorly understood limitations and possible resolutions \citep{Hadaya2018}. Our goal is to ground such rules in numerical simulations and scaling arguments.

\subsection{Linear Model with Self and Shared Signals}
\label{sec:model}
We consider a linear model with noise, $m_\text{self,X},m_\text{self,Y}$ self signals that are relevant to each modality independently, as well as $m_\text{shared}$ shared signals that capture the interrelationships between modalities.\footnote{This model is an extension of the model introduced by \cite{Nemenman2022}, and its probabilistic form has been studied by  \cite{Murphy2022}. In its turn, the latter is an extension of work by \cite{Kaski2012}, and \cite{Jordan2005}. However, within this model, we focus on the intensive limit, common in RMT \citep{potters2020first}, where the number of observations scales as the number of observed variables. This scenario is common in many real-world applications, and, to our knowledge, a similar extensive treatment to assess different DR methods as a function of various parameters of the system does not exist.} It results in $T$ observations of two high-dimensional standardized observables, $X$ and $Y$: 
\begin{eqnarray}
    \label{model}
    \nonumber \left[\tilde{X} \in \mathbb{R}^{N_X}\right] &= \underbrace{R_X}_{\text{Independent white noise}} + \underbrace{U_X V_X}_{\text{Self-Signal for X}} + \underbrace{P Q_X}_{\text{Shared-Signal}},\\
    \left[\tilde{Y} \in \mathbb{R}^{N_Y}\right] &= \underbrace{R_Y}_{\text{Independent white noise}} + \underbrace{U_Y V_Y}_{\text{Self-Signal for Y}} + \underbrace{P Q_Y}_{\text{Shared-Signal}},\\
  \label{model3}
    &X=\tilde{X}/\sigma_{\tilde{X}}, Y=\tilde{Y}/\sigma_{\tilde{Y}}.
\end{eqnarray}
The observations of $X$ and $Y$ are linear combinations of the following: (a) Independent white noise components $R_X$ and $R_Y$ with variances $\sigma^2_{R_X}$ and $\sigma^2_{R_Y}$. (b) Self-signal components $U_X$ and $U_Y$ residing in lower-dimensional subspaces $\mathbb{R}^{m_{\mathrm{self},X}}$ and $\mathbb{R}^{m_{\mathrm{self},Y}}$ with variances $\sigma^2_{U_X}$ and $\sigma^2_{U_Y}$. (c) Shared-signal components $P$ in a shared lower-dimensional subspace $\mathbb{R}^{m_{\mathrm{shared}}}$ with variance $\sigma^2_P$. These components are projected into their respective high-dimensional spaces $\mathbb{R}^{N_X}$ and $\mathbb{R}^{N_Y}$ using fixed quenched projection matrices $V_X$, $V_Y$, $Q_X$, and $Q_Y$ with specified variances $\sigma^2_{V_X}$, $\sigma^2_{V_Y}$, $\sigma^2_{Q_X}$, and $\sigma^2_{Q_Y}$, all respectively. Entries in these matrices are drawn from a Gaussian distribution with a zero mean and the corresponding variances.
Further, division by $\sigma_{\tilde{X}}$ and $\sigma_{\tilde{Y}}$ standardizes each column of the data matrices by their empirical standard deviations. The total variance in the matrix $\tilde{X}$ can be calculated as the sum of the variances of its individual components: 
$\sigma^2_{\tilde{X}} = \sigma^2_{R_X} + m_{\mathrm{self},X} \times \sigma^2_{U_X} \sigma^2_{V_X} + m_{\mathrm{shared}}\times  \sigma^2_P \sigma^2_{Q_X}$. A similar calculation can be done for the total variance in $\tilde{Y}$.

We define self and shared signal-to-noise ratios $\gamma_{\mathrm{self},X/Y}, \gamma_{\mathrm{shared},X/Y}$  as the relative strength of signals compared to background noise per component in each modality. These definitions allow us to examine how easily self or shared signals in each dimension can be distinguished from the noise.

\begin{equation}
\label{gamma}
\gamma_{\mathrm{self},X/Y} = \frac{\sigma^2_{U_{X/Y}}\sigma^2_{V_{X/Y}}}{\sigma^2_{R_{X/Y}}}, \quad \gamma_{\mathrm{shared},X/Y} = \frac{\sigma^2_{P}\sigma^2_{Q_{X/Y}}}{\sigma^2_{R_{X/Y}}}
\end{equation}

Our main goal is to evaluate the ability of linear SDR and IDR methods to reconstruct the shared signal $P$, while overlooking the effects of the self signals $U_{X/Y}$ on the statistics of the shared ones.

\section{Methods}

We apply DR techniques to $X$ and $Y$ to obtain their reduced dimensional forms $Z_X$ and $Z_Y$, respectively. $Z_X, Z_Y$ are of  sizes that can range from $T \times 1$ to $T \times N_X$ and $T \times N_Y$, respectively. As an IDR method, we use PCA \citep{Hotelling1933}. As SDR methods, we apply PLS \citep{Eriksson2001} and  CCA \citep{Hotelling1936, Vinod1976, Strother1998}, including both normal and regularized versions of the latter. Each of these methods focuses on specific parts of the overall covariance matrix 
\begin{equation} 
C_{X,Y} =
\begin{bmatrix}
C_{XX} & C_{XY}\\
C_{YX} & C_{YY}
\end{bmatrix} = \begin{bmatrix}
\frac{1}{T}X^\top X& \frac{1}{T}X^\top Y\\
\frac{1}{T}Y^\top X & \frac{1}{T}Y^\top Y
\end{bmatrix}.
\end{equation}
PCA aims to identify the most significant features that explain the majority of the {\em variance} in  $C_{XX}$ and $C_{YY}$, independently. PLS, on the other hand, focuses on singular values and vectors that explain the {\em covariance} component $C_{XY}$. Along the same lines, CCA aims to find linear combinations of $X$ and $Y$ that are responsible for the {\em correlation} ($C_{XY}/\sqrt{C_{XX}C_{YY}}$) between $X$ and $Y$. A detailed description of each method is in the next section \ref{a2}.

For every numerical experiment, we generate training and test data sets $(X_{\text{train}},Y_{\text{train}})$ and $(X_{\text{test}},Y_{\text{test}})$ according to Eqs.~(\ref{model}-\ref{model3})\footnote{We fix $\sigma^2_{R_{X/Y}},\sigma^2_{V_{X/Y}},\sigma^2_{Q_{X/Y}}$ and allow $\sigma^2_{U_{X/Y}},\sigma^2_{P}$ to vary when we choose $\gamma_{\mathrm{self},X/Y}, \gamma_{\mathrm{shared},X/Y}$. We first generate the fixed projection matrices $V_{X/Y}, Q_{X/Y}$, and we vary $R_{X/Y},U_{X/Y},P$  for each  trial.}. We apply PCA, PLS, CCA, and regularized CCA (rCCA) to the training to obtain the singular directions  $W_{X_\text{train}}$ and $W_{Y_{\text{train}}}$ for each method (see Appendix \ref{a2}). We then obtain the projections of the test data on these singular directions 
\begin{align}
\label{Ztest}
    \nonumber Z_{X} &= X_{\text{test}} W_{X_\text{train}},\\
    Z_{Y} &= Y_{\text{test}} W_{Y_\text{train}}.
\end{align}
Finally, we evaluate the {\em reconstructed correlations} metric $\mathcal{RC}'$, which measures how well these singular directions recover the shared signals in the data, corrected by the expected positive bias due to the sampling noise, see Appendix~\ref{a4} for details. $\mathcal{RC}'=0$ corresponds to no overlap between the true and the recovered shared directions, and $\mathcal{RC}'=1$ corresponds to perfect recovery. 

\subsection{Linear Dimensionality Reduction Methods}\label{a2}
\subsubsection{Principal Component Analysis
 - PCA}\label{pca}
PCA is a widely used linear IDR method that aims to find the orthogonal principal directions, such that a few of them explain the largest possible fraction of the variance within the data. PCA decomposes the covariance matrix of the data matrix $X$, $C_{XX} = \frac{1}{T}X^\top X$, into its eigenvectors and eigenvalues through singular value decomposition (SVD). The SVD yields orthogonal directions, represented by the vectors $w^{(i)}_X$, that capture the most significant variability in the data. In most numerical implementations \citep{Duchesnay2011}, these directions are obtained consecutively, one by one, such that the dot product between any two directions is zero $w_X^{(i)}\cdot w_X^{(j)}  =  \delta_{ij}$. The eigenvectors $w_X^{(i)}$ are obtained as the best solution to the optimization problem: \begin{equation}
    w^{*(i)}_{X} = \underset{w^{(i)}_{X}}{\mathrm{arg\max}} \frac{{w^{(i)}_{X}}^\top {X^{(i)}}^\top X^{(i)} w^{(i)}_{X}}{{w^{(i)}_{X}}^\top w^{(i)}_{X}}.
\end{equation}

Here $X^{(i)}$ is the $i$th deflated matrix where $X^{(1)}$ is the original matrix, and for every subsequent $i+1$, the matrix is deflated by subtracting the projection of $X$ on the obtained weights: $X^{(i+1)} = X-\Sigma_{s=1}^{i}X w_{(s)}w_{(s)}^\top$. The eigenvectors are sorted in decreasing order according to their corresponding eigenvalues, and the first $k$ eigenvectors $w_X^{(i=1:k)}$ are selected to form the projection matrix $W_X$. The obtained vectors determine the size of the reduced form $Z_X$, where $|Z_X| = k$ is the number of vectors retained from the decomposition of $X$. The vectors $w_X^{(i)}$ are then stacked together to form the projection matrix $W_X$. The low-dimensional representation $Z_X$ is then obtained by multiplying the original data matrix $X$ with this projection matrix, resulting in the reduced data matrix $Z_X = XW_X$. Similar treatment is done for $Y$ in order to obtain $Z_Y = YW_Y$

One of the main advantages of PCA is its simplicity and efficiency. However, one of the drawbacks of this method is that it performs DR for $X$ and $Y$ independently, and one then searches for relations between $Z_X$ and $Z_Y$ by regressing one on the other. Thus obtained low-dimensional descriptions may capture variance but not the covariance between the two datasets. 

\subsubsection{Partial Least Squares - PLS}\label{pls}
PLS, or Partial Least Squares, performs SDR by finding the shared signals that explain the maximum covariance between two sets of data \citep{Eriksson2001}. PLS performs the SVD of the covariance matrix $C_{XY} = \frac{1}{T}X^\top Y$ (or equivalently $C_{YX} = \frac{1}{T}Y^\top X$). The left and right singular vectors ($w^{*(i)}_{X}, w^{*(i)}_Y$) are obtained consecutively pair by pair such that $w_X^{(i)}\cdot w_Y^{(j)} = \delta_{ij}$. They are solutions of the optimization problem: 
\begin{equation}
(w^{*(i)}_{X},w^{*(i)}_{Y}) = \underset{w^{(i)}_{X},w^{(i)}_{Y}}{\mathrm{arg\max}} \frac{{w^{(i)}_{X}}^\top {X^{(i)}}^\top Y^{(i)} w^{(i)}_{Y}}{\sqrt{({w^{(i)}_{X}}^\top w^{(i)}_{X})({w^{(i)}_{Y}}^\top w^{(i)}_{Y})}}
\end{equation}
The matrices ${X^{(i)}}, Y^{(i)}$ are deflated in a similar manner to PCA \ref{pca}. The singular vectors are sorted in the decreasing order of their corresponding singular values, and the first $k$ vectors are selected to form the projection matrices $(W_X, W_Y)$. The obtained vectors determine the size of the reduced form $(Z_X, Z_Y)$, where $|Z_X| = |Z_Y| = k$ is the number of vectors retained. The vectors $(w_X^{(i)}, w_Y^{(i)})$ are then stacked together to form the projection matrices $(W_X, W_Y)$ respectively. The low-dimensional representations $(Z_X, Z_Y)$ are obtained by projecting the original data matrices $(X, Y)$ onto these projection matrices: $Z_X = XW_X$, and $Z_Y = YW_Y$.

In summary, PLS performs simultaneous reduction on both datasets, maximizing the covariance between the reduced representations $Z_X$ and $Z_Y$. This property makes PLS a powerful tool for studying the relationships between two datasets and identifying the underlying factors that explain their joint variability.

\subsubsection{Canonical Correlation Ananlysis - CCA}\label{cca}
\subsubsubsection{Normal CCA}
CCA is another SDR method, which aims to find the directions that explain the maximum correlation between two datasets \cite{Hotelling1936}. However, unlike PLS, CCA obtains the shared signals by performing SVD on the correlation matrix $\frac{C_{XY}}{\sqrt{C_{XX}}\sqrt{C_{YY}}}$. The singular vectors ($w^{*(i)}_{X}, w^{*(i)}_Y$) are obtained consecutively pair by pair such that $w_X^{(i)}\cdot w_Y^{(j)} = \delta_{ij}$.  CCA enforces the orthogonality of $w_X^{(i)}, w_Y^{(i)}$ independently as well, such that $w_X^{(i)}\cdot w_X^{(j)} = w_Y^{(i)}\cdot w_Y^{(j)} = \delta_{ij}$. The singular vectors are  obtained by solving  the optimization problem: 
\begin{equation}
    (w^{*(i)}_{X},w^{*(i)}_{Y}) = \underset{w^{(i)}_{X},w^{(i)}_{Y}}{\mathrm{arg\max}} \frac{{w^{(i)}_{X}}^\top {X^{(i)}}^\top Y^{(i)} w^{(i)}_{Y}}{\sqrt{({w^{(i)}_{X}}^\top {X^{(i)}}^\top X^{(i)} w^{(i)}_{X})({w^{(i)}_{Y}}^\top {Y^{(i)}}^\top Y^{(i)} w^{(i)}_{Y})}}.
\end{equation}
Like in PLS \ref{pls}, the matrices ${X^{(i)}}, Y^{(i)}$ are deflated in a similar manner. In addition, the first $k$ singular vectors $(w^{*(i)}_{X},w^{*(i)}_{Y})$ are stacked together to form the projection matrices $(W_X, W_Y)$, which then are used to obtain the reduced data matrices $Z_X = XW_X$, and $Z_Y = YW_Y$.

One of the key differences between PLS and CCA is that while both perform SDR, CCA also simultaneously performs IDR implicitly. Indeed, it involves multiplication of  $C_{XY}$ by $C_{XX}^{-1/2}$ on the left and $C_{YY}^{-1/2}$ on the right, which, in turn, requires finding singular values of the $X$ and the $Y$ data matrices independently.

\subsubsubsection{Regularized CCA - rCCA}
\label{rcca}
While CCA is a useful method for finding the maximum correlating features between two sets of data, it does have some limitations. Specfically,  in the undersampled regime, where $T\leq \max(N_X,N_Y)$, the matrices $C_{XX}$ and $C_{YY}$ are singular and their inverses do not exist. Using the pseudoinverse to solve the problem can lead to numerical instability and sensitivity to noise. Regularized CCA (rCCA) \citep{Vinod1976, Strother1998} overcomes this problem by adding a small regularization term to the covariance matrices, allowing them to be invertible. Specifically, one tales
\begin{align}
     \tilde{C}_{XX} & = C_{XX} + c_X I_X,\\
    \tilde{C}_{YY} & = C_{YY} + c_Y I_Y, 
\end{align}
where $\tilde{C}_{XX}, \tilde{C}_{YY}$ are the new regularized matrices, $c_X, c_Y>0$ are small regularization parameters and $I_X, I_Y$ are identity matrices with sizes $N_X\times N_X, N_Y\times N_Y$ respectively.

This original implementation of rCCA resulted in correlation matrices with diagonals not equal to one. Thus, a better implementation uses a different form of regularization \citep{Strother1998} by adding the regularization parameters $c_X$ and $c_Y$ individually to the equations as an affine combination (i.~e., $\sum_{i}^{n}c_i = 1$) as the following:
\begin{align}
    \tilde{C}_{XX} &= \frac{1}{T} (c_{X_1}w_X^\top X^\top X w_X+c_{X_2}w_X^\top w_X)\\
\tilde{C}_{YY} &= \frac{1}{T} ( c_{Y_1}w_Y^\top Y^\top Yw_Y+c_{Y_2}w_Y^\top w_Y).
\end{align}
This results in the regularized equations for $X$ and $Y$ to be:
\begin{eqnarray}
    \tilde{C}_{XX} & = \frac{1}{T} \big( (1-c_X)w_X^\top X^\top Xw_X+c_Xw_X^\top w_X \big)\\
    \tilde{C}_{YY} & = \frac{1}{T} \big( (1-c_Y)w_Y^\top Y^\top Yw_Y+c_Yw_Y^\top w_Y \big),
\end{eqnarray}
where $c_X$ and $c_Y$ are the regularization parameters, with values between 0 and 1, resulting in solving the optimization problem: 
\begin{equation}
\begin{multlined}
    (w^{*(i)}_{X},w^{*(i)}_{Y}) = \underset{w^{(i)}_{X},w^{(i)}_{Y}}{\mathrm{arg\max}} \quad {w^{(i)}_{X}}^\top {X^{(i)}}^\top Y^{(i)} w^{(i)}_{Y}\\
    \bigg{/} \bigg{(} \sqrt{(1-c_X)({w^{(i)}_{X}}^\top {X^{(i)}}^\top X^{(i)} w^{(i)}_{X})+c_X({w^{(i)}_{X}}^\top w^{(i)}_{X})} \quad \cdot \\
    \sqrt{(1-c_Y)({w^{(i)}_{Y}}^\top {Y^{(i)}}^\top Y^{(i)} w^{(i)}_{Y})+c_Y({w^{(i)}_{Y}}^\top w^{(i)}_{Y})}\bigg{)}
\end{multlined}
\end{equation}

Writing the regularization conditions in this form is in fact a convex interpolation problem between PLS and CCA, which is a more robust solution and does not suffer from shortening the length of correlations due to the added regularization. As a result, this implementation of rCCA achieves the best accuracy among all other methods.

\subsection{Assessing Success and Sampling Noise Treatment}\label{a4}

To assess the success of DR, we calculated the ratio between the total correlation between $Z_{X_{\rm test}}$ and $Z_{Y_{\rm test}}$, defined as in Eq.~(\ref{Ztest}), and the total correlation between $X$ and $Y$, which we input into the model. Specifically, we take the total correlation as the Frobenius norm of the correlation matrix, $||{A}||_F = \sqrt{\sum_{i}{\sigma_{i}^2(A)}}$, where $\sigma(A)$ are the singular values of the matrix $A$. Therefore, the metric of the quality of the DR is
\begin{equation}
\mathcal{RC} = \frac{||\text{Corr}(Z_{X_{\rm test}},Z_{Y_{\rm test}})||_{F}}{||\text{Corr}(P,P)||_{F}}= \frac{||\text{Corr}(Z_{X_{\rm test}},Z_{Y_{\rm test}})||_{F}}{m_{\rm shared}},
\label{eq:metric}
\end{equation}
where $\text{Corr}$ stands for the correlation matrix between its arguments, and we use $||\text{Corr}(P,P)||_{F}= m_\text{shared}$ as the total shared correlation that one needs to recover. Statistical fluctuations aside, $\mathcal{RC}$ should vary between zero (bad reconstruction of the shared variables) and one (perfect reconstruction).

\begin{figure}[h]
  \begin{center}
    \includegraphics[width=0.75\textwidth]{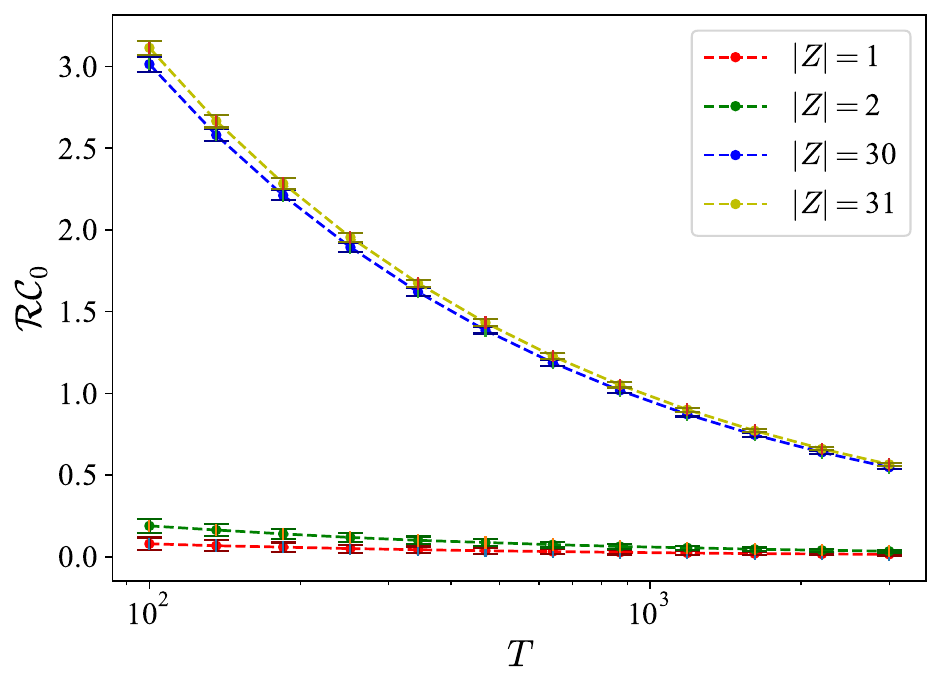}
  \end{center}
  \caption[Correlations due to random matrices]{The resulting correlations are averages of all the points in the phasespace, then averaged over 10 different realizations of the matrices. The error bars are for two standard deviations around the mean}
  \label{pd1-noise}
\end{figure}

In many real-world applications, the number of available samples, $T$, is often limited compared to the dimensionality of the data, $N_X$ and $N_Y$. This undersampling can introduce spurious correlations. We are not aware of analytical results to calculate the effects of the sampling noise on estimating singular values in the model in Eq.~(\ref{model}) \citep{Potters2017}. Thus, to estimate the effect of the sampling noise, we adopt an empirical approach. Specifically, we generate two random matrices, $Z_{X_{\text{random}}}$ and $Z_{Y_{\text{random}}}$, of sizes $T \times |Z_X|$ and $T \times |Z_Y|$, respectively. We then calculate the correlation between these matrices, denoted as $\mathcal{RC}_0$, for multiple such trials using the metric in Eq.~(\ref{eq:metric}). For random $Z_{X_{\text{random}}}$ and $Z_{Y_{\text{random}}}$, $\mathcal{RC}$ should be zero. However, Fig.~\ref{pd1-noise} shows that, especially for large dimensionalities of the compressed variables and small $T$, the sampling noise results in a significant spurious $\mathcal{RC}_0>0$, which may even be larger than 1! Crucially, $\mathcal{RC}_0$ does not fluctuate around its mean across trials, so that the sampling bias is narrowly distributed. 

To compensate for this sampling bias, we subtract it from the reconstruction quality metric, 
\begin{equation}
\label{rcprime}
    \mathcal{RC}'=\mathcal{RC} -\mathcal{RC}_0.
\end{equation}
It is this $\mathcal{RC}'$ that we plot in all Figures in this chapter as the ultimate metric of the reconstruction quality. While subtracting the bias is not the most rigorous mathematically,  it provides a practical approach for reducing the effects of the sampling noise. 

\subsection{Implementation}
\label{a5}
We used Python and the \url{scikit-learn} \citep{Duchesnay2011} library for performing PCA, PLS, and CCA, while the \url{cca-zoo} \citep{Wang2021} library was used for rCCA. For PCA, SVD was performed with default parameters. For PLS, the PLS Canonical method was used with the NIPALS algorithm. For both PLS and CCA, the tolerance was set to $10^{-4}$ with a maximum convergence limit of 5000 iterations. For rCCA, regularization parameters were set as $c_1 = c_2 = 0.1$. All other parameters not explicitly here were set to their default values.

All figures shown in this chapter were averaged over 10 independent realizations of $R_X, R_Y, U_X, U_Y, P$, while fixing the projection matrices $V_X, V_Y, Q_X, Q_Y$. We then performed an additional round of averaging everything over 10 realizations of the projection matrices themselves. The simulations were parallelized and run on Amazon Web Services (AWS) servers of instance types \url{ml.c5.2xlarge}.

\section{Results}
\subsection{Results of the Linear Model}
We perform numerical experiments to explore the undersampled regime, $T\lesssim N_X, N_Y $. We use $T = \{100, 300, 1000, 3000\}$ samples, $N_X = N_Y = 1000$. We explore the case of one shared signal only, $m_\text{shared} = 1$ and we mask this shared signal by a varying number of self signals and noise. We vary the number of retained dimensions, ($|Z_X|, |Z_Y|$), and explore how many of them are needed to recover the shared signal in the noise and the self signal background with different SNR.

For brevity, we explore two cases: (1) One self-signal in $X$ and $Y$ in addition to the shared signal ($m_\text{self}=1$); (2) many self-signals in $X$ and $Y$. For both cases, we calculate the quality of reconstruction as the function of the shared and the self SNR,  $\gamma_\text{shared}$ and  $\gamma_\text{self}$. In all figures, we show $\mathcal{RC}'$ for severely undersampled (first row, $T=300$) and relatively well sampled (second row, $T=3000$) regimes. We also show the value of $\mathcal{RC}_0$, the bias that we removed from our reconstruction quality metric, for completeness, see section \ref{a4} for details.

\subsection{One self-signal in \texorpdfstring{$X$} and \texorpdfstring{$Y$} in addition to the shared signal (\texorpdfstring{$m_\text{self}=1$}))}
\subsubsection{Keeping 1 dimension after reduction ($|Z_{X/Y}| = 1$)}

\begin{figure}[hp]
    \centering
    \includegraphics[width=\textwidth]{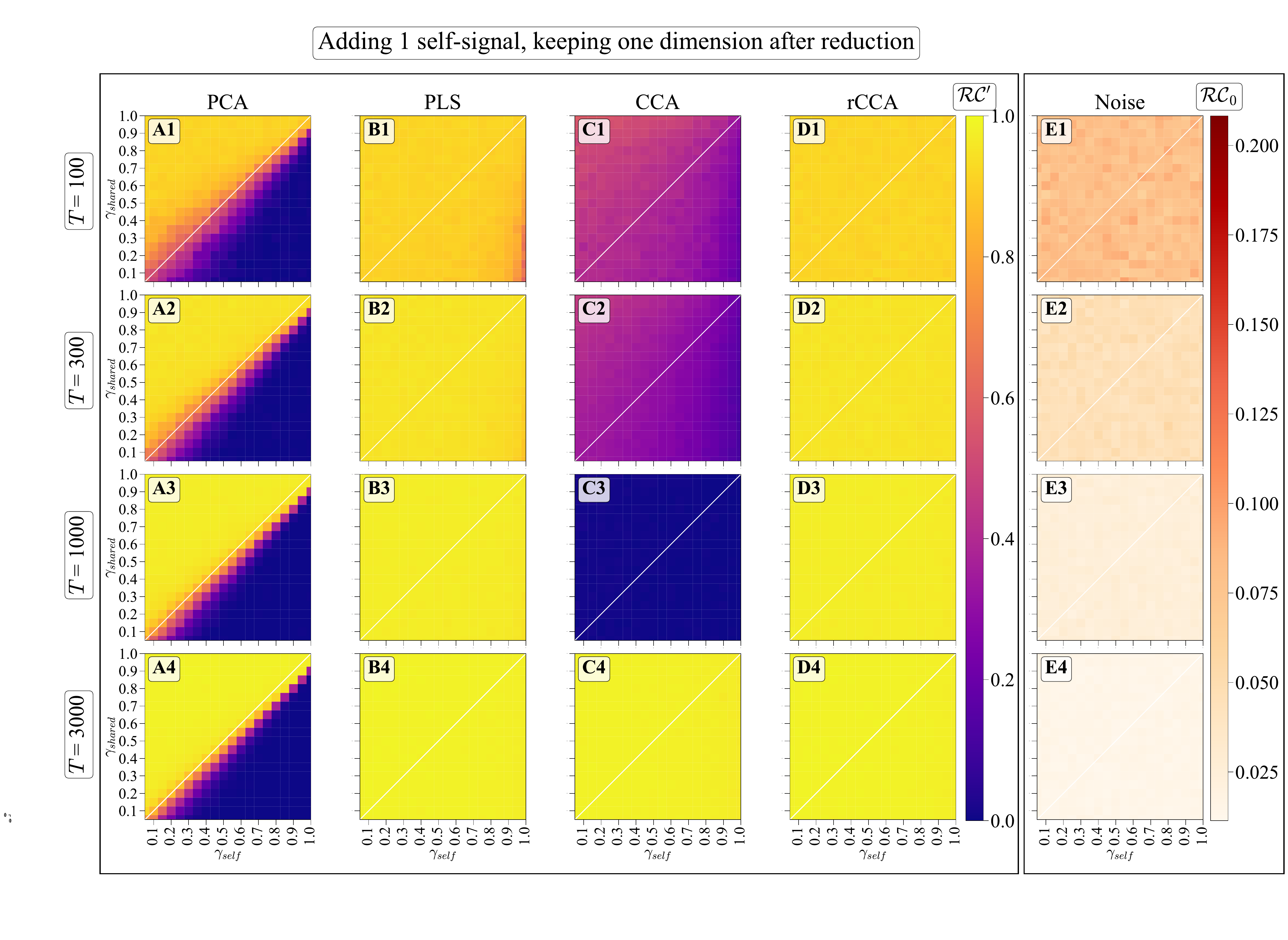}
    \caption[Results of One Shared Signal and One Self Signal, One Dimension Kept After Reduction]{Performance of PCA, PLS, CCA, rCCA, and noise in recovery of the shared signal for $|Z_X| = |Z_Y| = 1 = m_\text{self}$. PCA struggles to detect shared signals when they are weaker than the self signals. PLS and rCCA demonstrate nearly perfect reconstruction. CCA  displays no reconstruction in the undersampled regime $T\ll N_X$, and it is nearly perfect for large $T$.}
    \label{pd1-mx1-zx1}
\end{figure}

Figure \ref{pd1-mx1-zx1} shows that, in Case 1, when one dimension is retained in DR of $X$ and $Y$, PCA populates the compressed variable with the largest variance signals and hence struggles to retain the shared signal when $\gamma_{\rm self}>\gamma_{\rm shared}$, regardless of the number of samples. However, both PLS and rCCA excel in achieving nearly perfect reconstructions. When $T \ll N_X$, straightforward CCA cannot be applied (see \ref{cca}-\ref{rcca}), but it too achieves a perfect reconstruction when $T > N_X$.

\subsubsection{Keeping 2 dimensions after reduction ($|Z_{X/Y}| = 2$)}

\begin{figure}[hp]
    \centering
    \includegraphics[width=\textwidth]{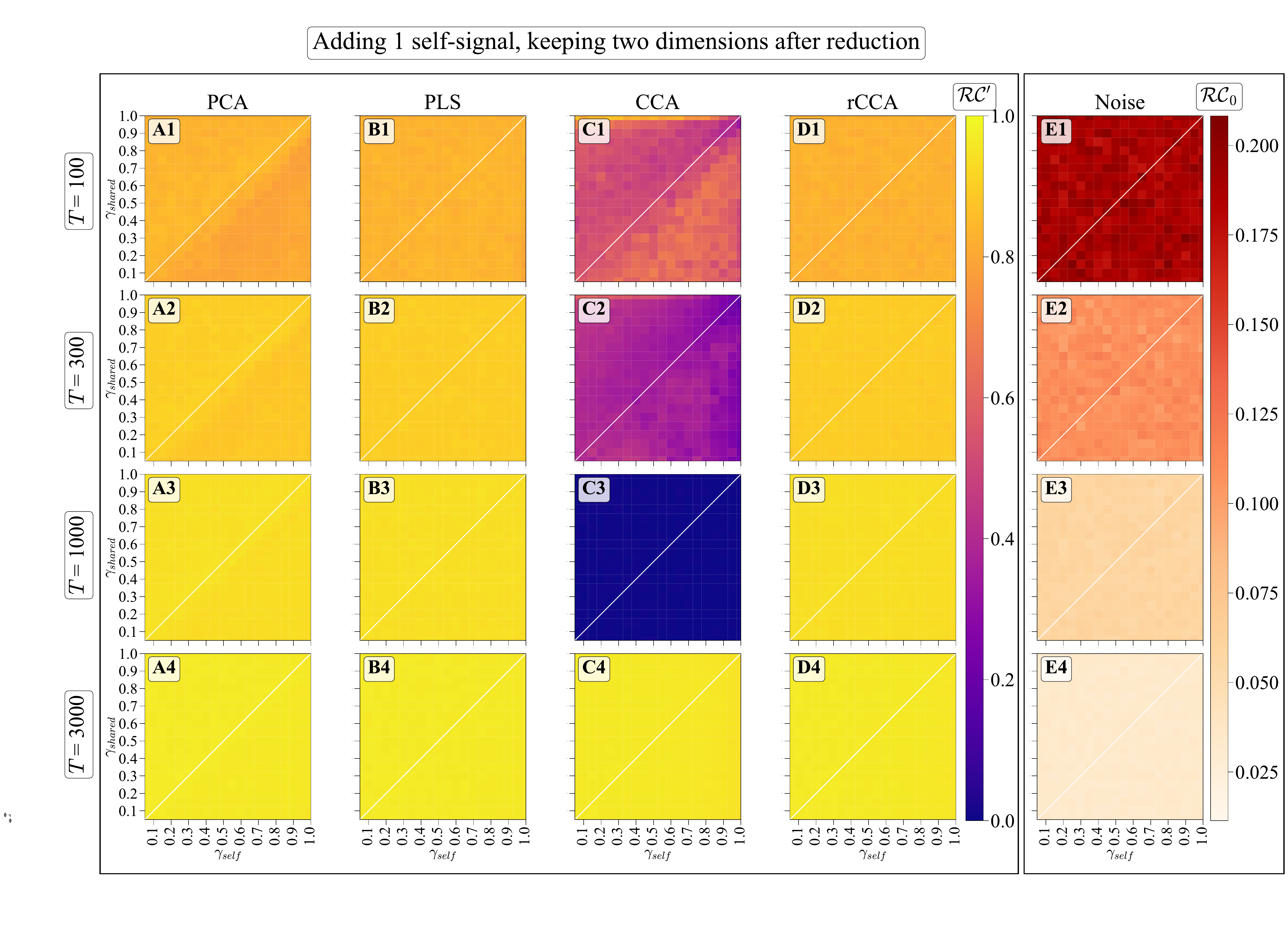}
    \caption[Results of One Shared Signal and One Self Signal, Two Dimensions Kept After Reduction]{Same as Fig.~\ref{pd1-mx1-zx1}, but for $|Z_X| = |Z_Y| = 2 = m_\text{self} + m_\text{shared}$. Now there are enough compressed variables for PCA to detect the shared signal. Other methods perform similarly to Fig.~\ref{pd1-mx1-zx1}, albeit the noise is larger.}
    \label{pd1-mx1-zx2}
\end{figure}

In Fig.~\ref{pd1-mx1-zx2}, we allow two dimensions in the reduced variables. For PCA, we expect this to be sufficient to preserve both the self and the shared signals. Indeed, PCA now works for all $\gamma$s and $T$s, although with a slightly reduced accuracy for large shared signals compared to Fig.~\ref{pd1-mx1-zx1}. PLS and rCCA continue to deliver highly accurate reconstructions. So does the CCA for $T>N_X$. Spurious correlations, as measured by $\mathcal{RC}_0$ grow slightly with the increasing dimensionality of $Z_X$, $Z_Y$ compared to Fig.~\ref{pd1-mx1-zx1}. This is expected since more projections must now be inferred from the same amount of data.

\subsection{Many self-signal in \texorpdfstring{$X$} and \texorpdfstring{$Y$} in addition to the shared signal (\texorpdfstring{$m_\text{self}=30$}))}
\subsubsection{Keeping 1 dimension after reduction ($|Z_{X/Y}| = 1$)}

\begin{figure}[hp]
    \centering
    \includegraphics[width=\textwidth]{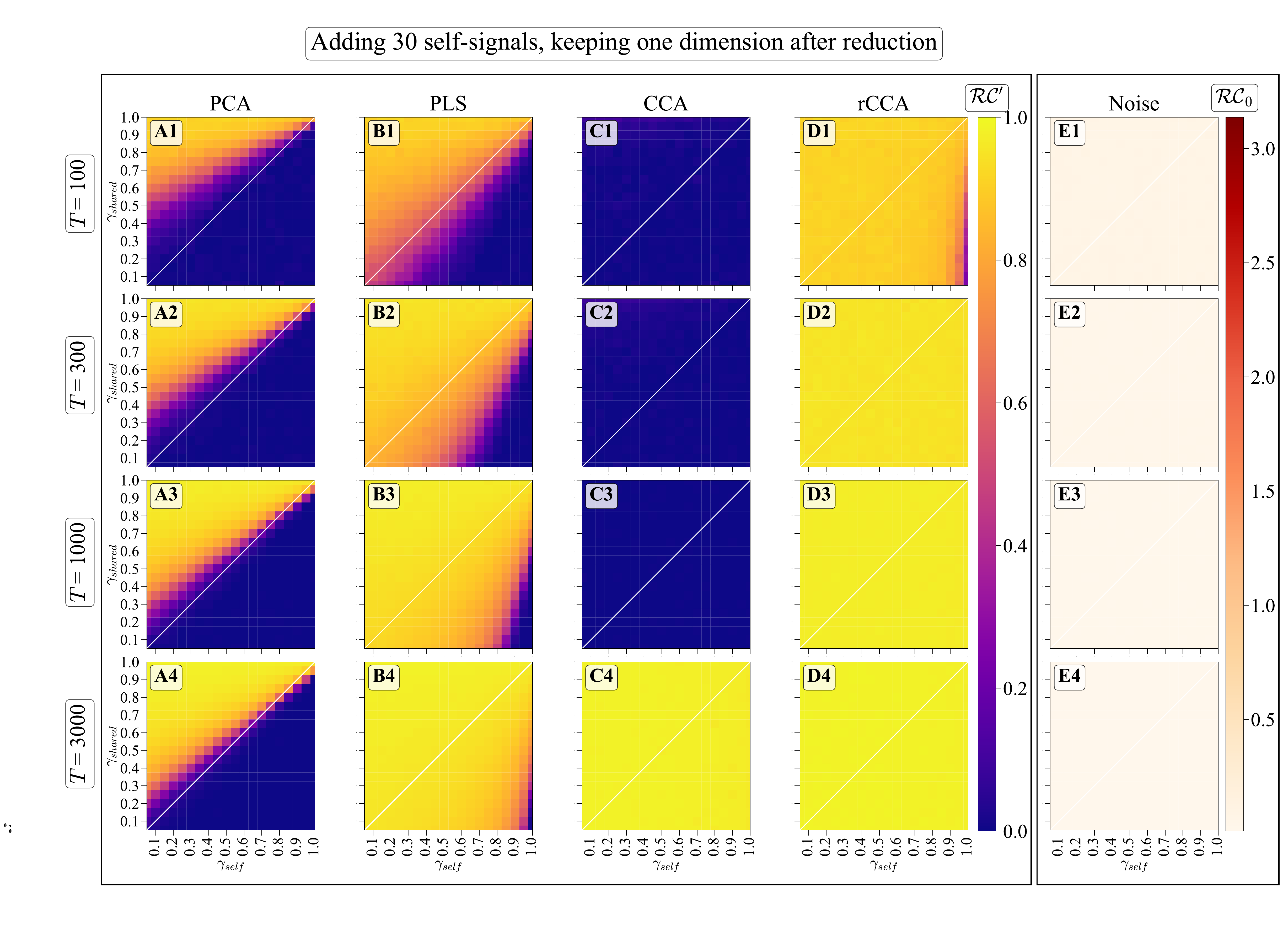}
    \caption[Results of One Shared Signal and Thirty Self Signals, One Dimension Kept After Reduction]{Reconstruction results for $m_{\rm self}=30$, $m_{\rm shared}=1$, and $|Z_X| = |Z_Y| = 1$. PCA struggles to detect any shared signals when they are even comparable to the self ones.  PLS performance also degrades.  CCA displays its usual impotence at small $T$. Finally, rCCA demonstrates nearly perfect reconstruction for all parameter values.}
    \label{pd1-mx30-zx1}
\end{figure}

We now turn to $m_{\rm self}\gg m_{\rm shared}$. We use $m_{\text{shared}} = 1$, $m_{\text{self}} = 30$ for concreteness. We expect that the performance of SDR methods will degrade weakly, as they are designed to be less sensitive to the masking effects of the self signals. In contrast, we expect IDR to be more easily confused by the many strong self-signals, degrading the performance. Indeed, Fig.~\ref{pd1-mx30-zx1} shows that PCA now faces challenges in detecting shared signals, even when the self signals are weaker than in Fig.~\ref{pd1-mx1-zx1}. Increasing $T$ improves its performance only slightly. Somewhat surprisingly, PLS performance also degrades, with improvements at $T\gg N_X$. CCA again displays no reconstruction when $T\ll N_X$, switching to near perfect reconstruction at large $T$. Crucially, rCCA again shines, maintaining its strong performance, consistently demonstrating nearly perfect reconstruction.

\subsubsection{Keeping 30 dimensions after reduction ($|Z_{X/Y}| = 30$)}

\begin{figure}[hp]
    \centering
    \includegraphics[width=\textwidth]{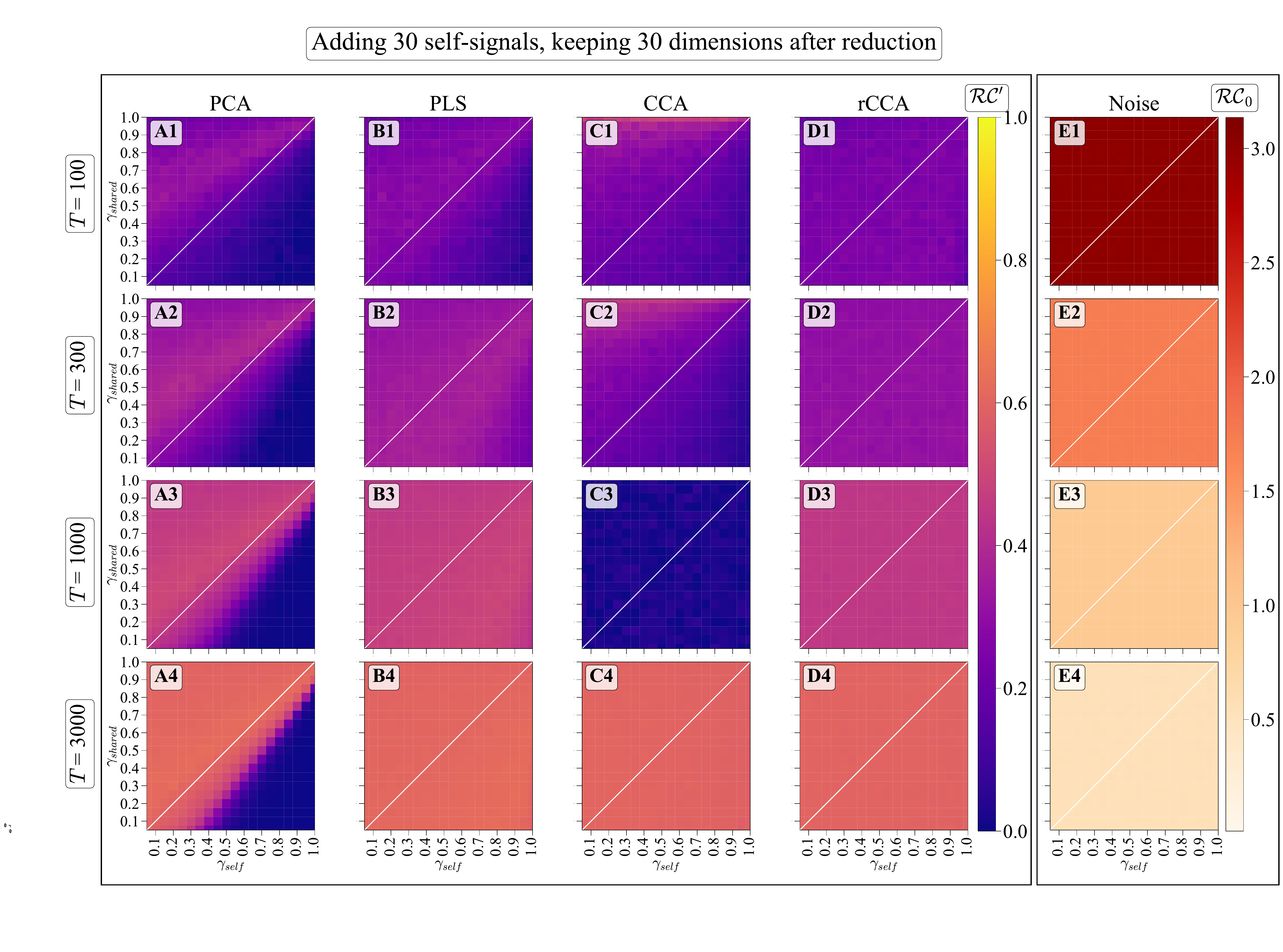}
    \caption[Results of One Shared Signal and Thirty Self Signals, Thirty Dimension Kept After Reduction]{DR performance for $|Z_X| = |Z_Y| = m_\text{self} > m_\text{shared}$). PCA now detects shared signals even when they are weaker than the self signals. However, the quality of reconstruction is significantly lower than in Fig.~\ref{pd1-mx1-zx2}. PLS detects signals in a larger part of the phase space, but also with a significant reduction in quality, which improves with sampling. CCA has its usual problem for $T\ll N_X$, and, like PLS, it has a significantly lower reconstruction quality than in the regime in Fig.~\ref{pd1-mx30-zx1}. rCCA is able to detect the signal in the whole phase space, but again with worse quality. Finally, spurious correlations are high, though they decrease with better sampling.}
    \label{pd1-mx30-zx30}
\end{figure}

Since one retained dimension is not sufficient for PCA to represent the shared signal when $\gamma_{\rm shared}\lesssim \gamma_{\rm self}$, we increase the dimensionality of reduced variables $|Z_X| = |Z_Y| = m_\text{self} \gg m_\text{shared}$), cf.~Fig.~\ref{pd1-mx30-zx30}. PCA now detects shared signals even when they are weaker than the self-signals, $\gamma_{\rm shared}<\gamma_{\rm self}$, but at a cost of the reconstruction accuracy plateauing significantly below 1. In other words, when self and shared signals are comparable, they mix, allowing for partial reconstruction. However, even at $T\gg N_X$, PCA cannot break into the phase diagram's lower right corner. Other methods perform similarly, reconstructing shared signals over the same or wider ranges of sampling and the SNR ratios than in Fig.~\ref{pd1-mx30-zx1}. For all of them, the improvement comes at the cost of the decreased asymptotic performance. The most distinct feature of this regime is the dramatic effect of noise, where 30-dimensional compressed variables can accumulate enough sampling fluctuations to recover correlations that are supposedly nearly twice as high as the data actually has.

\subsubsection{Keeping 31 dimensions after reduction ($|Z_{X/Y}| = 31$)}
\begin{figure}[hp]
    \centering
    \includegraphics[width=\textwidth]{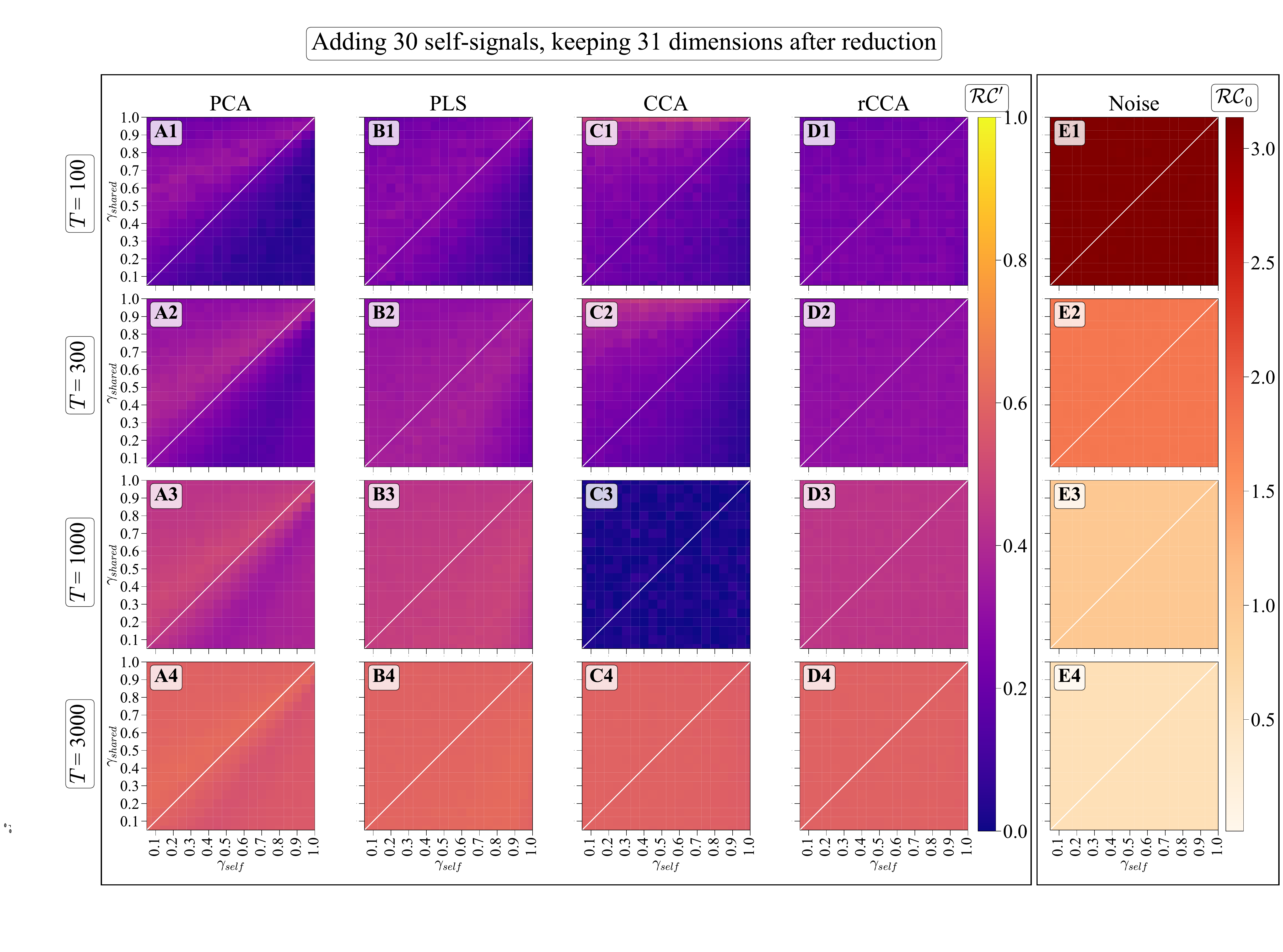}
    \caption[Results of One Shared Signal and Thirty Self Signals, Thirty-one Dimension Kept After Reduction]{PCA, PLS, CCA, rCCA, and noise results when 31 dimensions are kept after reduction ($|Z_X| = |Z_Y| = m_\text{self} + m_\text{shared}$). PCA now can detect more shared signals when they are weaker than the self signals (A1), however, with a significantly lower quality compared to figure \ref{pd1-mx1-zx2}, but suddenly explores the whole phase space, still with lower accuracy than Case 1. PLS, CCA, rCCA, and noise show similar behavior to figure \ref{pd1-mx30-zx30}.}
    \label{pd1-mx30-zx31}
\end{figure}

Figure \ref{pd1-mx30-zx31} now explores a regime when the dimensionality of the compressed variables is enough to store both the self and the shared interactions at the same time, $|Z_X| = |Z_Y| = m_\text{self} + m_\text{shared}=31$. With just one more dimension than Fig.~\ref{pd1-mx30-zx30}, PCA abruptly transitions to being able to recover shared signals for all SNRs, albeit still saturating at a far from perfect performance at large $T$. PLS, CCA, rCCA, and noise show behavior remain similar to Fig.~\ref{pd1-mx30-zx30}.

\subsection{Key Parameters and Testing Technique for Dimensionality of Self and Shared Signals}

Our analysis suggests that there are three relevant factors that determine the ability of DR to reconstruct shared signals. The first is the strength of the shared and the self signals compared to each other and to noise. For brevity, in the following  analysis, we fix $\gamma_{\text{self}}$ and define the ratio $\tilde{\gamma} = \gamma_{\text{shared}} / \gamma_{\text{self}}$ to represent this effect. The second factor affecting the performance is the ratio between the number of shared and self signals, denoted by $\tilde{m} = m_{\text{shared}} / m_{\text{self}}$. The third factor is the number of samples per dimension of the reduced variable, denoted by $\tilde{q} = T/|Z|$. 

\begin{figure}[hp]
    \centering
    \includegraphics[width=\textwidth]{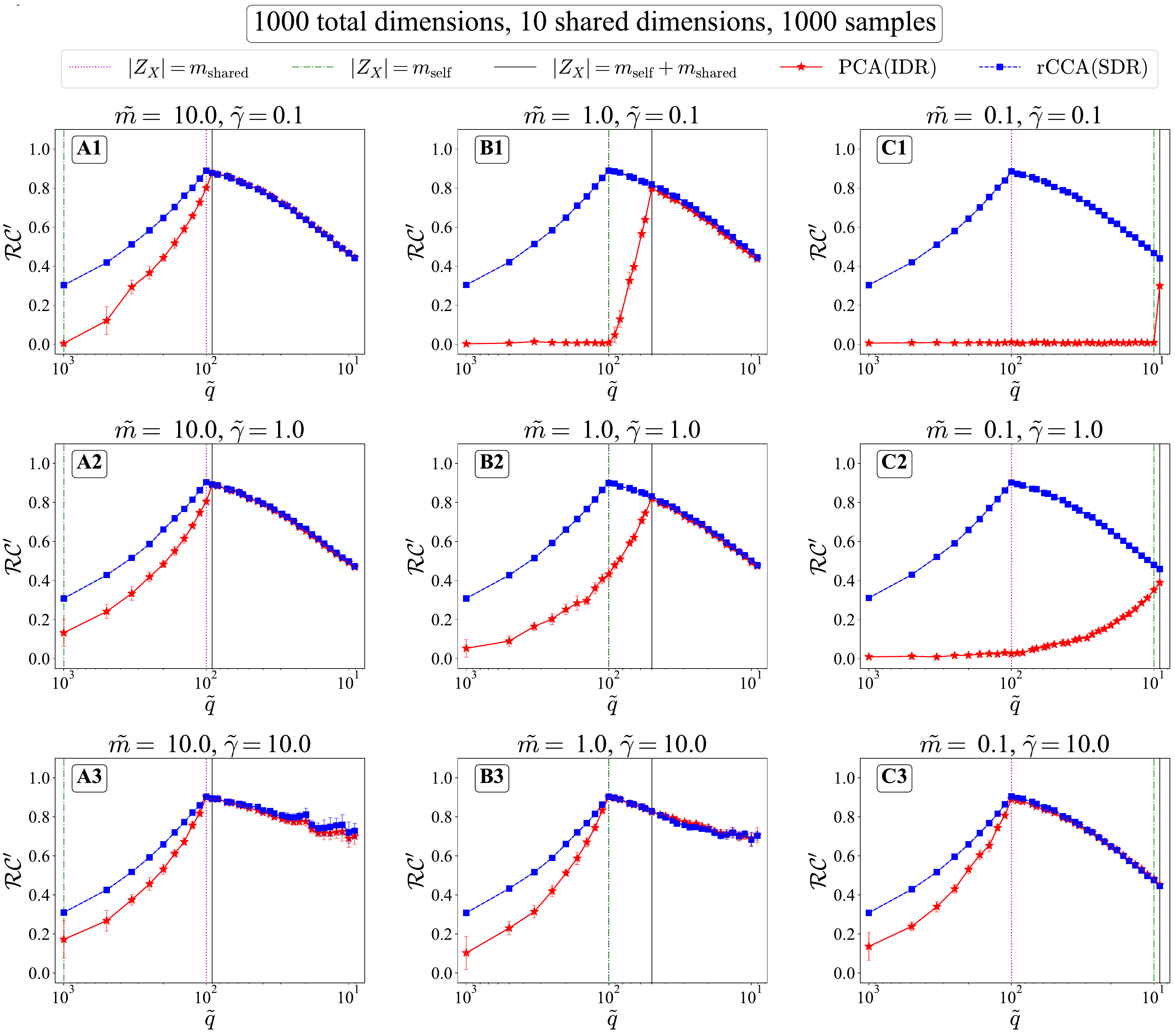}
    \caption[Summary of Results Across Various Regimes and Parameters]{Performance of PCA (IDR) and rCCA (SDR) for different values of the relevant parameters of the model: the number of samples per dimension of the compressed variable ($\tilde{q}$), the strength of shared signals relative to the self ones ($\tilde{\gamma}$), and the ratio of the number of shared to self signal components ($\tilde{m}$), while fixing the number of samples ($T = 1000$) and the number of shared dimensions ($m_{\rm shared} = 10$). Note that decreasing $\tilde{q}$ (left to right) corresponds to increasing the dimension of the latent space $|Z_X|$ at a fixed number of samples $T$.}
    \label{pd2-t1000}
\end{figure}

In Fig.~\ref{pd2-t1000}, we illustrate how these parameters influence the performance of DR, $\mathcal{RC}'$. Each subplot varies $\tilde{q}$, while holding $T$ constant and changing $|Z_X|$. We compare the results of PCA (representing IDR) and rCCA (representing SDR). Each curve is averaged over 10 trials, with error bars indicating 1 standard deviation around the mean, using algorithmic parameters as described in section \ref{a5}.

We see that the relative strength of signals, as represented by $\tilde{\gamma}$, plays a significant role in determining which method performs better. If the shared signals are larger (bottom) both approaches work.  However, for weak shared signals (top), SDR is generally more effective. Further, the ratio between the number of shared and self signals, $\tilde{m}$, also plays an important role. When $\tilde{m}$ is large (left), IDR is more likely to detect the shared signal before the self signals, and it approaches the performance of SDR.  However, when $\tilde{m}$ is small, IDR is more likely to capture the self signals before moving on to the shared signals, degrading performance (right). Finally, not surprisingly, the number of samples per dimension of the compressed variables, $\tilde{q}$, is also critical to the success. If $\tilde{q}$ is small, the signal is drowned in the sampling noise, and adding more retained dimensions hurts the DR process. This expresses itself as a peak for SDR performance around $|Z_X|=m_{\text{shared}}$. For IDR, the peak is around $|Z_X|=m_{\text{self}}+m_{\text{shared}}$, thus requiring more data to achieve performance similar to SDR.

We observe that the performance of rCCA (SDR) is almost independent of changing $\tilde{m}$ or $\tilde{\gamma}$, indicating that it focuses on shared dimensions even if the latter is masked by self signals. The algorithm crucially depends on $\tilde{q}$, where adding more dimensions (decreasing $\tilde{q}$) than needed hurts the reduction. This is because, for a fixed number of samples, the reconstruction of each dimension then gets worse. In contrast, for PCA (IDR), the performance depends on all three relevant parameters, $\tilde{q}$, $\tilde{m}$, and $\tilde{\gamma}$. At some parameter combinations, the performance of IDR in reconstructing shared signals approaches SDR. However, in all cases, SDR never performs worse than IDR on this task.

\subsection{Beyond Linear Models - Noisy MNIST}

\label{noisy_mnist}
\subsubsection{The Dataset}

\begin{figure}[ht]
\begin{center}
\includegraphics[width=.85\textwidth]{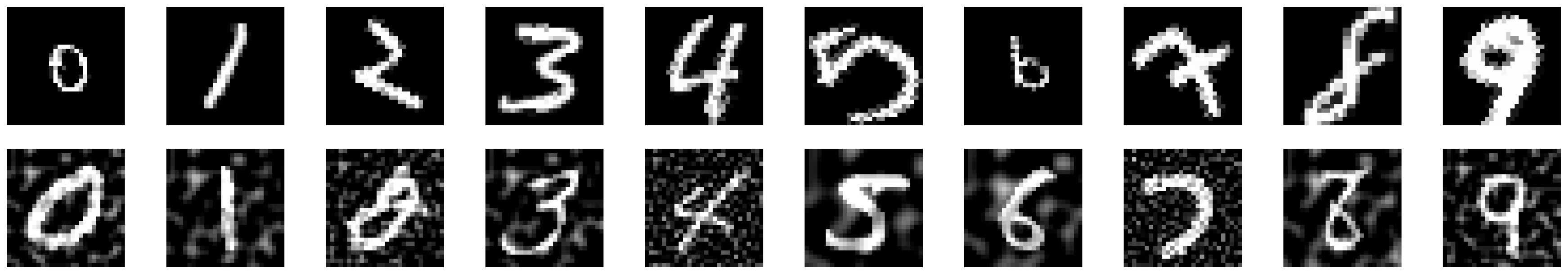}
\includegraphics[width=.85\textwidth]{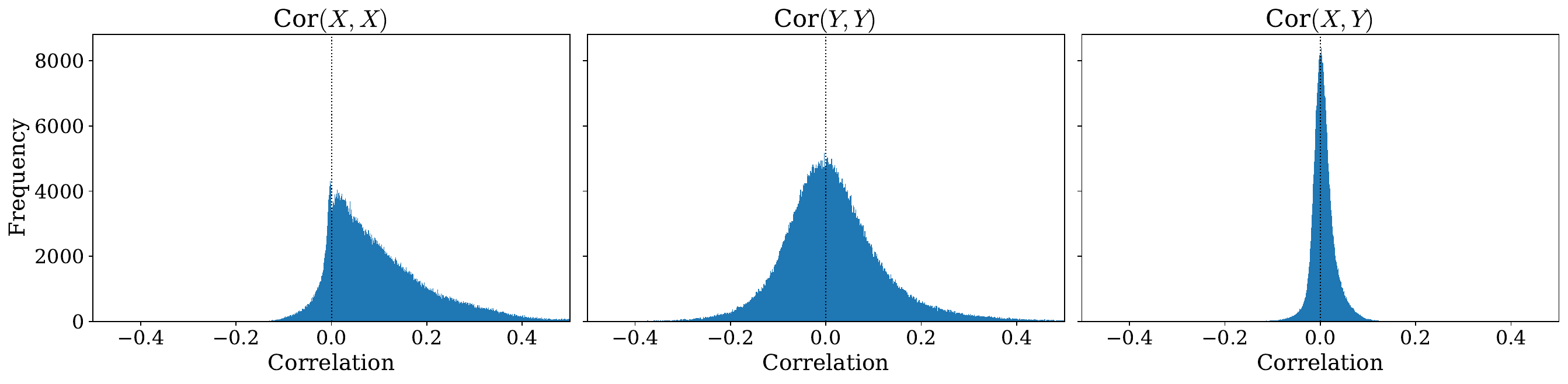}
\end{center}
\caption{Dataset containing paired MNIST digit samples sharing only the same identity \textit{(shared signal)}. The first row ($X$) shows MNIST digits randomly subjected to scaling, $(0.5-1.5)$, and rotation with an angle of $(0-\pi/2)$, while the second row ($Y$) shows MNIST digits with an added background Perlin noise \textit{(self signals)}.
In the bottom row, histograms of self correlations for the $X$ and $Y$ datasets (left and middle, respectively) illustrate a wide range of correlations, while the histogram of the cross correlation between $X$ and $Y$ (right) demonstrates a smaller range.}
\label{Fig:data_ch1}
% \vspace{-.2in}
\end{figure}

To analyze linear DR methods on nonlinear data, we followed the same procedure as in Fig.~\ref{pd2-t1000} for a dataset inspired by the noisy MNIST dataset \citep{Haffner1998, Bilmes2015, Livescu2016, Abdelaleem2023}. This dataset has two distinct views of data, each of dimensionality $28 \times 28$ pixels, examples of which are shown in Fig.~\ref{Fig:data_ch1}. The first view is an image of the digit subjected to a random rotation within an angle uniformly sampled between $0$ and $\frac{\pi}{2}$, along with scaling by a factor uniformly distributed between $0.5$ and $1.5$. The second view consists of another image with the same digit identity with an additional background layer of Perlin noise \citep{Perlin1985}, with the noise factor uniformly distributed between $0$ and $1$. Both views are normalized to an intensity range of $[0,1)$, then flattened to form an array of $784$ dimensions.

To cast this dataset into our language, we shuffled the images within labels, retaining the shared label identity (that is the shared signal), but we still have the view-specific details (which is the self signal). This resulted in a total dataset size of $\sim 56k$ images for training and $\sim 7k$ images for testing.
The correlation histogram of $X$ (or $Y$) with itself shows a relatively wide spectrum when compared to the cross correlation between $X$ and $Y$, highlighting that the self signal is stronger, and can lead to different DR methods overseeing the shared one. The complexity of the tasks makes it sufficiently challenging, serving as a good benchmark for evaluating the performance of the different DR techniques.

\subsubsection{Results}
\begin{figure}[b!]
    \centering
    \includegraphics[width=0.95\textwidth]{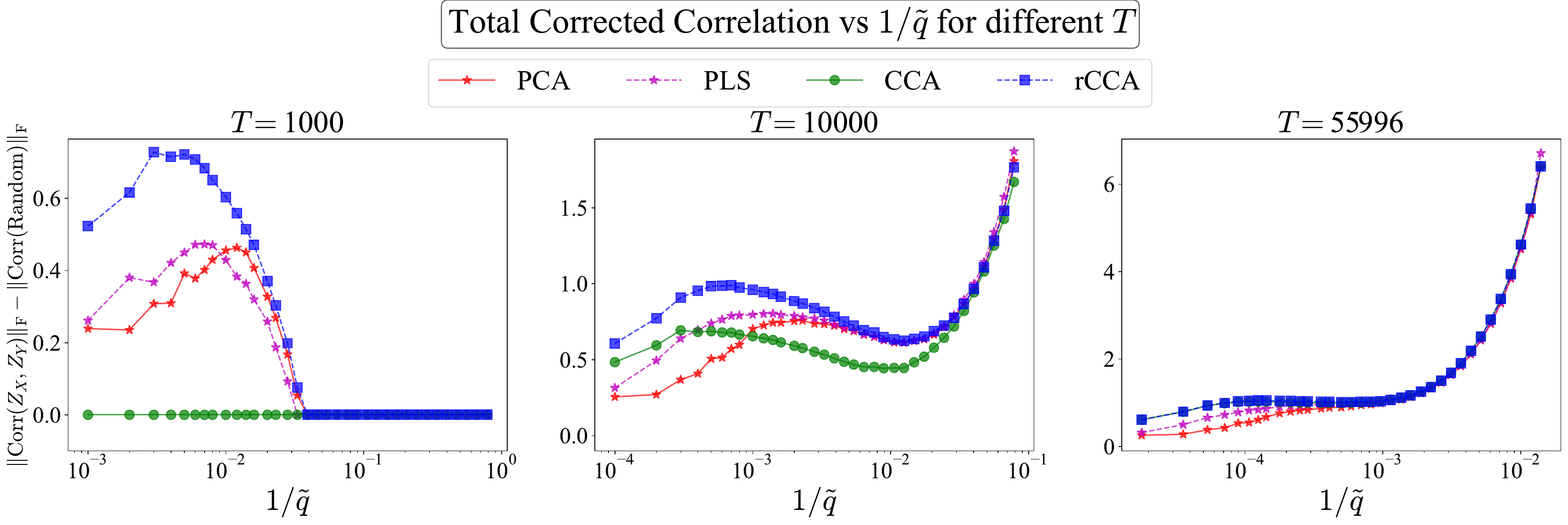}
    \caption{Performance of PCA, PLS, CCA, rCCA applied to the modified Noisy MNIST dataset across varying sampling scenarios. Each panel represents different sample sizes ($1000$, $10,000$, and approximately $56,000$ samples). The x-axis denotes the inverse of the number of samples per retained dimensions $(1/\tilde{q})$, while the y-axis represents the total corrected correlation between the obtained low-dimensional representations $Z_X$ and $Z_Y$.}
    \label{pd2-mnist}
\end{figure}

Figure \ref{pd2-mnist} shows the performance of PCA, PLS, CCA, and rCCA applied to the modified Noisy MNIST dataset for varying sampling scenarios. The three panels are evaluated for different sample sizes ($1000$, $10,000$, and $\sim 56,000$ samples), from undersampled to the full dataset.

In each scenario, the training samples are used for the DR methods. Subsequently, the learned projection matrices onto the singular directions are used to transform a separate test dataset of around $7,000$ samples into low-dimensional spaces, yielding $Z_X$ and $Z_Y$. The correlation between these transformed spaces is computed using the Frobenius norm of the correlation matrix. As before, we then subtracted from it the correlation value obtained from a random matrix of the same size. This difference is then plotted against $1/\tilde{q}$, which is the measure of how many dimensions are retained at each sampling ratio.

In the undersampled scenario ($1000$ samples), rCCA and PLS demonstrate an early detection (in terms of the number of kept dimensions after reduction) of shared signals, whereas PCA initially lags behind. As the number of dimensions increases, all methods exhibit a decline in correlation due to increased noise as we have fewer samples per dimension. CCA does not work in this scenario, since covariance matrices are degenerate.

Upon increasing the sample size ($10,000$ samples), a similar pattern emerges initially, where all methods experience an increase in total correlation till a certain number of kept dimensions is reached, then a decline when adding more dimensions. The decline is because one needs to estimate more singular vectors from the same number of samples. However, beyond a certain number of singular vectors, an increase in correlation is observed. This is because the number of vectors is now sufficient to learn both the shared and the self signals. We observe that rCCA maintains superior performance, while PCA reaches peak correlation at a higher number of kept dimensions, providing a rough estimation of the number of true self and shared signals.
With the full dataset (approximately 56,000 samples), a similar trend is seen. Yet CCA's performance approaches that of rCCA.

Notably, the consistent superiority of Simultaneous Dimensionality Reduction (SDR) over Independent Dimensionality Reduction (IDR) is reaffirmed, emphasizing its effectiveness in detecting shared signals even in nonlinear datasets.

\section{Discussions}

\subsection{Extensions and Generalizations}
We used a generative linear model which captures multiple desired features of multimodal data with shared and non-shared signals. The model focused only on data with two measured modalities. However, while not a part of this study,  the model can be readily extended to accommodate more than two modalities (e.~g., $X_i = R_i + U_i V_i + PQ_i$ for $i=1,...,n$, where $n$ represents the number of modalities). Then, methods such as Tensor CCA, which can handle more than two modalities \citep{Wen2015}, can be used to get insight into DR on such data.

\subsection{Explaining Observations in the Literature}\label{observations in the litrature}
We analyzed different DR methods on data from this model in different parameter regimes. Linear SDR methods were clearly superior to their IDR counterparts for detecting shared signals. We observed similar results on a nonlinear dataset as well. We thus make a strong practical suggestion that, whenever the goal is to reconstruct a low dimensional representation of covariation between two components of the data, IDR methods (PCA) should always be avoided in favor of SDR. Of the examined SDR approaches, rCCA is a clear winner in all parameter regimes and should always be preferred.  These findings explain the results of, for example, \cite{Shanechi2021} and others that SDR can recover joint neuro-behavioral latent spaces with fewer latent dimensions and using fewer samples than IDR methods. Further, our observation that SDR is always superior to IDR in the context of our model corroborates the theoretical findings of \cite{martini2024data}, who proved a similar result in the context of discrete data and a different SDR algorithm, namely the  Symmetric Information Bottleneck \citep{Tishby2013}. \cite{Maggioni2021} made similar conclusions using conditional covariance matrices for the reduction in the context of classification. More recent work of \cite{Abdelaleem2023} showed similar results using deep variational methods. Collectively, these diverse investigations, linear and nonlinear, theoretical, computational, and empirical, provide strong evidence that generic (not just linear) SDR methods are likely to be more efficient in extracting covariation than their IDR analogs.

\subsection{Is SDR strictly effective in low sampling situations?}
Our study answers an open question in the literature surrounding the effectiveness of SDR techniques. Specifically, there has been debate about whether PLS, an SDR method, is effective at low sampling \citep{Newsted1999, Sarstedt2011, Thompson2006, Thompson2012}. Our results show that SDR is not necessarily effective in the undersampled regime. It works well when the number of samples per retained dimension is high (even if the number of samples per observed dimension is low), but only when the dimensionality of the reduced description is matched to the actual dimensionality of the shared signals. 

\subsection{Diagnostic Test for number of latent signals}
In addition to the previous, our results can be used as a diagnostic test to determine the number of shared versus self signals in data. As demonstrated in Fig.~\ref{pd2-t1000}, total correlations between $Z_X$ and $Z_Y$ obtained by applying PCA and rCCA increase monotonically as the dimensionality of $Z$s increases, until this dimensionality becomes larger than the signal dimensionality. For PCA, the signal dimensionality is equal to the sum of the number of the shared and the self signals, $m_\text{shared} + m_\text{self}$. For rCCA, it is only the number of the shared signal. Thus increasing the dimensionality of the compressed variables and tracking the performance of rCCA and PCA until they diverge can be used to identify the number of self signals in the data, provided that the data, indeed,  has a low-dimensional latent structure. This approach can be a valuable tool in various applications, where the characterization of shared and self signals in complex systems can provide insights into their structure and function.

\subsection{Limitations, and Future Work}
\subsubsection{Linearity of the model}
While this work has provided useful insight,  the assumptions made here may not fully capture the complexity of real-world data. Specifically, our data is generated by a linear model with random Gaussian features. It is unlikely that real data have this exact structure. Therefore, there is a need for further exploration of the advantages and limitations of linear DR methods on data that have a low-dimensional, but nonlinear shared structure. 
This can be done using more complex nonlinear generative models, such as nonlinearly transforming the data generated by Eq.~(\ref{model}-\ref{model3}), or random feature two-layered neural network models \citep{Mehta2022}. Alternatively, analyzing the model, Eq.~(\ref{model}) using various theoretical techniques \citep{Knutsson1997, Maggioni2021,potters2020first} is likely to offer even more insights into its properties. Collectively, these diverse approaches would aid our understanding of different DR methods under diverse conditions.

\subsubsection{Linearity of the methods}
A different possible future research direction is to explore the performance of nonlinear DR methods on data from generative models with a latent low-dimensional nonlinear structure. Autoencoders and their variational extensions are a natural extension of IDR to learn nonlinear reduced dimensional representations \citep{Salakhutdinov2006, Welling2014, Lerchner2016}. Meanwhile, Deep CCA and its variational extensions \citep{Livescu2013, Bilmes2015, Ravindran2015, Livescu2016} should be explored as a nonlinear version of SDR. Both of these types of methods can potentially capture more complex relationships between the modalities and improve the quality of the reduced representations, and while recent work suggests that \citep{Abdelaleem2023}, it is not clear if the SDR class of methods is always more efficient than the IDR one.

\subsubsection{Linearity of the metric}
Our analysis also depends on the choice of metric used to quantify the performance of DR, and different choices should also be explored.  For example,  to capture nonlinear correlations, mutual information can be utilized to quantify the relationships between the reduced representations.

\subsection{Conclusion}

In conclusion, we highlight a general principle that, when searching for a shared signal between different modalities of data, SDR methods are preferable to IDR methods. Additionally, the differences in performance between the two classes of methods can tell us a lot about the underlying structure of the data. Finally, for a limited number of samples, naive approaches, such as increasing the number of compressed dimensions indefinitely to overcome the masking of shared signals by self signals are infeasible. Thus, the use of SDR methods becomes even more essential in such cases, and despite the aforementioned limitations, we believe that our work provides a compelling addition to the body of knowledge that  SDR outperforms IDR in detecting shared signals quite generally.

\section{Limitations, and Future Work}

While this work has provided useful insight,  the assumptions made here may not fully capture the complexity of real-world data. Specifically, our data is generated by a linear model with random Gaussian features. It is unlikely that real data have this exact structure. Therefore, there is a need for further exploration of the advantages and limitations of linear DR methods on data that have a low-dimensional, but nonlinear shared structure. 
This can be done using more complex nonlinear generative models, such as nonlinearly transforming the data generated by Eq.~(\ref{model}-\ref{model3}), or random feature two-layered neural network models \citep{Mehta2022}. 

A different possible future research direction is to explore the performance of nonlinear DR methods on data from generative models with a latent low-dimensional nonlinear structure. Autoencoders and their variational extensions are a natural extension of IDR to learn nonlinear reduced dimensional representations \citep{Salakhutdinov2006, Welling2014, Lerchner2016}. Meanwhile, Deep CCA and its variational extensions \citep{Livescu2013, Bilmes2015, Ravindran2015, Livescu2016} should be explored as a nonlinear version of SDR. Both of these types of methods can potentially capture more complex relationships between the modalities and improve the quality of the reduced representations, and it is not clear if the SDR class of methods is always more efficient than the IDR one.

Further, our analysis depends on the choice of metric used to quantify the performance of DR, and different choices should also be explored.  For example,  to capture nonlinear correlations, mutual information can be utilized to quantify the relationships between the reduced representations.

Despite the aforementioned limitations, we believe that our work provides a compelling addition to the body of knowledge that  SDR outperforms IDR in detecting shared signals quite generally.

\chapter{Deep Variational Multivariate Information Bottleneck Framework}\label{ch3}

\section{Summary}
\footnote{This chapter presents the paper \cite{Abdelaleem2023} with the title \textit{Deep Variational Multivariate Information Bottleneck -- A Framework for Variational Losses}. This work was conducted in collaboration with K.\ Michael Martini and Ilya Nemenman. Michael and I contributed equally to all the analytics, coding, result production, and manuscript writing. All authors contributed to conceiving the framework and reviewed the manuscript.}Variational dimensionality reduction methods are known for their high accuracy, generative abilities, and robustness. We introduce a framework to unify many existing variational methods and design new ones. The framework is based on an interpretation of the multivariate information bottleneck, in which an encoder graph, specifying what information to compress, is traded-off against a decoder graph, specifying a generative model. Using this framework, we rederive existing dimensionality reduction methods including the deep variational information bottleneck and variational auto-encoders. The framework naturally introduces a trade-off parameter extending the deep variational CCA (DVCCA) family of algorithms to beta-DVCCA. We derive a new method, the deep variational symmetric informational bottleneck (DVSIB), which simultaneously compresses two variables to preserve information between their compressed representations. We implement these algorithms and evaluate their ability to produce shared low dimensional latent spaces on Noisy MNIST dataset. We show that algorithms that are better matched to the structure of the data (in our case, beta-DVCCA and DVSIB) produce better latent spaces as measured by classification accuracy, dimensionality of the latent variables, and sample efficiency. We believe that this framework can be used to unify other multi-view representation learning algorithms and to derive and implement novel problem-specific loss functions.

\section{Introduction}

Large dimensional multi-modal datasets are abundant in multimedia systems utilized for language modeling \citep{Rui2016, Corso2018, Schiele2017, Metze2018, Parikh2017, Bowman2018, Steinhardt2020},  neural control of behavior studies \citep{Harris2021, Churchland2022, Poeppel2017, Fairhall2016}, multi-omics approaches in systems biology \citep{Clark2013, Bielas2017, Teichmann2018, O'Donovan2015, Lorenzi2018}, and many other domains. Such data come with the curse of dimensionality, making it hard to learn the relevant statistical correlations from samples. The problem is made even harder by the data often containing information that is irrelevant to the specific questions one asks. To tackle these challenges, a myriad of dimensionality reduction (DR) methods have emerged. By preserving certain aspects of the data while discarding the remainder, DR can decrease the complexity of the problem, yield clearer insights, and provide a foundation for more refined modeling approaches.

DR techniques span linear methods like Principal Component Analysis (PCA) \citep{Hotelling1933}, Partial Least Squares (PLS) \citep{Eriksson2001}, Canonical Correlations Analysis (CCA) \citep{Hotelling1936}, and regularized CCA \citep{Vinod1976, Strother1998}, as well as nonlinear approaches, including Autoencoders (AE) \citep{Salakhutdinov2006}, Deep CCA \citep{Livescu2013}, Deep Canonical Correlated AE \citep{Bilmes2015}, Correlational Neural Networks \citep{Ravindran2015}, Deep Generalized CCA \citep{Arora2017}, and Deep Tensor CCA \citep{Zeng2021}. Of particular interest to us are variational methods, such as Variational Autoencoders (VAE) \citep{Welling2014}, beta-VAE \citep{Lerchner2016}, Joint Multimodal VAE (JMVAE) \citep{Matsuo2016}, Deep Variational CCA (DVCCA) \citep{Livescu2016}, Deep Variational Information Bottleneck (DVIB) \citep{Murphy2017}, Variational Mixture-of-experts AE \citep{shi2019}, and Multiview Information Bottleneck \citep{federici2020}. These  DR methods use deep neural networks and variational approximations to learn robust and accurate representations of the data, while, at the same time, often serving as generative models for creating samples from the learned distributions.

There are many theoretical derivations and justifications for variational DR methods \citep{Welling2014, Lerchner2016, Matsuo2016, Livescu2016, Schuurmans2021, Lin2022, Murphy2017, Bao2021, VanderSchaar2021, Zhou2019, Hu2021, Akata2020, Elgamal2022, Ye2020}. This diversity of derivations, while enabling adaptability, often leaves researchers with no principled ways for choosing a method for a particular application, for designing new methods with distinct assumptions, or for comparing methods to each other.

Here, we introduce the Deep Variational Multivariate Information Bottleneck (DVMIB) framework, offering a unified mathematical foundation for many variational DR methods. Our framework is grounded in the multivariate information bottleneck loss function \citep{Bialek2000, Tishby2013}. This loss, amenable to approximation through upper and lower variational bounds, provides a system for implementing diverse DR variants using deep neural networks. We demonstrate the framework's efficacy by deriving the loss functions of many existing variational DR methods starting from the same principles. Furthermore, our framework naturally allows the adjustment of trade-off parameters, leading to generalizations of these existing methods. For instance, we generalize DVCCA to $\beta$-DVCCA. The framework further allows us to introduce and implement in software novel DR methods. We view the DVMIB framework, with its uniform information bottleneck language, conceptual clarity of translating statistical dependencies in data via graphical models of encoder and decoder structures into variational losses, the ability to unify existing approaches, and easy adaptability to new scenarios as one of the main contributions of our work.

Beyond its unifying role, our framework offers a principled approach for deriving problem-specific loss functions using domain-specific knowledge. Thus, we anticipate its application for multi-view representation learning across diverse fields. To illustrate this, we use the framework to derive a novel dimensionality reduction method,  the Deep Variational Symmetric Information Bottleneck (DVSIB), which compresses two random variables into two distinct latent variables that are maximally informative about one another. This new method produces better representations of classic datasets than previous approaches. The introduction of DVSIB is another major contribution of our work.

In summary, this chapter makes the following contributions to the field:
\begin{enumerate}
    \item \textbf{Introduction of the Variational Multivariate Information Bottleneck Framework:} We provide both intuitive and mathematical insights into this framework, establishing a robust foundation for further exploration.
    
    \item \textbf{Rederivation and Generalization of Existing Methods within a Common Framework:} We demonstrate the versatility of our framework by systematically rederiving and generalizing various existing methods from the literature, showcasing the framework's ability to unify diverse approaches.
    
    \item \textbf{Design of a Novel Method — Deep Variational Symmetric Information Bottleneck (DVSIB):} Employing our framework, we introduce DVSIB as a new method, contributing to the growing repertoire of techniques in variational dimensionality reduction. The method constructs high-accuracy latent spaces from substantially fewer samples than comparable approaches. 
\end{enumerate}

The chapter is structured as follows. First, we introduce the underlying mathematics and the implementation of the DVMIB framework. We then explain how to use the framework to generate new DR methods. In Tbl.~\ref{table:methods}, we present several known and newly derived variational methods, illustrating how easily they can be derived within the framework. As a proof of concept, we then benchmark {\em simple} computational implementations of methods in Tbl.~\ref{table:methods} against the Noisy MNIST dataset. Appendices present detailed treatment of all terms in variational losses in our framework, discussion of multi-view generalizations, and more details ---including visualizations--- of the performance of many methods on the Noisy MNIST.
\section{Multivariate Information Bottleneck Framework}

We represent DR problems similar to the Multivariate Information Bottleneck (MIB) of \citet{Tishby2013}, which is a generalization of the more traditional Information Bottleneck algorithm \citep{Bialek2000} to multiple variables. The reduced representation is achieved as a trade-off between two Bayesian networks. Bayesian networks are directed acyclic graphs that provide a factorization of the joint probability distribution, $P(X_1,X_2,X_3,..,X_N)=\prod_{i=1}^N P(X_i|Pa_{X_i}^G)$, where $Pa_{X_i}^G$ is the set of parents of $X_i$ in graph $G$.  The multiinformation \citep{Vejnarova1998} of a Bayesian network is defined as the Kullback-Leibler divergence between the joint probability distribution and the product of the marginals, and it serves as a measure of the total correlations among the variables, $I(X_1,X_2,X_3,...,X_N)=D_{KL}(P(X_1,X_2,X_3,...,X_N) \Vert P(X_1)P(X_2)P(X_3)...P(X_N))$. For a Bayesian network, the multiinformation reduces to the sum of all the local informations $I(X_1,X_2,..X_N)=\sum_{i=1}^N I(X_i;Pa_{X_i}^G)$ \citep{Tishby2013}. 

The first of the Bayesian networks is an encoder (compression) graph, which models how compressed (reduced, latent) variables are obtained from the observations. The second network is a decoder graph, which specifies a generative model for the data from the compressed variables, i.e., it is an alternate factorization of the distribution. In MIB, the information of the encoder graph is minimized, ensuring strong compression (corresponding to the approximate posterior).  The information of the decoder graph is maximized, promoting the most accurate model of the data (corresponding to maximizing the log-likelihood). As in  IB \citep{Bialek2000}, the trade-off between the compression and reconstruction is controlled by a trade-off parameter $\beta$:
\begin{equation}
L=I_{\text{encoder}}-\beta I_{\text{decoder}}.
\label{multiInfo}
\end{equation}

In this work, our key contribution is in writing an explicit variational loss for typical information terms found in both the encoder and the decoder graphs.  All terms in the decoder graph use samples of the compressed variables as determined from the encoder graph.  If there are two terms that correspond to the same information in Eq.~(\ref{multiInfo}), one from each of the graphs, they do not cancel each other since they correspond to two different variational expressions. For pedagogical clarity, we do this by first analyzing the Symmetric Information Bottleneck (SIB), a {\em special case} of MIB. We derive the bounds for three types of information terms in SIB, which we then use as building blocks for all other variational MIB methods in subsequent Sections.

\subsection{Deep Variational Symmetric Information Bottleneck}\label{sec:DVSIB}
The Deep Variational Symmetric Information Bottleneck (DVSIB)  simultaneously reduces a pair of datasets $X$ and $Y$ into two separate lower dimensional compressed versions $Z_X$ and $Z_Y$. These compressions are done at the same time to ensure that the latent spaces are maximally informative about each other. The joint compression is known to decrease dataset size requirements compared to individual ones \citep{martini2024data}.  Having distinct latent spaces for each modality usually helps with interpretability. For example, $X$ could be the neural activity of thousands of neurons, and $Y$ could be the recordings of joint angles of the animal. Rather than one latent space representing both, separate latent spaces for the neural activity and the joint angles are sought. By maximizing compression as well as $I(Z_X,Z_Y)$, one constructs the latent spaces that capture only the neural activity pertinent to joint movement and only the movement that is correlated with the neural activity (cf.~\cite{Fairhall2016}). Many other applications could benefit from a similar DR approach.

\begin{wrapfigure}{r}{0.4\textwidth}
\begin{center}
\vspace{-.2in}
\adjustbox{width=.4\textwidth}{
    \begin{tikzpicture}[node distance={15mm}, thick,
    main/.style = {draw, circle,minimum size=11mm},
    comp/.style = {draw, circle,minimum size=9mm},
    labs/.style = {}
    ] 
    \node[labs] (a) {$G_{\text{encoder}}$};
    \node[main] (x) [below=4mm of a] {$X$}; 
    \node[main] (y) [right of=x] {$Y$};
    \node[comp] (zx) [below of=x] {$Z_X$}; 
    \node[comp] (zy) [below of=y] {$Z_Y$};
    \draw (x) -- (y);
    \draw[->] (x) -- (zx);
    \draw[->] (y) -- (zy);

    \node[main] (mx) [right of=y] {$X$}; 
    \node[labs] (b) [above=4mm of mx] {$G_{\text{decoder}}$};
    \node[main] (my) [right of=mx] {$Y$};
    \node[comp] (mzx) [below of=mx] {$Z_X$}; 
    \node[comp] (mzy) [below of=my] {$Z_Y$};
    \draw[->] (mzx) -- (mzy);
    \draw[->] (mzx) -- (mx);
    \draw[->] (mzy) -- (my);
    \end{tikzpicture}
}
\end{center}
\caption{The encoder and decoder graphs for DVSIB.}
\label{Fig:Graph}
%\vspace{-.2in}
\end{wrapfigure}

In Fig.~\ref{Fig:Graph}, we define two Bayesian networks for DVSIB, $G_{\text{encoder}}$ and $G_{\text{decoder}}$. $G_{\text{encoder}}$ encodes the compression of $X$ to $Z_X$ and $Y$ to $Z_Y$. It corresponds to the factorization $p(x,y,z_x,z_y)=p(x,y)p(z_x|x)p(z_y|y)$ and the resultant $I_{\text{encoder}} = I^{E}(X;Y)+I^{E}(X;Z_X)+I^{E}(Y;Z_Y)$. The $I^{E}(X,Y)$ term does not depend on the compressed variables, does not affect the optimization problem, and hence is discarded in what follows. $G_{\text{decoder}}$ represents  a generative model for $X$ and $Y$ given the compressed latent variables $Z_X$ and $Z_Y$. It corresponds to the factorization $p(x,y,z_x,z_y)=p(z_x)p(z_y|z_x)p(x|z_x)p(y|z_y)$ and the resultant $I_{\text{decoder}} = I^{D}(Z_X;Z_Y)+I^{D}(X;Z_X)+I^{D}(Y;Z_Y)$.  Combing the informations from both graphs and  using Eq.~(\ref{multiInfo}), we find the SIB loss:
\begin{equation}
L_{\text{SIB}}=I^{E}(X;Z_X)+I^{E}(Y;Z_Y)
-\beta \left(I^{D}(Z_X;Z_Y)+I^{D}(X;Z_X)+I^{D}(Y;Z_Y)\right).
\label{eq:sib}
\end{equation}

Note that information in the encoder terms is minimized, and information in the decoder terms is maximized. Thus, while it is tempting to simplify Eq.~(\ref{eq:sib}) by canceling $I^{E}(X;Z_X)$ and $I^{D}(X;Z_X)$, this would be a mistake.  Indeed,  these terms come from different factorizations:  the encoder corresponds to learning $p(z_x|x)$, and the decoder to $p(x|z_x)$.

While the DVSIB loss may appear similar to previous models, such as MultiView Information Bottleneck (MVIB) \citep{federici2020} and Barlow Twins \citep{zbontar2021}, it is distinct both conceptually and in practice. For example, MVIB aims to generate latent variables that are as similar to each other as possible, sharing the same domain. DVSIB, however, endeavors to produce distinct latent representations, which could potentially have different units or dimensions, while maximizing mutual information between them. Barlow Twins architecture on the other hand appears to have two latent subspaces while in fact they are one latent subspace that is being optimized by a regular information bottleneck.

We now follow a procedure and notation similar to \citet{Murphy2017} and construct variational bounds on all $I^{E}$ and $I^{D}$ terms. Terms without leaf nodes, i.~e., $I^{D}(Z_X,Z_Y)$, require new approaches.

\subsection{Variational Bounds on DVSIB Encoder Terms}
\label{sec:encoder}
The information $I^{E}(Z_X;X)$  corresponds to compressing the random variable $X$ to $Z_X$. Since this is an encoder term, it needs to be minimized in Eq.~(\ref{eq:sib}). Thus, we seek a variational bound $I^{E}(Z_X;X)\le \Tilde{I}^{E}(Z_X;X)$, where $\Tilde{I}^{E}$ is the variational version of $I^{E}$, which can be implemented using a deep neural network. We find $\Tilde{I}^E$ by using the positivity of the Kullback–Leibler divergence. We make $r(z_x)$ be a variational approximation to $p(z_x)$. Then $D_{\rm KL} ( p(z_x) \Vert r(z_x) )\ge 0$, so that $-\int dz_x p(z_x) \ln(p(z_x))\le-\int dz_x p(z_x) \ln(r(z_x))$. Thus, $-\int dx dz_x p(z_x,x) \ln(p(z_x))\le-\int dx dz_x p(z_x,x) \ln(r(z_x))$. We then add $\int dx dz_x p(z_x,x) \ln(p(z_x|x))$ to both sides and find: 

\begin{align}
    I^{E}(Z_X;X) &= \int dx dz_x p(z_x,x) \ln\left(\frac{p(z_x|x)}{p(z_x)}\right) \nonumber\\
    &\le \int dx dz_x p(z_x,x) \ln\left(\frac{p(z_x|x)}{r(z_x)}\right) \equiv \Tilde{I}^{E}(Z_X;X).
\end{align}
We further simplify the variational loss by approximating $p(x)\approx\frac{1}{N}\sum_{i=1}^N \delta(x-x_i)$, so that:
\begin{equation}
\Tilde{I}^{E}(Z_X;X)\approx\frac{1}{N}\sum_{i=1}^N\int dz_x p(z_x|x_i) \ln\left(\frac{p(z_x|x_i)}{r(z_x)}\right)=\frac{1}{N}\sum_{i=1}^N D_{\rm KL}(p(z_x|x_i) \Vert r(z_x)).
\end{equation}

The term $I^{E}(Y;Z_Y)$ can be treated in an analogous manner, resulting in:
\begin{equation}
\Tilde{I}^{E}(Z_Y;Y) \approx \frac{1}{N}\sum_{i=1}^N D_{KL}(p(z_y|y_i) \Vert r(z_y)).
\end{equation}

\subsection{Variational Bounds on DVSIB Decoder Terms}
\label{sec:decoder}
%model terms:
The term $I^{D}(X;Z)$  corresponds to a decoder of $X$ from the compressed variable $Z_X$. It is maximized in Eq.~(\ref{eq:sib}). Thus, we seek its variational version $\Tilde{I}^{D}$, such that $I^{D}\ge\Tilde{I}^{D}$. Here, $q(x|z_x)$ will serve as a variational approximation to $p(x|z_x)$. We use the positivity of the Kullback-Leibler divergence, $D_{\rm KL}(p(x|z_x) \Vert q(x|z_x)) \ge 0$, to find $\int dx\, p(x|z_x)\ln(p(x|z_x))\ge\int dx\, p(x|z_x)\ln(q(x|z_x))$. This gives $\int dz_x dx\, p(x,z_x)\ln(p(x|z_x))\allowbreak\ge\int dz_x dx\, p(x,z_x)\ln(q(x|z_x))$. We add the entropy of $X$ to both sides to arrive at the variational bound:
\begin{align}
    I^{D}(X;Z_X)&=\int dz_x dx\, p(x,z_x)\ln\frac{p(x|z_x)}{p(x)}\nonumber \\
    &\ge \int dz_x dx p(x,z_x)\ln\frac{q(x|z_x)}{p(x)}\equiv\Tilde{I}^{D}(X;Z_X).
\end{align}

We further simplify $\Tilde{I}^{D}$ by replacing $p(x)$ by samples, $p(x)\approx\frac{1}{N}\sum_i^N\delta(x-x_i)$ and using the $p(z_x|x)$ that we learned previously from the encoder:
\begin{equation}
\Tilde{I}^{D}(X;Z_X)\approx H(X)+\frac{1}{N}\sum_{i=1}^N\int dz_x p(z_x|x_i)\ln(q(x_i|z_x)).
\end{equation}
Here $H(X)$ does not depend on $p(z_x|x)$ and, therefore, can be dropped from the loss. The variational version of $I^{D}(Y;Z_Y)$ is obtained  analogously:
\begin{equation}
\Tilde{I}^{D}(Y;Z_Y)\approx H(Y)+\frac{1}{N}\sum_{i=1}^N\int dz_x p(z_y|y_i)\ln(q(y_i|z_y)).
\end{equation}

%second type of model term:
\subsection{Variational Bounds on Decoder Terms not on a Leaf - MINE}
\label{sec:MINE}
The variational bound above cannot be applied to the information terms that do not contain leaves in $G_{\rm decoder}$. For SIB, this corresponds to the $I^{D}(Z_X,Z_Y)$ term. This information is maximized. To find a variational bound such that $I^{D}(Z_X,Z_Y)\ge\Tilde{I}^{D}(Z_X,Z_Y)$, we use the MINE mutual information estimator \citep{Hjelm2018}, which samples both $Z_X$ and $Z_Y$ from their respective variational encoders. Other mutual information estimators, such as $I_{\rm Info NCE}$ \citep{Tucker2019}, can be used as long as they are differentiable\footnote{Further details and discussions for different estimators are presented in Chapter~\ref{ch4}.}. Other estimators might be better suited for different problems, but for our current application, $I_{\rm MINE}$ was sufficient. We variationally approximate $p(z_x,z_y)$ as $p(z_x)p(z_y)e^{T(z_x,z_y)}/\mathcal{Z}_{\text{norm}}$, where $\mathcal{Z}_{\text{norm}}=\int dz_x dz_y p(z_x)p(z_y)e^{T(z_x,z_y)}$ is the normalization factor. Here $T(z_x,z_y)$ is parameterized by a neural network that takes in samples of the latent spaces $z_x$ and $z_y$ and returns a single number. We again  use the positivity of the Kullback-Leibler divergence, $D_{\rm KL}(p(z_x,z_y) \Vert p(z_x)p(z_y)e^{T(z_x,z_y)}/\mathcal{Z}_{\text{norm}}) \ge 0$, which implies $\int dz_x dz_y p(z_x,z_y)\ln(p(z_x,z_y))\ge\int dz_x dz_y p(z_x,z_y)\ln\frac{p(z_x)p(z_y)e^{T(z_x,z_y)}}{\mathcal{Z}_{\text{norm}}}$. Subtracting $\int dz_x dz_y p(z_x,z_y)\ln(p(z_x)p(z_y))$ from both sides, we find:
\begin{equation}
I^{D}(Z_X;Z_Y) \ge
\int dz_x dz_y p(z_x,z_y)\ln\frac{e^{T(z_x,z_y)}}{\mathcal{Z}_{\text{norm}}}\equiv \Tilde{I}_{\text{MINE}}^{D}(Z_X;Z_Y).
\end{equation}

\subsection{Parameterizing the Distributions and the Reparameterization Trick}
$H(X)$, $H(Y)$, and $I(X,Y)$ do not depend on $p(z_x|x)$ and $p(z_y|y)$ and are dropped from the loss. Further, we can use any ansatz for the variational distributions we introduced. We choose parametric probability distribution families and learn the nearest distribution in these families consistent with the data. We assume $p(z_x|x)$ is a normal distribution with mean $\mu_{Z_X}(x)$ and a diagonal variance $\Sigma_{Z_X}(x)$. We learn the mean and the log variance as neural networks. We also assume that $q(x|z_x)$ is normal with a mean $\mu_{X}(z_x)$ and a unit variance. In principle, we could also learn the variance for this distribution, but practically we did not find the need for that, and the approach works well as is. Finally, we assume that $r(z_x)$ is a standard normal distribution.  We use the reparameterization trick to produce samples of ${z_x}_{i,j}=z_{x_j}(x_i)=\mu(x_i)+\sqrt{\Sigma_{Z_X}(x_i)} \eta_j$  from $p(z_x|x_i)$, where $\eta_j$ is drawn from a standard normal distribution \citep{Welling2014}. We choose the same types of distributions for the corresponding $z_y$ terms. 

To sample from $p(z_x,z_y)$ we use  $p(z_x,z_y)=\int dx dy\, p(z_x,z_y,x,y)=\int dx dy\, p(z_x|x)p(z_y|y)\times p(x,y)\approx\frac{1}{N}\sum_{i=1}^N p(z_x|x_i)p(z_y|y_i)=\frac{1}{N M^2}\sum_{i=1}^N (\sum_{j=1}^M\delta(z_x-{z_x}_{i,j}))(\sum_{j=1}^M\delta(z_y-{z_y}_{i,j}))$, where ${z_x}_{i,j} \in p(z_x|x_i)$ and ${z_y}_{i,j} \in p(z_y|y_i)$, and $M$ is the number of new samples being generated. To sample from $p(z_x)p(z_y)$, we generate samples from $p(z_x,z_y)$ and scramble the generated entries $z_x$ and $z_y$, destroying all correlations.
With this, the components of the loss function become
%=\frac{1}{N}\sum_{i=1}^N D_{\rm KL}(p(z_x|x_i) \Vert r(z_x))\nonumber\\
\begin{align}
\Tilde{I}^{E}(X;Z_X) &\approx \frac{1}{2N}\sum_{i=1}^N \left[\text{Tr}({\Sigma_{Z_X}(x_i)}) +||\vec{\mu}_{Z_X}(x_i)||^2-k_{Z_X}-\ln \det(\Sigma_{Z_X}(x_i)) \right],\label{IExzx}\\
\Tilde{I}^{D}(X;Z_X)&\approx
%\frac{1}{N}\sum_{i=1}^N\int dz_x p(z_x|x_i)\ln  q(x_i|z_x) = 
\frac{1}{MN}\sum_{i,j=1}^{N,M} {-\frac{1}{2}}||(x_i - \mu_{X}({z_x}_{i,j}))||^2,\label{IDxzx}\\
\Tilde{I}^{D}_{\rm MINE}(Z_X;Z_Y)&
%=
%\int dz_x dz_y p(z_x,z_y)\ln \frac{e^{T(z_x,z_y)}}%{\mathcal{Z}_{\text{norm}}}
\approx
\frac{1}{M^2N}\sum_{i,j_x,j_y=1}^{N,M,M}\left[T({z_x}_{i,j_x},{z_y}_{i,j_y}) - \ln \mathcal{Z}_{\text{norm}}\right],\label{Imine}
\end{align}
where $\mathcal{Z}_{\text{norm}}=\mathbb{E}_{z_x \sim p(z_x), z_y \sim p(z_y)}[e^{T(z_x,z_y)}]$, $k_{Z_X}$ is the dimension of $Z_X$, and the corresponding terms for $Y$ are similar. Combining these terms results in the variational loss for DVSIB:
\begin{equation}
L_{\text{DVSIB}}=\Tilde{I}^{E}(X;Z_X)+\Tilde{I}^{E}(Y;Z_Y)
-\beta \left(\Tilde{I}^{D}_{\rm MINE}(Z_X;Z_Y)+\Tilde{I}^{D}(X;Z_X)+\Tilde{I}^{D}(Y;Z_Y)\right).
\end{equation}

\begin{table}
    \vspace{-.5in}
    \caption{Method descriptions, variational losses, and the Bayesian Network graphs for each DR method derived in our framework. See Appendix \ref{App:Library} for details. For methods where we can reduce either $X$ or $Y$, only $X$ graphs/loss are shown}
    \label{table:methods}
    \vspace{-0.2in}
    \begin{center}
    \begin{tabular}{|m{0.64\textwidth}|m{0.13\textwidth}|m{0.13\textwidth}|}
        \hline
        \textbf{Method Description}
        & \textbf{$G_{\text{encoder}}$} & \textbf{$G_{\text{decoder}}$} \\
        \hline
        \textbf{beta-VAE} \citep{Welling2014, Lerchner2016}: Two independent Variational Autoencoder (VAE) models trained, one for each view, $X$ and $Y$.\newline
        $\sloppy L_{\text{VAE}}= \Tilde{I}^{E}(X;Z_X)-\beta \Tilde{I}^{D}(X;Z_X)$
        &
        \centering
        \adjustbox{height=18mm}{
            \begin{tikzpicture}[node distance={15mm}, thick,
                main/.style = {draw, circle,minimum size=11mm},
                comp/.style = {draw, circle,minimum size=9mm},
                labs/.style = {}
            ] 
            \node[main] (1) {$X$}; 
            \node[comp] (2) [below of=1] {$Z_X$};
            \draw[->] (1) -- (2);
            
            \end{tikzpicture}
        }
        &
        \adjustbox{height=18mm}{
            \begin{tikzpicture}[node distance={15mm}, thick,
                main/.style = {draw, circle,minimum size=11mm},
                comp/.style = {draw, circle,minimum size=9mm},
                labs/.style = {}
            ]

            \node[main] (3) {$X$};
            \node[labs] (a) [left=5mm of 3] {};%for aligning
            \node[comp] (4) [below of=3] {$Z_X$};
            \draw[->] (4) -- (3);
            \end{tikzpicture}
        }
        \\
        \hline
        \textbf{DVIB} \citep{Murphy2017}: Two bottleneck models trained, one for each view, $X$ and $Y$, using the other view as the supervising signal.\newline
        $\sloppy L_{\text{DVIB}}=\Tilde{I}^{E}(X;Z_X)-\beta \Tilde{I}^{D}(Y;Z_X)$
        &
        \adjustbox{height=18mm}{
        \begin{tikzpicture}[node distance={15mm}, thick,
            main/.style = {draw, circle,minimum size=11mm},
            comp/.style = {draw, circle,minimum size=9mm},
            labs/.style = {}
            ] 
            \node[main] (x) {$X$}; 
            \node[main] (y) [right of=x] {$Y$};
            \node[comp] (zx) [below of=x] {$Z_X$}; 
            \draw (x) -- (y);
            \draw[->] (x) -- (zx);
            
            \end{tikzpicture}
        }
        &
        \adjustbox{height=18mm}{
            \begin{tikzpicture}[node distance={15mm}, thick,
                main/.style = {draw, circle,minimum size=11mm},
                comp/.style = {draw, circle,minimum size=9mm},
                labs/.style = {}
            ]
            
            \node[main] (mx) {$X$}; 
            \node[main] (my) [right of=mx] {$Y$};
            \node[comp] (mzx) [below of=mx] {$Z_X$}; 
            \draw[->] (mzx) -- (my);
            \end{tikzpicture}    
        } \\
        \hline
        \textbf{beta-DVCCA}: Similar to DVIB \citep{Murphy2017}, but with reconstruction of both views. Two models trained, compressing either $X$ or $Y$,  while reconstructing both $X$ and $Y$.
        \newline
        $\sloppy L_{\text{DVCCA}}=\Tilde{I}^{E}(X;Z_X)-\beta (\Tilde{I}^{D}(Y;Z_X)+\Tilde{I}^{D}(X;Z_X))$
        \newline
        \textbf{DVCCA} \citep{Livescu2016}: $\beta$-DVCCA with $\beta=1$.
        &
        \adjustbox{height=18mm}{
        \begin{tikzpicture}[node distance={15mm}, thick,
            main/.style = {draw, circle,minimum size=11mm},
            comp/.style = {draw, circle,minimum size=9mm},
            labs/.style = {}
            ] 
            \node[main] (x) {$X$}; 
            \node[main] (y) [right of=x] {$Y$};
            \node[comp] (zx) [below of=x] {$Z_X$}; 
            \draw (x) -- (y);
            \draw[->] (x) -- (zx);
                        
            \end{tikzpicture}
        }
        &
        \adjustbox{height=18mm}{
            \begin{tikzpicture}[node distance={15mm}, thick,
                main/.style = {draw, circle,minimum size=11mm},
                comp/.style = {draw, circle,minimum size=9mm},
                labs/.style = {}
            ] 
            
            \node[main] (mx) {$X$}; 
            \node[main] (my) [right of=mx] {$Y$};
            \node[comp] (mzx) [below of=mx] {$Z_X$}; 
            \draw[->] (mzx) -- (my);
            \draw[->] (mzx) -- (mx);
            \end{tikzpicture}    
        } \\
        \hline
        \textbf{beta-joint-DVCCA}: A single model trained using a concatenated variable $[X,Y]$, learning one latent representation $Z$. \newline
        $\sloppy L_{\text{jDVCCA}}=\Tilde{I}^{E}((X,Y);Z)-\beta (\Tilde{I}^{D}(Y;Z)+\Tilde{I}^{D}(X;Z))$
        \newline
        \textbf{joint-DVCCA} \citep{Livescu2016}: $\beta$-jDVCCA with $\beta=1$.
        &
        \adjustbox{height=12mm}{
                    \begin{tikzpicture}[node distance={15mm}, thick,
            main/.style = {draw, circle,minimum size=11mm},
            comp/.style = {draw, circle,minimum size=9mm},
            labs/.style = {}
            ] 
            \node[main] (1) {$X$};
            \node[comp] (3) [below right of=1] {$Z$};
            \node[main] (2) [above right of=3] {$Y$};
            \draw (1) -- (2);
            \draw[->] (1) -- (3);
            \draw[->] (2) -- (3);
                        
            \end{tikzpicture}
        }
        &
        \adjustbox{height=12mm}{
            \begin{tikzpicture}[node distance={15mm}, thick,
                main/.style = {draw, circle,minimum size=11mm},
                comp/.style = {draw, circle,minimum size=9mm},
                labs/.style = {}
            ]

            \node[main] (x) {$X$};
            \node[comp] (z) [below right of=x] {$Z$};
            \node[main] (y) [above right of=z] {$Y$};
            \draw[->] (z) -- (x);
            \draw[->] (z) -- (y);
            \end{tikzpicture}
        } \\
        \hline
        \textbf{beta-DVCCA-private}: Two models trained, compressing either $X$ or $Y$, while reconstructing both $X$ and $Y$, and simultaneously learning private information $W_X$ and $W_Y$
        
        $\sloppy L_{\text{DVCCA-p}}=\Tilde{I}^{E}(X;Z)+\Tilde{I}^{E}(X;W_X)+\Tilde{I}^{E}(Y;W_Y)-\beta (\Tilde{I}^{D}(X;(W_X,Z))+\Tilde{I}^{D}(Y;(W_Y,Z)))$

        \textbf{DVCCA-private} \citep{Livescu2016}: $\beta$-DVCCA-p with $\beta=1$.
        &
        \adjustbox{height=15mm}{
        \begin{tikzpicture}[node distance={15mm}, thick,
            main/.style = {draw, circle,minimum size=11mm},
            comp/.style = {draw, circle,minimum size=9mm},
            labs/.style = {}
            ]
            \node[main] (1) {$X$};
            \node[comp] (3) [below right of=1] {$Z$};
            \node[main] (2) [above right of=3] {$Y$};
            \node[comp] (4) [below of=1] {$W_X$};
            \node[comp] (5) [below of=2] {$W_Y$};
            \draw (1) -- (2);
            \draw[->] (1) -- (3);
            \draw[->] (1) -- (4);
            \draw[->] (2) -- (5);
        \end{tikzpicture}
        }
        &
        \adjustbox{height=15mm}{
        \begin{tikzpicture}[node distance={15mm}, thick,
            main/.style = {draw, circle,minimum size=11mm},
            comp/.style = {draw, circle,minimum size=9mm},
            labs/.style = {}
        ] 
            \node[main] (x) {$X$};
            \node[comp] (z) [below right of=x] {$Z$};
            \node[main] (y) [above right of=z] {$Y$};
            \node[comp] (wx) [below of=x] {$W_X$};
            \node[comp] (wy) [below of=y] {$W_Y$};
            \draw[->] (wx) -- (x);
            \draw[->] (z) -- (x);
            \draw[->] (wy) -- (y);
            \draw[->] (z) -- (y);
        \end{tikzpicture}
        
        }\\
        \hline
        \textbf{beta-joint-DVCCA-private}: A single model trained using a concatenated variable $[X,Y]$, learning one latent representation $Z$, and simultaneously learning private information $W_X$ and $W_Y$.
        $\sloppy L_{\text{jDVCCA-p}}=\Tilde{I}^{E}((X,Y);Z)+\Tilde{I}^{E}(X;W_X)+$$\sloppy\Tilde{I}^{E}(Y;W_Y)-\beta (\Tilde{I}^{D}(X;(W_X,Z))+\Tilde{I}^{D}(Y;(W_Y,Z)))$ 

        \textbf{joint-DVCCA-private}\citep{Livescu2016}: $\beta$-jDVCCA-p with $\beta=1$.
        &
        \adjustbox{height=15mm}{
        \begin{tikzpicture}[node distance={15mm}, thick,
            main/.style = {draw, circle,minimum size=11mm},
            comp/.style = {draw, circle,minimum size=9mm},
            labs/.style = {}
            ] 
            \node[main] (1) {$X$};
            \node[comp] (3) [below right of=1] {$Z$};
            \node[main] (2) [above right of=3] {$Y$};
            \node[comp] (4) [below of=1] {$W_X$};
            \node[comp] (5) [below of=2] {$W_Y$};
            \draw (1) -- (2);
            \draw[->] (1) -- (3);
            \draw[->] (2) -- (3);
            \draw[->] (1) -- (4);
            \draw[->] (2) -- (5);
        \end{tikzpicture}
        }
        &
        \adjustbox{height=15mm}{
        \begin{tikzpicture}[node distance={15mm}, thick,
            main/.style = {draw, circle,minimum size=11mm},
            comp/.style = {draw, circle,minimum size=9mm},
            labs/.style = {}
        ] 
            \node[main] (x) {$X$};
            \node[comp] (z) [below right of=x] {$Z$};
            \node[main] (y) [above right of=z] {$Y$};
            \node[comp] (wx) [below of=x] {$W_X$};
            \node[comp] (wy) [below of=y] {$W_Y$};
            \draw[->] (wx) -- (x);
            \draw[->] (z) -- (x);
            \draw[->] (wy) -- (y);
            \draw[->] (z) -- (y);
        \end{tikzpicture}
        
        }\\
        \hline

        \textbf{DVSIB}: A symmetric model trained, producing $Z_X$ and $Z_Y$.\newline
        $\sloppy L_{\text{DVSIB}}=\Tilde{I}^{E}(X;Z_X)+\Tilde{I}^{E}(Y;Z_Y) $\newline $- \beta \left(\Tilde{I}^{D}_{\text{MINE}}(Z_X;Z_Y)+\Tilde{I}^{D}(X;Z_X)+\Tilde{I}^{D}(Y;Z_Y)\right)$
    &
    \adjustbox{height=18mm}{
    \begin{tikzpicture}[node distance={15mm}, thick,
        main/.style = {draw, circle,minimum size=11mm},
        comp/.style = {draw, circle,minimum size=9mm},
        labs/.style = {}
        ] 
        \node[main] (x) {$X$}; 
        \node[main] (y) [right of=x] {$Y$};
        \node[comp] (zx) [below of=x] {$Z_X$}; 
        \node[comp] (zy) [below of=y] {$Z_Y$};
        \draw (x) -- (y);
        \draw[->] (x) -- (zx);
        \draw[->] (y) -- (zy);
    \end{tikzpicture}
    }
    &
    \adjustbox{height=18mm}{
    \begin{tikzpicture}[node distance={15mm}, thick,
        main/.style = {draw, circle,minimum size=11mm},
        comp/.style = {draw, circle,minimum size=9mm},
        labs/.style = {}
        ] 
        \node[main] (mx) {$X$}; 
        \node[main] (my) [right of=mx] {$Y$};
        \node[comp] (mzx) [below of=mx] {$Z_X$}; 
        \node[comp] (mzy) [below of=my] {$Z_Y$};
        \draw[->] (mzx) -- (mzy);
        \draw[->] (mzx) -- (mx);
        \draw[->] (mzy) -- (my);
    \end{tikzpicture}
    }\\
    \hline
    \textbf{DVSIB-private}: A symmetric model trained, producing $Z_X$ and $Z_Y$, while simultaneously learning private information $W_X$ and $W_Y$.\newline
    $\sloppy L_{\text{DVSIBp}}=\Tilde{I}^{E}(X;W_X)+\Tilde{I}^{E}(X;Z_X) +$\newline$\Tilde{I}^{E}(Y;Z_Y)+\Tilde{I}^{E}(Y;W_Y)
-$\newline
$\sloppy\beta \left(\Tilde{I}^{D}_{\text{MINE}}(Z_X;Z_Y)+\Tilde{I}^{D}(X;(Z_X,W_X))+\Tilde{I}^{D}(Y;(Z_Y,W_Y))\right)$
    &
    \adjustbox{height=11mm}{
    \begin{tikzpicture}[node distance={15mm}, thick,
        main/.style = {draw, circle,minimum size=11mm},
        comp/.style = {draw, circle,minimum size=9mm},
        labs/.style = {}
        ] 
        \node[main] (x) {$X$}; 
        \node[main] (y) [right of=x] {$Y$};
        \node[comp] (zx) [below of=x] {$Z_X$}; 
        \node[comp] (zy) [below of=y] {$Z_Y$};
        \node[comp] (wx) [below left of=x] {$W_X$}; 
        \node[comp] (wy) [below right of=y] {$W_Y$};
        \draw (x) -- (y);
        \draw[->] (x) -- (zx);
        \draw[->] (y) -- (zy);
        \draw[->] (x) -- (wx);
        \draw[->] (y) -- (wy);
    \end{tikzpicture}
    }
    &
    \adjustbox{height=11mm}{
    \begin{tikzpicture}[node distance={15mm}, thick,
        main/.style = {draw, circle,minimum size=11mm},
        comp/.style = {draw, circle,minimum size=9mm},
        labs/.style = {}
        ] 
        \node[comp] (mwx) {$W_X$}; 
        \node[main] (mx) [above right of=mwx] {$X$}; 
        \node[main] (my) [right of=mx] {$Y$};
        \node[comp] (mzx) [below of=mx] {$Z_X$}; 
        \node[comp] (mzy) [below of=my] {$Z_Y$};
        \node[comp] (mwy) [below right of=my] {$W_Y$};
        \draw[->] (mzx) -- (mzy);
        \draw[->] (mzx) -- (mx);
        \draw[->] (mzy) -- (my);
        \draw[->] (mwx) -- (mx);
        \draw[->] (mwy) -- (my);
        
    \end{tikzpicture}
    }\\
    \hline
    
    \end{tabular}
    \end{center}
\end{table}

\section{Deriving Other DR Methods}
The variational bounds used in DVSIB can be used to implement loss functions that correspond to other encoder-decoder graph pairs and hence to other DR algorithms. The simplest is the beta variational auto-encoder. Here $G_{\rm encoder}$  consists of one term: $X$ compressed into $Z_X$. Similarly $G_{\rm decoder}$  consists of one term: $X$ decoded from $Z_X$ (see Table~\ref{table:methods}). Using this simple set of Bayesian networks, we find the variational loss:
\begin{equation}
L_{\text{beta-VAE}}=\Tilde{I}^{E}(X;Z_X)-\beta\Tilde{I}^{D}(X;Z_X).
\label{beta-VAE}
\end{equation}

Both terms in Eq.~(\ref{beta-VAE}) are the same as Eqs.~(\ref{IExzx}, \ref{IDxzx}) and can be approximated and implemented by neural networks. 

Similarly, we can re-derive the DVCCA family of losses \citep{Livescu2016}.
 Here $G_{\rm encoder}$  is $X$ compressed into $Z_X$. $G_{\rm decoder}$ reconstructs both $X$ and $Y$ from the same compressed latent space $Z_X$. In fact, our loss function is more general than the DVCCA loss and has an additional compression-reconstruction trade-off parameter $\beta$. We call this more general loss $\beta$-DVCCA, and the original  DVCCA emerges when $\beta=1$:
\begin{equation}
L_{\rm DVCCA}=\Tilde{I}^{E}(X;Z_X)-\beta (\Tilde{I}^{D}(Y;Z_X)+\Tilde{I}^{D}(X;Z_X)).
\end{equation}
Using the same library of terms as we found in DVSIB, Eqs.~(\ref{IExzx}, \ref{IDxzx}), we find:
\begin{multline}
L_{\text{DVCCA}}\approx\frac{1}{N}\sum_{i=1}^N D_{\rm KL}(p(z_x|x_i) \Vert r(z_x))\\
-\beta\left( \frac{1}{N}\sum_{i=1}^N\int dz_x p(z_x|x_i)\ln(q(y_i|z_x)) + \frac{1}{N}\sum_{i=1}^N\int dz_x p(z_x|x_i)\ln(q(x_i|z_x))\right).
\end{multline}
This is similar to the loss function of the deep variational CCA \citep{Livescu2016}, but now it has a trade-off parameter $\beta$. It trades off the compression into $Z_X$ against the reconstruction of $X$ and $Y$ from the compressed variable $Z_X$. 

Table~\ref{table:methods} shows how our framework reproduces and generalizes other DR losses (see Appendix \ref{App:Library}). Our framework naturally extends beyond two variables as well (see Appendix \ref{App:MultiViewLosses}).

\section{Results}
\begin{figure}[ht]
%\vspace{-0.5in}
\begin{center}
\includegraphics[width=.85\textwidth]{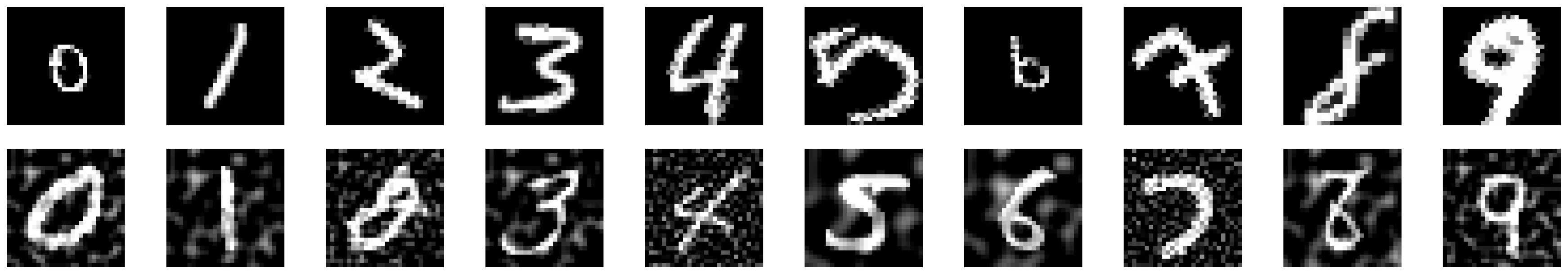}
\includegraphics[width=.32\textwidth, trim={0in 0in 0in 0.25in}, clip]{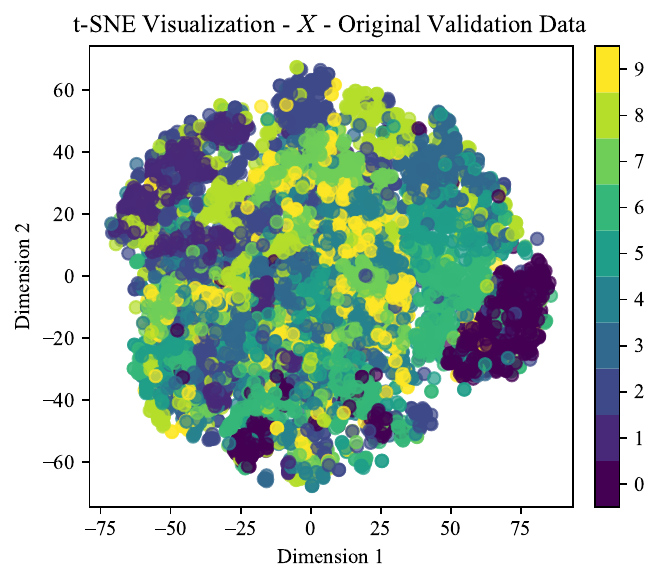}
\includegraphics[width=.32\textwidth, trim={0in 0in 0in 0.25in}, clip]{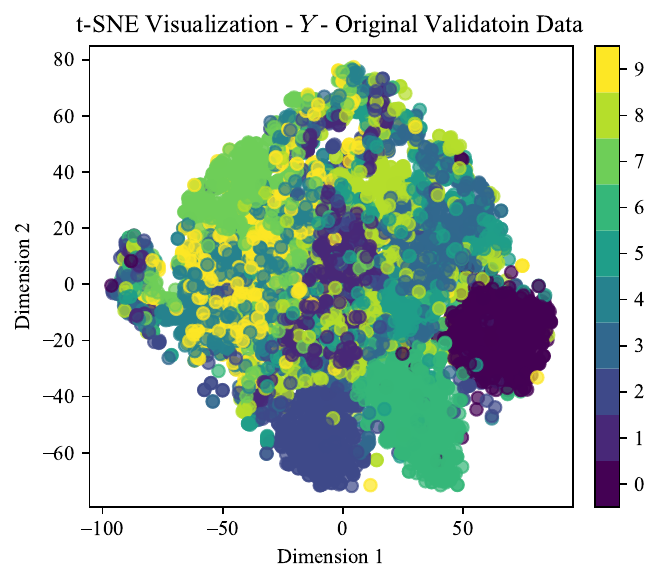}
\includegraphics[width=.32\textwidth]{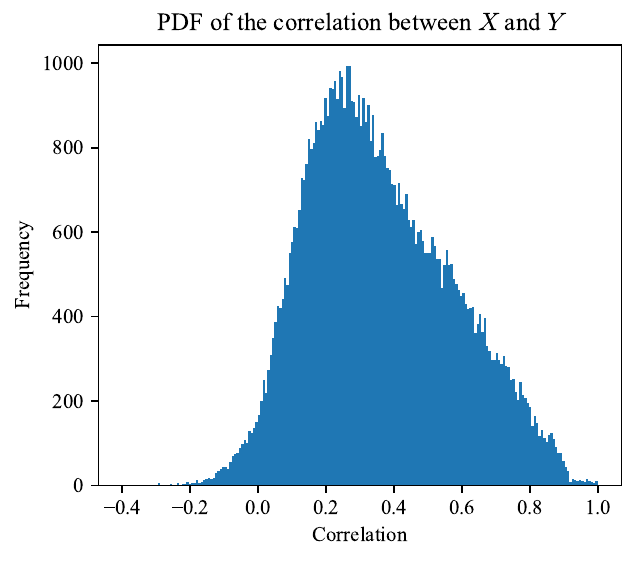}
\end{center}
% \vspace{-.25in}
\caption{Dataset consisting of pairs of digits drawn from MNIST that share an identity. Top row, $X$:  MNIST digits randomly scaled $(0.5-1.5)$ and rotated $(0-\pi/2)$. Bottom row, $Y$: MNIST digits with a background Perlin noise. t-SNE of $X$ and $Y$ datasets (left and middle) shows poor separation by digit, and there is a wide range of correlation between $X$ and $Y$ (right).}
\label{Fig:data}
%\vspace{-.15in}
\end{figure}

To test our methods, we created a dataset inspired by the noisy MNIST dataset \citep{Haffner1998, Bilmes2015, Livescu2016}, consisting of two distinct views of data, both with dimensions of $28 \times 28$~pixels, cf.~Fig.~\ref{Fig:data}. The first view comprises the original image randomly rotated by an angle uniformly sampled between $0$ and $\frac{\pi}{2}$ and scaled by a factor uniformly distributed between $0.5$ and $1.5$. The second view consists of the original image with an added background Perlin noise \citep{Perlin1985} with the noise factor uniformly distributed between $0$ and $1$. Both image intensities are scaled to the range of $[0,1)$. The dataset was shuffled within labels, retaining only the shared label identity between two images, while disregarding the view-specific details, i.e., the random rotation and scaling for $X$, and the correlated background noise for $Y$. The dataset, totaling $70,000$ images, was partitioned into training ($80\%$), testing ($10\%$), and validation ($10\%$) subsets. Visualization via t-SNE \citep{Roweis2002} plots of the original dataset suggest poor separation by digit, and the two digit views have diverse correlations,  making this a sufficiently hard problem. 

The DR methods we evaluated include all methods from  Tbl.~\ref{table:methods}. PCA and CCA \citep{Hotelling1933, Hotelling1936} served as a baseline for linear dimensionality reduction. Multi-view Information Bottleneck \cite{federici2020} was included for a specific comparison with DVSIB (see Appendix~\ref{App:MVIB}). We emphasize that none of the algorithms were given labeled data. They had to infer compressed latent representations that presumably should cluster into ten different digits based simply on the fact that images come in pairs, and the (unknown) digit label is the only information that relates the two images. 

Each method was trained for 100 epochs using fully connected neural networks with layer sizes $(\text{input\_dim}, 1024, 1024, (k_Z, k_Z))$, where $k_Z$ is the latent dimension size, employing ReLU activations for the hidden layers. The input dimension $(\text{input\_dim})$ was either the size of $X$ (784) or the size of the concatenated $[X, Y]$ (1568). The last two layers of size $k_Z$ represented the means and $\log(\text{variance})$ learned. For the decoders, we employed regular decoders, fully connected neural networks with layer sizes $(k_Z, 1024, 1024, \text{output\_dim})$, using ReLU activations for the hidden layers and sigmoid activation for the output layer. Again, the output dimension $(\text{output\_dim})$ could either be the size of $X$ (784) or the size of the concatenated $[X, Y]$ (1568). The latent dimension $(k_{Z})$ could be $k_{Z_X}$ or $k_{Z_Y}$ for regular decoders, or $k_{Z_X}+k_{W_X}$ or $k_{Z_Y}+k_{W_Y}$ for decoders with private information. Additionally, another decoder denoted as $\text{decoder\_MINE}$, based on the MINE estimator for estimating $I(Z_X, Z_Y)$, was used in DVSIB and DVSIB with private information. The $\text{decoder\_MINE}$ is a fully connected neural network with layer sizes $(k_{Z_X}+k_{Z_Y}, 1024, 1024, 1)$ and ReLU activations for the hidden layers. Optimization was conducted using the ADAM optimizer with default parameters.

\begin{table}[ht]
%\vspace{-.15 in}
\caption{Maximum accuracy from a linear SVM and the optimal $k_Z$ and $\beta$ for variational DR methods reported on the $Y$ (above the line) and the joint $[X,Y]$ (below the line) datasets. ($^\dag$ fixed values)}
\label{table:SVM-Y}
\begin{center}
\begin{tabular}{|l|c|c|c|c|c|c|}
\hline
\textbf{Method} & \textbf{Acc. \%} & \bm{${k_Z}_{\textbf{best}}$} & $95\%$ \bm{${k_Z}_{\text{range}}$} & \bm{$\beta_\text{best}$} & $95\%$ \textbf{$\beta_\text{range}$} & $\bm{C_\text{best}}$ \\
\hline
Baseline & 90.8 & 784$^\dag$ & - & - & - & 0.1 \\
%Baseline-stacked & 70.4* & 784* & - & 0.1 \\
PCA & 90.5 & 256 & [64,256*] & - & - & 1 \\
CCA & 85.7 & 256 & [32,256*] & - & - & 10 \\
$\beta$-VAE & 96.3 & 256 & [64,256*] & 32 & [2,1024*] & 10 \\
DVIB & 90.4 & 256 & [16,256*] & 512 & [8,1024*] & 0.003 \\
DVCCA & 89.6 & 128 & [16,256*] & 1$^\dag$ & - & 31.623 \\
$\beta$-DVCCA & 95.4 & 256 & [64,256*] & 16 & [2,1024*] & 10 \\
DVCCA-p & 92.1 & 16 & [16,256*] & 1$^\dag$ & - & 0.316 \\
$\beta$-DVCCA-p & 95.5 & 16 & [\textbf{4},256*] & 1024 & [1,1024*] & 0.316 \\
MVIB & \textbf{97.7} & 8 & [\textbf{4},64] & 1024 & [128,1024*] & 0.01 \\
DVSIB & \textbf{97.8} & 256 & [\textbf{8},256*] & 128 & [2,1024*] & 3.162 \\
DVSIB-p & \textbf{97.8} & 256 & [\textbf{8},256*] & 32 & [2,1024*] & 10 \\
\hline
jBaseline & 91.9 & 1568$^\dag$ & - & - & - & 0.003 \\
%Baseline-stacked & 70.4* & 784* & - & 0.1 \\
jDVCCA & 92.5 & 256 & [64,265*] & 1$^\dag$ & - & 10 \\
$\beta$-jDVCCA & 96.7 & 256 & [16,265*] & 256 & [1,1024*] & 1 \\
jDVCCA-p & 92.5 & 64 & [32,265*] & 1$^\dag$ & - & 10 \\
$\beta$-jDVCCA-p & 92.7 & 256 & [\textbf{4},265*] & 2 & [1,1024*] & 10 \\
\hline

\end{tabular}
% \vspace{-.3in}
\end{center}
\end{table}

\begin{figure}[t]
%\vspace{-0.5in}
\begin{center}
\includegraphics[width=.48\textwidth]{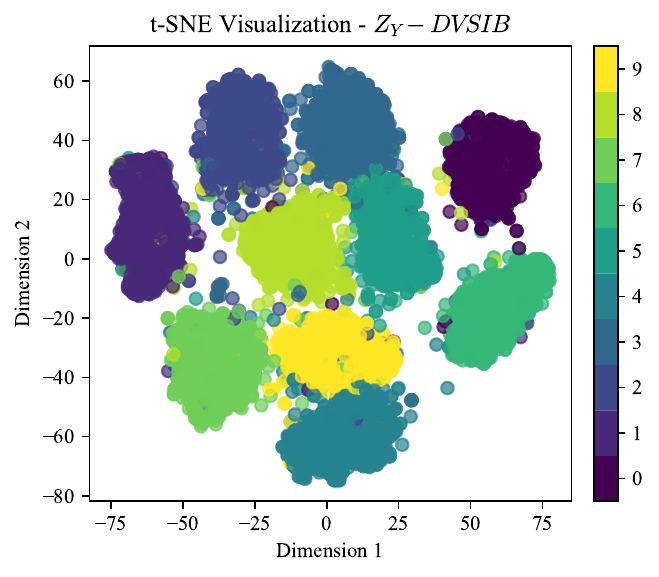}
\includegraphics[width=.5\textwidth]{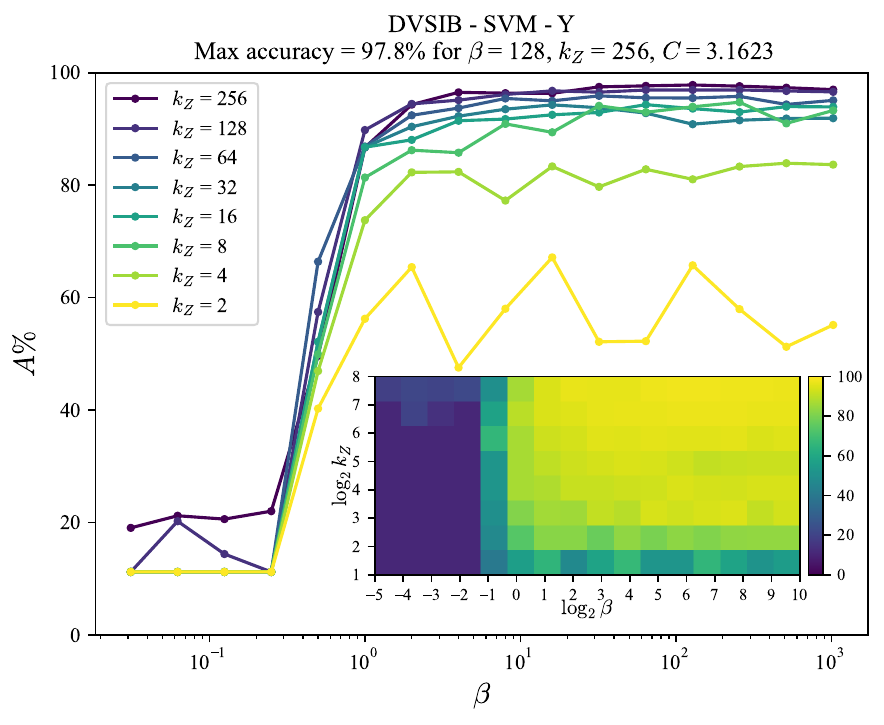}
\includegraphics[width=.85\textwidth]{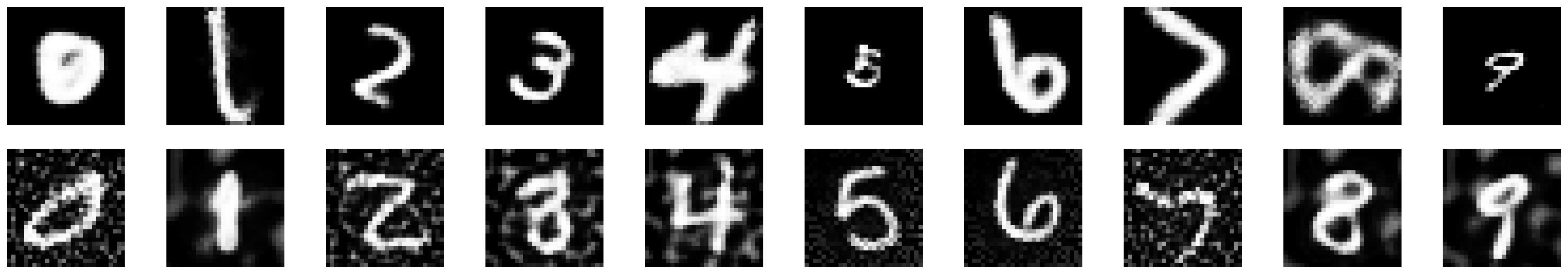}
\end{center}
\vspace{-.15in}
\caption{Top: t-SNE plot of the latent space $Z_Y$ of DVSIB colored by the identity of digits. Top Right: Classification accuracy of an SVM trained on DVSIB's $Z_Y$ latent space. The accuracy was evaluated for DVSIB with a parameter sweep of the trade-off parameter $\beta=2^{-5},...,2^{10}$ and the latent dimension $k_Z=2^1, ..., 2^8$. The max accuracy was  $97.8\%$ for $\beta=128$ and $k_Z=256$. Bottom: Example digits generated by sampling from the DVSIB decoder, $X$ and $Y$ branches.}
%\vspace{-.2in}
\label{Fig:DVSIB}
\end{figure}

To evaluate the methods, we trained them on the training portions of  $X$ and $Y$ without exposure to the true labels. Subsequently, we utilized the trained encoders to compute $Z_{\text{train}}$, $Z_{\text{test}}$, and $Z_{\text{validation}}$ on the respective datasets. To assess the quality of the learned representations, we revealed the labels of $Z_{\text{train}}$ and trained a linear SVM classifier with $Z_{\text{train}}$ and  $\text{labels}_{\text{train}}$. Fine-tuning of the classifier was performed to identify the optimal SVM slack parameter ($C$ value), maximizing accuracy on $Z_{\text{test}}$. This best classifier was then used to predict $Z_{\text{validation}}$, yielding the reported accuracy. We also conducted classification experiments using fully connected neural networks, with detailed results available in the Appendix \ref{App:tables}. For both SVM and the fully connected network, we find the baseline accuracy on the original training data and labels  $(X_\text{train}, \text{labels}_\text{train})$ and $(Y_\text{train}, \text{labels}_\text{train})$, fine-tuning with the test datasets, and reporting the results of the validation datasets. Using Linear SVM enables us to assess the linear separability of the clusters of $Z_X$ and $Z_Y$ obtained through the DR methods. While neural networks excel at uncovering nonlinear relationships that could result in higher classification accuracy, the comparison with a linear SVM establishes a level playing field. It ensures a fair comparison among different methods and is independent of the success of the classifier used for comparison in detecting nonlinear features in the data, which might have been missed by the DR methods. Here, we focus on the results of the $Y$ datasets (MNIST with correlated noise background); results for $X$ are in the Appendix \ref{App:tables}. A parameter sweep was performed to identify optimal $k_Z$ values, ranging from $2^1$ to $2^8$ dimensions on $\log_2$ scale, as well as optimal $\beta$ values, ranging from $2^{-5}$ to $2^{10}$. For methods with private information, $k_{W_X}$ and $k_{W_Y}$ were varied from $2^1$ to $2^6$. The highest accuracy is reported in Tbl.~\ref{table:SVM-Y}, along with the optimal parameters used to obtain this accuracy. 
Additionally, for every method we find the range of $\beta$ and the dimensionality $k_Z$ of the latent variable $Z_Y$ that gives 95\% of the method's maximum accuracy. If the range includes the limits of the parameter, this is indicated by an asterisk.

Figure~\ref{Fig:DVSIB} shows a t-SNE plot of DVSIB's latent space, $Z_Y$, colored by the identity of digits. The resulting latent space has 10 clusters, each corresponding to one digit. The clusters are well separated and interpretable. Further, DVSIB's $Z_Y$ latent space provides the best classification of digits using a linear method such as an SVM showing the latent space is linearly separable.
DVSIB maximum classification accuracy obtained for the linear SVM is 97.8\%. Crucially, DVSIB maintains  accuracy of at least 92.9\% (95\% of 97.8\%) for  $\beta\in [2,1024^*]$ and $k_Z\in  [8,256^*]$. This accuracy is high compared to other methods and has a large range of hyperparameters that maintain its ability to correctly capture information about the identity of the shared digit. DVSIB is a generative method, we have provided sample generated digits from the decoders that were trained from the model graph.

\begin{figure}[ht]
% \vspace{-.25in}
\begin{center}
\includegraphics[width=0.85\textwidth]{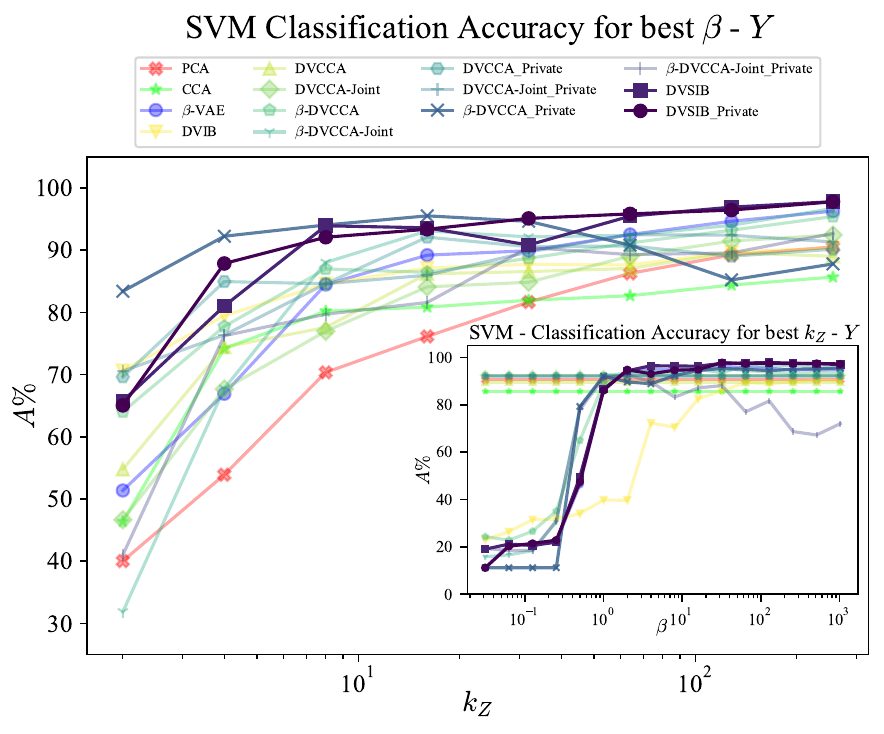}
%\vspace{-.35in}
\end{center}
\caption{The best SVM classification accuracy curves for each method. Here DVSIB and DVSIB-private obtained the best accuracy and, together with $\beta$-DVCCA-private, they had the best accuracy for low latent dimensional spaces.}
\label{Fig:SVM-Y}
\end{figure}

\begin{figure}[ht]
\begin{center}
\includegraphics[width=0.85\textwidth]{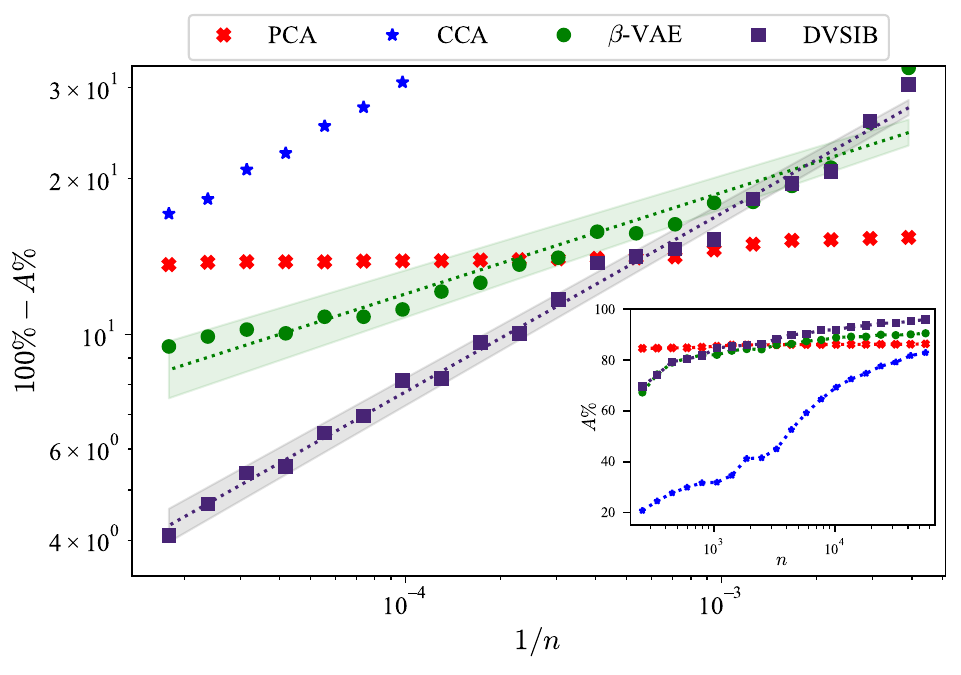}
\end{center}
\caption{Classification accuracy ($A$) of DVSIB has a better sample size ($n$) dependent scaling. Main: a log-log plot of $100\%-A$ vs $1/n$. Slope for fitted lines are $0.345\pm0.007$ for DVSIB, and $0.196\pm0.013$ for $\beta$-VAE, corresponding to a faster increase of accuracy of DVSIB with $n$. Inset: same data, but plotted as $A$ vs $n$. }
\label{Fig:Acc_T_SVM-Y}
\end{figure}

In Fig.~\ref{Fig:SVM-Y}, we show the highest SVM classification accuracy curves for each method. DVSIB and DVSIB-private tie for the best classification accuracy for $Y$. Together with $\beta$-DVCCA-private they have the highest accuracy for all dimensions of the latent space, $k_Z$. In theory, only one dimension should be needed to capture the identity of a digit, but our datasets also contain information about the rotation and scale for $X$ and the strength of the background noise for $Y$. $Y$ should then need at least two latent dimensions to be reconstructed and $X$ should need at least three. Since DVSIB, DVSIB-private, and $\beta$-DVCCA-private performed with the best accuracy starting with the smallest $k_Z$, we conclude that methods with the encoder-decoder graphs that more closely match the structure of the data produce higher accuracy with lower dimensional latent spaces. 

Next, in Fig.~\ref{Fig:Acc_T_SVM-Y}, we compare the sample training efficiency of DVSIB and $\beta$-VAE by training new instances of these methods on a geometrically increasing number of samples $n=[256, 339, 451,\dots,\sim42k, \sim56k]$, consisting of 20 subsamples of the full training data ($X,Y$) to get ($X_{\text{train}_n},Y_{\text{train}_n}$), where each larger subsample includes the previous one. Each method was trained for 60 epochs, and we used $\beta=1024$ (as defined by the DVMIB framework). Further, all reported results are with the latent space size $k_Z=64$. We explored other numbers of training epochs and latent space dimensions (see Appendix~\ref{App:subsamples}), but did not observe qualitative differences. We follow the same procedure as outlined earlier, using the 20 trained encoders for each method to compute $Z_{\text{train}_{n}}$, $Z_{\text{test}}$, and $Z_{\text{validation}}$ for the training, test, and validation datasets. As before, we then train and evaluate the classification accuracy of SVMs for the $Z_Y$ representation learned by each method. Fig.~\ref{Fig:Acc_T_SVM-Y}, inset, shows the classification accuracy of each method as a function of the number of samples used in training. Again, CCA and PCA serve as linear methods baselines. PCA is able to capture the linear correlations in the dataset consistently, even at low sample sizes. However,  it is unable to capture the nonlinearities of the data, and its accuracy does not improve with the sample size. Because of the iterative nature of the implementation of the PCA algorithm \citep{scikit-learn}, it is able to capture some linear correlations in a relatively low number of dimensions, which are sufficiently sampled even with small-sized datasets. Thus the accuracy of PCA barely depends on the training set size. CCA, on the other hand, does not work in the under-sampled regime (see \cite{abdelaleem2024simultaneous} for discussion of this). DVSIB performs uniformly better, at all training set sizes, than the $\beta$-VAE. Furthermore, DVSIB improves its quality faster, with a different sample size scaling. Specifically,  DVSIB and $\beta$-VAE accuracy ($A$, measured in percent) appears to follow the scaling form $A=100-c/n^m$, where $c$ is a constant, and the scaling exponent $m=0.345\pm0.007$ for DVSIB, and $0.196\pm0.013$ for $\beta$-VAE. We illustrate this scaling in Fig.~\ref{Fig:Acc_T_SVM-Y} by plotting a log-log plot of $100-A$ vs $1/n$ and observing a linear relationship.

\section{Conclusion}
We developed an MIB-based framework for deriving variational loss functions for DR applications. We demonstrated the use of this framework by developing a novel variational method, DVSIB. DVSIB compresses the variables $X$ and $Y$ into latent variables $Z_X$ and $Z_Y$ respectively, while maximizing the information between $Z_X$ and $Z_Y$.  The method generates two distinct latent spaces---a feature highly sought after in various applications---but it accomplishes this with superior data efficiency, compared to other methods. The example of DVSIB demonstrates the process of deriving variational bounds for terms present in all examined DR methods. A comprehensive library of typical terms is included in Appendix~\ref{App:Library} for reference, which can be used to derive additional DR methods. Further, we (re)-derive several DR methods, as outlined in Table~\ref{table:methods}. These include well-known techniques such as $\beta$-VAE, DVIB, DVCCA, and DVCCA-private.
MIB naturally introduces a trade-off parameter into the DVCCA family of methods, resulting in what we term the $\beta$-DVCCA DR methods, of which DVCCA is a special case. We implement this new family of methods and show that it produces better latent spaces than DVCCA at $\beta=1$, cf.~Tbl.~\ref{table:SVM-Y}.

We observe that methods that more closely match the structure of dependencies in the data can give better latent spaces as measured by the dimensionality of the latent space and the accuracy of reconstruction (see Figure~\ref{Fig:SVM-Y}).  This makes DVSIB, DVSIB-private, and $\beta$-DVCCA-private perform the best. DVSIB and DVSIB-private both have separate latent spaces for $X$ and $Y$. The private methods allow us to learn additional aspects about $X$ and $Y$ that are not important for the shared digit label, but allow reconstruction of the rotation and scale for $X$ and the background noise of $Y$. We also found that DVSIB can make more efficient use of data when producing latent spaces as compared to $\beta$-VAEs and linear methods. 

Our framework may be extended beyond variational approaches. For instance, in the deterministic limit of VAE, autoencoders can be retrieved by defining the encoder/decoder graphs as nonlinear neural networks $z=f(x)$ and $x=g(z)$. Additionally, linear methods like CCA can be viewed as special cases of the information bottleneck \citep{Weiss2003} and hence must follow from our approach. Similarly, by using specialized encoder and decoder neural networks, e.g., convolutional ones, our framework can implement symmetries and other constraints into the DR process. Overall, the framework serves as a versatile and customizable toolkit, capable of encompassing a wide spectrum of dimensionality reduction methods. With the provided tools and code \citep{githubCode}
, we aim to facilitate the adaptation of the approach to diverse problems.

\section{Supplementary Information}
\subsection{Deriving and Designing Variational Losses}
\label{App:Library}
In the next two subsections, we provide a library of typical terms found in encoder graphs, Appendix \ref{App:Comp}, and decoder graphs, Appendix\ref{App:Model}. In Appendix \ref{App:Methods}, we provide examples of combining these terms to produce variational losses corresponding to beta-VAE, DVIB, beta-DVCCA, beta-DVCCA-joint, beta-DVCCA-private, DVSIB, and DVSIB-private.

\subsubsection{Encoder Graph Components}
\label{App:Comp}
\begin{figure}[ht]
\begin{center}
\begin{tikzpicture}[node distance={15mm}, thick,
main/.style = {draw, circle,minimum size=11mm},
comp/.style = {draw, circle,minimum size=9mm},
labs/.style = {},
every edge quotes/.style = {auto, font=\footnotesize, sloped}
] 
\node[labs] (a) {a.};
\node[main] (1) [below right of=a] {$X$}; 
\node[comp] (2) [below of=1] {$Z_X$}; 
%\draw[->] (1) edge["$p(z_x|x)$"] (2);
\draw[->] (1) -- (2);

\node[labs] (b) [above right of=1] {b.};
\node[main] (3) [below right of=b] {$X$}; 
\node[comp] (4) [below left of=3] {$W_X$};
\node[comp] (5) [below right of=3] {$Z_X$}; 
%\draw[->] (3) edge["$p(w_x|x)$"] (4);
%\draw[->] (3) edge["$p(z_x|x)$"] (5);
\draw[->] (3) -- (4);
\draw[->] (3) -- (5);

\node[labs] (c) [above right of=3] {c.};
\node[main] (6) [below right of=c] {$X$}; 
\node[comp] (8) [below right of=6] {$Z$};
\node[main] (7) [above right of=8] {$Y$};

%\draw[auto, font=\footnotesize]  (8) +(0,.5) node[above]     {$p(z|x,y)$};
\draw[->] (6) -- (8);
\draw[->] (7) -- (8);

\node[labs] (d) [above right of=7] {d.};
\node[main] (9) [below right of=d] {$X$};
\node[main] (10) [right of=9] {$Y$};

\draw (9) -- (10);
\end{tikzpicture}
\end{center}
\caption{Encoder graph components.}
\label{Fig:Gencoder}
\end{figure}
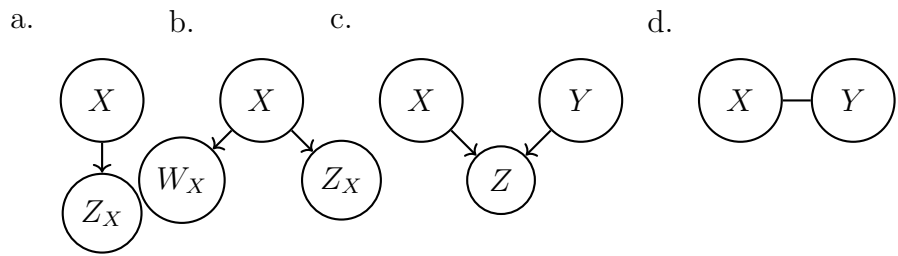

We expand  Sec.~\ref{sec:encoder} and present a range of common components found in encoder graphs across various DR methods, cf.~Fig.~(\ref{Fig:Gencoder}).

\begin{enumerate}[a.]
\item 
This graph corresponds to compressing the random variable $X$ to $Z_X$. Variational bounds for encoders of this type were derived in the main text in Sec.~\ref{sec:encoder} and correspond to the loss:
\begin{align}
\Tilde{I}^{E}(X;Z_X) &=\frac{1}{N}\sum_{i=1}^N D_{\rm KL}(p(z_x|x_i) \Vert r(z_x))\nonumber\\
&\approx \frac{1}{2N}\sum_{i=1}^N \left[\text{Tr}({\Sigma_{Z_X}(x_i)}) +||\vec{\mu}_{Z_X}(x_i)||^2-k_{Z_X}-\ln \det(\Sigma_{Z_X}(x_i)) \right].
\end{align}

\item
This type of encoder graph is similar to the first, but now with two outputs, $Z_X$ and $W_X$. This corresponds to making two encoders, one for $Z_X$ and one for $W_X$, $\Tilde{I}^{E}(Z_X;X)+\Tilde{I}^{E}(W_X;X)$, where
\begin{align}
\Tilde{I}^{E}(Z_X;X)&\approx\frac{1}{N}\sum_{i=1}^N D_{\rm KL}(p(z_x|x_i) \Vert r(z_x)),\\
\Tilde{I}^{E}(W_X;X)&\approx\frac{1}{N}\sum_{i=1}^N D_{\rm KL}(p(w_x|x_i) \Vert r(w_x)).
\end{align}

\item
This type of encoder consists of compressing $X$ and $Y$ into a single variable $Z$. It corresponds to the information loss $I^{E}(Z;(X,Y))$. This again has a similar encoder structure to type (a), but  $X$ is replaced by a joint variable $(X,Y)$. For this loss, we find a variational version:
\begin{equation}
\Tilde{I}^{E}(Z;(X,Y))\approx\frac{1}{N}\sum_{i=1}^N D_{\rm KL}(p(z|x_i,y_i) \Vert r(x_i,y_i)).
\end{equation}

\item
This final type of an encoder term corresponds to information $I^{E}(X,Y)$, which is constant with respect to our minimization. In practice, we drop terms of this type.

\end{enumerate}

\subsubsection{Decoder Graph Components}
\label{App:Model}
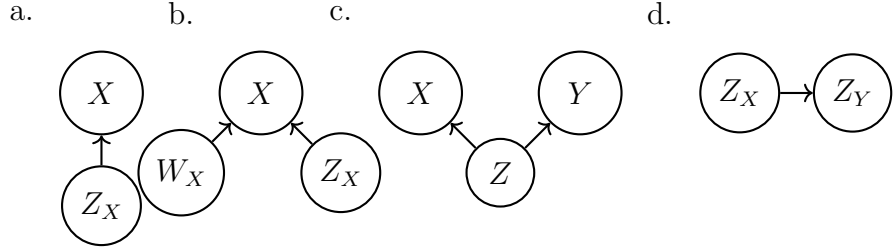
\begin{figure}[ht]
\begin{center}
\begin{tikzpicture}[node distance={15mm}, thick,
main/.style = {draw, circle,minimum size=11mm},
comp/.style = {draw, circle,minimum size=9mm},
labs/.style = {}
] 
\node[labs] (a) {a.};
\node[main] (1) [below right of=a] {$X$}; 
\node[comp] (2) [below of=1] {$Z_X$}; 
\draw[->] (2) -- (1);

\node[labs] (b) [above right of=1] {b.};
\node[main] (3) [below right of=b] {$X$}; 
\node[comp] (4) [below left of=3] {$W_X$};
\node[comp] (5) [below right of=3] {$Z_X$}; 
\draw[->] (4) -- (3);
\draw[->] (5) -- (3);

\node[labs] (c) [above right of=3] {c.};
\node[main] (6) [below right of=c] {$X$}; 
\node[comp] (8) [below right of=6] {$Z$}; 
\node[main] (7) [above right of=8] {$Y$};

\draw[->] (8) -- (6);
\draw[->] (8) -- (7);

\node[labs] (d) [above right of=7] {d.};
\node[comp] (9) [below right of=d] {$Z_X$};
\node[comp] (10) [right of=9] {$Z_Y$};

\draw[->] (9) -- (10);
\end{tikzpicture}
\end{center}
\caption{Decoder graph components.}
\label{Fig:Gdecoder}
\end{figure}

In this section, we elaborate on the decoder graphs that happen in our considered DR methods,  cf.~Fig.~(\ref{Fig:Gdecoder}).

All decoder graphs sample from their methods' corresponding encoder graph.
\begin{enumerate}[a.]
\item 
In this decoder graph, we decode $X$ from the compressed variable $Z_X$. 
Variational bounds for decoders of this type were derived in the main text, Sec.~\ref{sec:decoder}, and they correspond to the loss:
\begin{align}
\Tilde{I}^{D}(X;Z_X)&=H(X)+\frac{1}{N}\sum_{i=1}^N\int dz_x p(z_x|x_i)\ln  q(x_i|z_x)\nonumber\\ 
&\approx H(X)+\frac{1}{MN}\sum_{i,j=1}^{N,M} {-\frac{1}{2}}||(x_i - \mu_{X}({z_x}_{i,j}))||^2,
\end{align}
where $H(X)$ can be dropped from the loss since it doesn't change in optimization.

\item 
This type of decoder term is similar to that in part (a), but $X$ is decoded from two variables simultaneously. The corresponding loss term is $I^{D}(X;(Z_X,W_X))$. We find a variational loss by replacing $Z_X$ in part (a) by $(Z_X,W_X)$:
\begin{equation}
\Tilde{I}^{D}(X;(Z_X,W_X))\approx H(X)+\frac{1}{N}\sum_{i=1}^N\int dz_x dw_x p(z_x,w_x|x_i)\ln(q(x_i|z_x,w_x)),
\end{equation}
where, again, the entropy of $X$ can be dropped.

\item
This decoder term can be obtained by adding two decoders of type (a) together. In this case, the loss term is $I^{D}(X;Z)+I^{D}(Y;Z)$:
\begin{multline}
\Tilde{I}^{D}(X;Z)+\Tilde{I}^{D}(Y;Z)\approx H(X)+H(Y)\\+\frac{1}{N}\sum_{i=1}^N\int dz p(z|x_i)\ln(q(x_i|z))+\frac{1}{N}\sum_{i=1}^N\int dz p(z|y_i)\ln(q(y_i|z)),
\end{multline}
and the entropy terms can be dropped, again.
\item
Decoders of this type were discussed in the main text in Sec.~\ref{sec:MINE}. They correspond to the information between latent variables $Z_X$ and $Z_Y$. We use the MINE estimator to find variational bounds for such terms:
\begin{align}
\Tilde{I}^{D}_{\rm MINE}(Z_X;Z_Y)&=\int dz_x dz_y p(z_x,z_y)\ln \frac{e^{T(z_x,z_y)}}{\mathcal{Z}_{\text{norm}}}
\approx
\frac{1}{NM^2}\sum_{i,j_x,j_y=1}^{N,M,M}\left[T({z_x}_{i,j_x},{z_y}_{i,j_y}) - \ln \mathcal{Z}_{\text{norm}}\right].
\end{align}

\end{enumerate}

\subsubsection{Detailed Method Implementations}
\label{App:Methods}
For completeness, we provide detailed implementations of methods outlined in Tbl.~\ref{table:methods}.

\subsubsubsection{Beta Variational Auto-Encoder}

\begin{figure}[ht]
\begin{center}
\begin{tikzpicture}[node distance={15mm}, thick,
main/.style = {draw, circle,minimum size=11mm},
comp/.style = {draw, circle,minimum size=9mm},
labs/.style = {}
] 
\node[labs] (a) {$G_{\text{encoder}}$};
\node[main] (1) [below right of=a] {$X$}; 
\node[comp] (2) [below of=1] {$Z_X$};
\draw[->] (1) -- (2);

\node[labs] (b) [above right of=1] {$G_{\text{decoder}}$};
\node[main] (3) [below right of=b] {$X$}; 
\node[comp] (4) [below of=3] {$Z_X$};
\draw[->] (4) -- (3);
\end{tikzpicture}
\end{center}
\caption{Encoder and decoder graphs for the beta-variational auto-encoder method}
\label{Fig:betaVAE}
\end{figure}
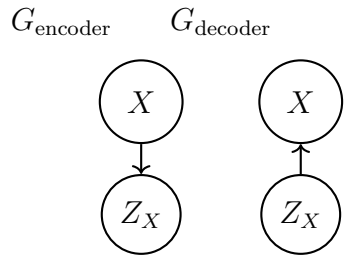

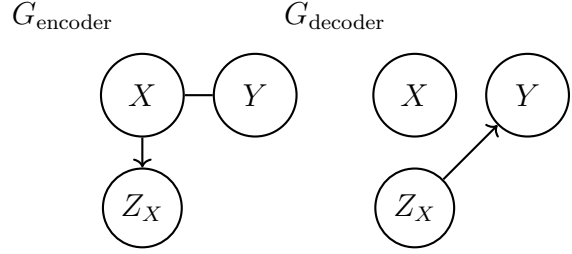
\begin{figure}[ht]
\begin{center}
\begin{tikzpicture}[node distance={15mm}, thick,
main/.style = {draw, circle,minimum size=11mm},
comp/.style = {draw, circle,minimum size=9mm},
labs/.style = {}
] 
\node[labs] (a) {$G_{\text{encoder}}$};
\node[main] (x) [below right of=a] {$X$}; 
\node[main] (y) [right of=x] {$Y$};
\node[comp] (zx) [below of=x] {$Z_X$}; 
\draw (x) -- (y);
\draw[->] (x) -- (zx);

\node[labs] (b) [above right of=y] {$G_{\text{decoder}}$};
\node[main] (mx) [below right of=b] {$X$}; 
\node[main] (my) [right of=mx] {$Y$};
\node[comp] (mzx) [below of=mx] {$Z_X$}; 
\draw[->] (mzx) -- (my);
\end{tikzpicture}
\end{center}
\caption{Encoder and decoder graphs for the Deep Variational Information Bottleneck.}
\label{Fig:betaDVIB}
\end{figure}

A variational autoencoder \citep{Welling2014, Lerchner2016} compresses $X$ into a latent variable $Z_X$ and then reconstructs $X$ from the latent variable, cf.~Fig.~(\ref{Fig:betaDVIB}). The overall loss is a trade-off between the compression $I^{E}(X;Z_X)$ and the reconstruction $I^{D}(X;Z_X)$:
\begin{multline}
I^{E}(X;Z_X)-\beta I^{D}(X;Z_X) \le
\Tilde{I}^{E}(X;Z_X)-\beta \Tilde{I}^{D}(X;Z_X) \\
\lesssim \frac{1}{N}\sum_{i=1}^N D_{\rm KL}(p(z_x|x_i) \Vert r(z_x)) -\beta \left(H(X)+\frac{1}{N}\sum_{i=1}^N\int dz_x p(z_x|x_i)\ln(q(x_i|z_x))\right).
\end{multline}
$H(X)$ is a constant with respect to the minimization, and it can be omitted from the loss. Similar to the main text, DVSIB case, we make ansatzes for forms of each of the variational distributions. We choose parametric distribution families and learn the nearest distribution in these families consistent with the data. Specifically, we assume $p(z_x|x)$ is a normal distribution with mean $\mu_{Z_X}(X)$ and variance $\Sigma_{Z_X}(X)$. We learn the mean and the log-variance as neural networks. We also assume that $q(x|z_x)$ is normal with a mean $\mu_{X}(z_x)$ and a unit variance.  Finally, we assume that $r(z_x)$ is drawn from a standard normal distribution. We then use the re-parameterization trick to produce samples of $z_{x_j}(x)=\mu(x)+\sqrt{\Sigma_{Z_X}(x)} \eta_j$ from $p(z_x|x)$, where $\eta$ is drawn from a standard normal distribution. Overall, this gives:
\begin{multline}
L_{\text{VAE}}=\frac{1}{2N}\sum_{i=1}^N \left[\text{Tr}({\Sigma_{Z_X}(x_i)}) +\vec{\mu}_{Z_X}(x_i)^T\vec{\mu}_{Z_X}(x_i)-k_{Z_X}-\ln\det(\Sigma_{Z_X}(x_i)) \right]\\ -\beta \left(\frac{1}{MN}\sum_{i=1}^N\sum_{j=1}^M {-\frac{1}{2}}(x_i - \mu_{X}(z_{x_j}))^T(x_i - \mu_{X}(z_{x_j})) \right).
\end{multline}
This is the same loss as for a beta auto-encoder. However, following the convention in the Information Bottleneck literature \citep{Bialek2000,Tishby2013}, our $\beta$ is the inverse of the one typically used for beta auto-encoders. A small $\beta$ in our case results in a stronger compression, while a large $\beta$ results in a better reconstruction.

\subsubsubsection{Deep Variational Information Bottleneck}

Just as in the beta auto-encoder, we immediately write down the loss function for the information bottleneck. Here, the encoder graph compresses $X$ into $Z_X$, while the decoder tries to maximize the information between the compressed variable and the relevant variable $Y$, cf.~Fig.~(\ref{Fig:betaDVIB}). The resulting loss function is:
\begin{equation}
L_{\text{IB}} = I^{E}(X;Y)+I^{E}(X;Z_X)-\beta I^{D}(Y;Z_X).
\end{equation}
Here the information between $X$ and $Y$ does not depend on $p(z_x|x)$ and can dropped in the optimization.

Thus the Deep Variational Information Bottleneck \citep{Murphy2017} becomes :
\begin{multline}
L_{\text{DVIB}}\approx \frac{1}{N}\sum_{i=1}^N D_{\rm KL}(p(z_x|x_i) \Vert r(z_x)) -\beta\left( \frac{1}{N}\sum_{i=1}^N\int dz_x p(z_x|x_i)\ln(q(y_i|z_x))\right),
\end{multline}
where we dropped $H(Y)$ since it doesn't change in the optimization. 

As we have been doing before, we choose to parameterize all these distributions by Gaussians and their means and their log variances are learned by neural networks. Specifically, we parameterize $p(z_x|x)=N(\mu_{z_x}(x),\Sigma_{z_x})$, $r(z_x)=N(0,I)$, and $q(y|z_x)=N(\mu_Y,I)$. Again we can use the reparameterization trick and sample from $p(z_x|x_i)$ by $z_{x_j}(x)=\mu(x)+\sqrt{\Sigma_{z_x}(x)} \eta_j$ where $\eta$ is drawn from a standard normal distribution.

\subsubsubsection{Beta Deep Variational CCA}
\begin{figure}[ht]
\begin{center}
\begin{tikzpicture}[node distance={15mm}, thick,
main/.style = {draw, circle,minimum size=11mm},
comp/.style = {draw, circle,minimum size=9mm},
labs/.style = {}
] 
\node[labs] (a) {$G_{\text{encoder}}$};
\node[main] (x) [below right of=a] {$X$}; 
\node[main] (y) [right of=x] {$Y$};
\node[comp] (zx) [below of=x] {$Z_X$}; 
\draw (x) -- (y);
\draw[->] (x) -- (zx);

\node[labs] (b) [above right of=y] {$G_{\text{decoder}}$};
\node[main] (mx) [below right of=b] {$X$}; 
\node[main] (my) [right of=mx] {$Y$};
\node[comp] (mzx) [below of=mx] {$Z_X$}; 
\draw[->] (mzx) -- (my);
\draw[->] (mzx) -- (mx);
\end{tikzpicture}
\end{center}
\caption{Encoder and decoder graphs for beta Deep Variational CCA.}
\label{Fig:betaDCCA}
\end{figure}
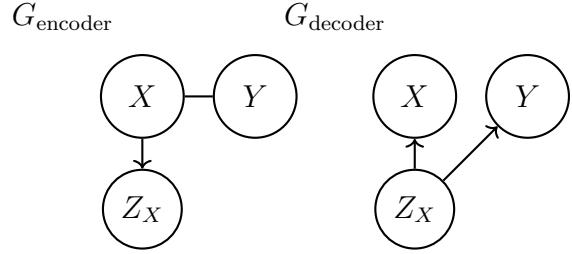
beta-DVCCA, cf.~Fig.~\ref{Fig:betaDCCA}, is similar to the traditional information bottleneck, but now $X$ and $Y$ are both used as relevance variables:
\begin{equation}
L_{\rm DVCCA}=\Tilde{I}^{E}(X;Y)+\Tilde{I}^{E}(X;Z_X)-\beta (\Tilde{I}^{D}(Y;Z_X)+\Tilde{I}^{D}(X;Z_X))
\end{equation}

Using the same library of terms as before, we find:
\begin{multline}
L_{\text{DVCCA}}\approx\frac{1}{N}\sum_{i=1}^N D_{\rm KL}(p(z_x|x_i) \Vert r(z_x))\\
-\beta\left( \frac{1}{N}\sum_{i=1}^N\int dz_x p(z_x|x_i)\ln(q(y_i|z_x)) + \frac{1}{N}\sum_{i=1}^N\int dz_x p(z_x|x_i)\ln(q(x_i|z_x))\right).
\end{multline}
This is similar to the loss function of the deep variational CCA \citep{Livescu2016}, but now it has a trade-off parameter $\beta$. It trades off the compression into $Z$ against the reconstruction of $X$ and $Y$ from the compressed variable $Z$.

\subsubsubsection{Beta joint-Deep Variational CCA}
\begin{figure}[ht]
\begin{center}
\begin{tikzpicture}[node distance={15mm}, thick,
main/.style = {draw, circle,minimum size=11mm},
comp/.style = {draw, circle,minimum size=9mm},
labs/.style = {}
] 
\node[labs] (a) {$G_{\text{encoder}}$};
\node[main] (1) [below right of=a] {$X$};
\node[comp] (3) [below right of=1] {$Z$};
\node[main] (2) [above right of=3] {$Y$};
\draw (1) -- (2);
\draw[->] (1) -- (3);
\draw[->] (2) -- (3);

\node[labs] (b) [above right of=2] {$G_{\text{decoder}}$};
\node[main] (x) [below right of=b] {$X$};
\node[comp] (z) [below right of=x] {$Z$};
\node[main] (y) [above right of=z] {$Y$};
\draw[->] (z) -- (x);
\draw[->] (z) -- (y);
\end{tikzpicture}
\end{center}
\caption{Encoder and decoder graphs for beta joint-Deep Variational CCA.}
\label{Fig:betaDCCAj}
\end{figure}
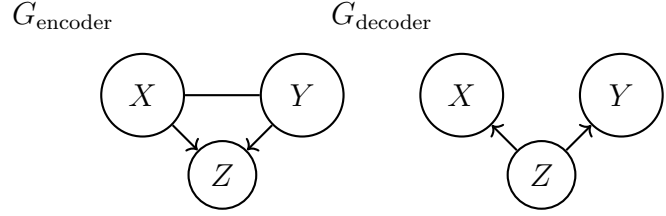

Joint deep variational CCA \citep{Livescu2016}, cf.~Fig.~\ref{Fig:betaDCCAj}, compresses $(X,Y)$ into one $Z$ and then reconstructs the individual terms $X$ and $Y$,
\begin{equation}
L_{\rm DVCCA}=I^{E}(X;Y)+I^{E}((X,Y);Z)-\beta (I^{D}(Y;Z)+I^{D}(X;Z)).
\end{equation}

Using the terms we derived, the loss function is:
\begin{multline}
L_{\text{DVCCA}}\approx\frac{1}{N}\sum_{i=1}^N D_{KL}(p(z|x_i,y_i) \Vert r(z)) \\-\beta \left(\frac{1}{N}\sum_{i=1}^N\int dz p(z|x_i)\ln(q(y_i|z)) + \frac{1}{N}\sum_{i=1}^N\int dz p(z|x_i)\ln(q(x_i|z))\right).
\end{multline}
The information between $X$ and $Y$ does not change under the minimization and can be dropped.

\subsubsubsection{Beta joint-Deep Variational CCA-private}

\begin{figure}[ht]
\begin{center}
\begin{tikzpicture}[node distance={15mm}, thick,
main/.style = {draw, circle,minimum size=11mm},
comp/.style = {draw, circle,minimum size=9mm},
labs/.style = {}
] 
\node[labs] (a) {$G_{\text{encoder}}$};
\node[main] (1) [below right of=a] {$X$};
\node[comp] (3) [below right of=1] {$Z$};
\node[main] (2) [above right of=3] {$Y$};
\node[comp] (4) [below of=1] {$W_X$};
\node[comp] (5) [below of=2] {$W_Y$};
\draw (1) -- (2);
\draw[->] (1) -- (3);
%\draw[->] (2) -- (3);
\draw[->] (1) -- (4);
\draw[->] (2) -- (5);

\node[labs] (b) [above right of=2] {$G_{\text{decoder}}$};
\node[main] (x) [below right of=b] {$X$};
\node[comp] (z) [below right of=x] {$Z$};
\node[main] (y) [above right of=z] {$Y$};
\node[comp] (wx) [below of=x] {$W_X$};
\node[comp] (wy) [below of=y] {$W_Y$};
\draw[->] (wx) -- (x);
\draw[->] (z) -- (x);
\draw[->] (wy) -- (y);
\draw[->] (z) -- (y);
\end{tikzpicture}
\end{center}
\caption{Encoder and decoder graphs for beta Deep Variational CCA-private}
\label{Fig:betaDVCCAp}
\end{figure}
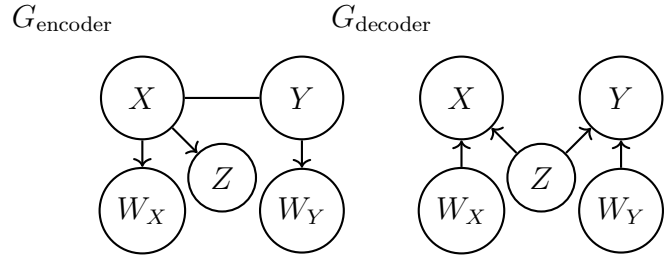

This is a generalization of the Deep Variational CCA \cite{Livescu2016} to include private information, cf.~Fig.~\ref{Fig:betaDVCCAp}. Here $X$ is encoded into a shared latent variable $Z$ and a private latent variable $W_X$. Similarly $Y$ is encoded into the same shared variable and a different private latent variable $W_Y$. $X$ is reconstructed from $Z$ and $W_X$, and $Y$ is reconstructed from $Z$ and $W_Y$. In the joint version $(X,Y)$ are compressed jointly in $Z$ similar to the previous joint methods. What follows is the loss $X$ version of beta Deep Variational CCA-private.
\begin{multline}
L_{\rm DVCCAp}=I^{E}(X;Y)+I^{E}((X,Y);Z)+I^{E}(X;W_X)+I^{E}(Y;W_Y)\\-\beta (I^{D}(X;(W_X,Z))+I^{D}(Y;(W_Y,Z))).
\end{multline}
After the usual variational manipulations, this becomes:
\begin{multline}
L_{\text{DVCCAp}}\approx\frac{1}{N}\sum_{i=1}^N D_{\rm KL}(p(z|x_i) \Vert r(z))+\frac{1}{N}\sum_{i=1}^N D_{\rm KL}(p(w_x|x_i) \Vert r(w_x))\\+
\frac{1}{N}\sum_{i=1}^N D_{\rm KL}(p(w_y|y_i) \Vert r(w_y)) -\beta \left(\frac{1}{N}\sum_{i=1}^N\int dz dw_x p(w_x|x_i)p(z|x_i)\ln(q(y_i|z,w_x)) \right.\\
+ \left. \frac{1}{N}\sum_{i=1}^N\int dz dw_y p(w_y|y_i)p(z|x_i)\ln(q(x_i|z,w_y))\right).
\end{multline}

\subsubsubsection{Deep Variational Symmetric Information Bottleneck}
This has been analyzed in detail in the main text, Sec.~\ref{sec:DVSIB}, and will not be repeated here.

\subsubsubsection{Deep Variational Symmetric Information Bottleneck-private}
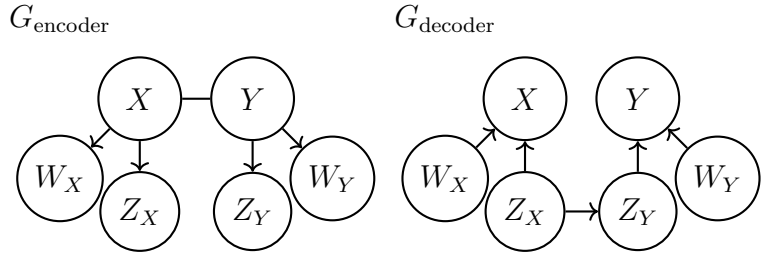
\begin{figure}[ht]
\begin{center}
\begin{tikzpicture}[node distance={15mm}, thick,
main/.style = {draw, circle,minimum size=11mm},
comp/.style = {draw, circle,minimum size=9mm},
labs/.style = {}
] 
\node[labs] (a) {$G_{\text{encoder}}$};
\node[main] (x) [below right of=a] {$X$}; 
\node[main] (y) [right of=x] {$Y$};
\node[comp] (zx) [below of=x] {$Z_X$}; 
\node[comp] (zy) [below of=y] {$Z_Y$};
\node[comp] (wx) [below left of=x] {$W_X$}; 
\node[comp] (wy) [below right of=y] {$W_Y$};
\draw (x) -- (y);
\draw[->] (x) -- (zx);
\draw[->] (y) -- (zy);
\draw[->] (x) -- (wx);
\draw[->] (y) -- (wy);

\node[comp] (mwx) [right of=wy] {$W_X$}; 
\node[main] (mx) [above right of=mwx] {$X$}; 
\node[labs] (b) [above left of=mx] {$G_{\text{decoder}}$};
\node[main] (my) [right of=mx] {$Y$};
\node[comp] (mzx) [below of=mx] {$Z_X$}; 
\node[comp] (mzy) [below of=my] {$Z_Y$};
\node[comp] (mwy) [below right of=my] {$W_Y$};
\draw[->] (mzx) -- (mzy);
\draw[->] (mzx) -- (mx);
\draw[->] (mzy) -- (my);
\draw[->] (mwx) -- (mx);
\draw[->] (mwy) -- (my);

\end{tikzpicture}
\end{center}
\caption{Encoder and decoder graphs for DVSIB-private.}
\label{Fig:betaDVSIBp}
\end{figure}

This is a generalization of the Deep Variational Symmetric Information Bottleneck to include private information. Here $X$ is encoded into a shared latent variable $Z_X$ and a private latent variable $W_X$. Similarly, $Y$ is encoded into its own shared $Z_Y$ variable and a private latent variable $W_Y$. $X$ is reconstructed from $Z_X$ and $W_X$, and $Y$ is reconstructed from $Z_Y$ and $W_Y$. $Z_X$ and $Z_Y$ are constructed to be maximally informative about each another. This results in
\begin{multline}
L_{\text{DVSIBp}}=I^{E}(X;W_X)+I^{E}(X;Z_X)+I^{E}(Y;Z_Y)+I^{E}(Y;W_Y)\\
-\beta \left(I^{D}(Z_X;Z_Y)+I^{D}(X;(Z_X,W_X))+I^{D}(Y;(Z_Y,W_Y))\right).
\end{multline}
After the usual variational manipulations, this becomes (see also main text):
\begin{multline}
L_{\text{DVSIBp}}\approx \frac{1}{N}\sum_{i=1}^N D_{\rm KL}(p(z_x|x_i) \Vert r(z_x)) + \frac{1}{N}\sum_{i=1}^N D_{\rm KL}(p(z_y|x_i) \Vert r(z_y)) \\+\frac{1}{N}\sum_{i=1}^N D_{\rm KL}(p(w_x|x_i) \Vert r(w_x))+\frac{1}{N}\sum_{i=1}^N D_{\rm KL}(p(w_y|y_i) \Vert r(w_y)) \\-\beta \left(\int dz_x dz_y p(z_x,z_y)\ln\frac{e^{T(z_x,z_y)}}{\mathcal{Z}_{\text{norm}}}+\frac{1}{N}\sum_{i=1}^N\int dz_ydw_y p(w_y|y_i)p(z_y|y_i)\ln(q(y_i|z_y,w_y))\right. \\\left.+ \frac{1}{N}\sum_{i=1}^N\int dz_x dw_x p(w_x|x_i)p(z_x|x_i)\ln(q(x_i|z_x,w_x))\right),
\end{multline}
where 
\begin{equation}
\mathcal{Z}_{\rm norm}=\int dz_x dz_y p(z_x)p(z_y)e^{T(z_x,z_y)}.
\end{equation}
\subsection{Multi-variable Losses (More than 2 Views / Variables)}
\label{App:MultiViewLosses}
It is possible to rederive several multi-variable losses that have appeared in the literature within our framework.

\subsubsection{Multi-view Total Correlation Auto-encoder}

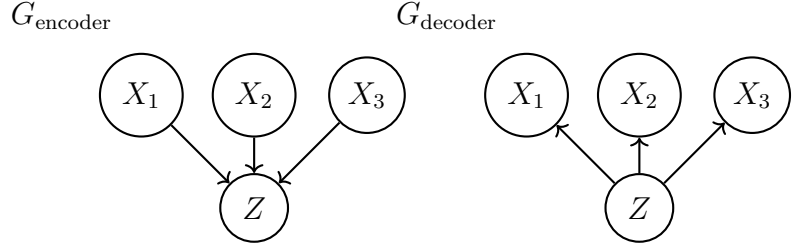
\begin{figure}[ht]
\begin{center}
\begin{tikzpicture}[node distance={15mm}, thick,
main/.style = {draw, circle,minimum size=11mm},
comp/.style = {draw, circle,minimum size=9mm},
labs/.style = {}
] 
\node[labs] (a) {$G_{\text{encoder}}$};
\node[main] (x1) [below right of=a] {$X_1$}; 
\node[main] (x2) [right of=x1] {$X_2$};
\node[comp] (x3) [right of=x2] {$X_3$}; 
\node[comp] (z) [below of=x2] {$Z$};

\draw[->] (x1) -- (z);
\draw[->] (x2) -- (z);
\draw[->] (x3) -- (z);

\node[labs] (b) [above right of=x3] {$G_{\text{decoder}}$};
\node[main] (mx1) [below right of=b] {$X_1$}; 
\node[main] (mx2) [right of=mx1] {$X_2$};
\node[comp] (mx3) [right of=mx2] {$X_3$}; 
\node[comp] (mz) [below of=mx2] {$Z$};

\draw[->] (mz) -- (mx1);
\draw[->] (mz) -- (mx2);
\draw[->] (mz) -- (mx3);

\end{tikzpicture}
\end{center}
\caption{Encoder and decoder graphs for a multi-view auto-encoder.}
\label{Fig:beta}
\end{figure}

Here we demonstrate several graphs for multi-variable losses. This first example consists of a structure, where all the views $X_1$, $X_2$, and $X_3$ are compressed into the same latent variable $Z$. The corresponding decoder produces reconstructed views from the same latent variable $Z$. This is known in the literature as a multi-view auto-encoder.
\begin{equation}
L_{\rm MVAE}=\Tilde{I}^{E}((X_1,X_2,X_3);Z)-\beta (\Tilde{I}^{D}(X_1;Z)+\Tilde{I}^{D}(X_2;Z)+\Tilde{I}^{D}(X_3;Z)).
\end{equation}

Using the same library of terms as before, we find:
\begin{multline}
L_{\text{MVAE}}\approx\frac{1}{N}\sum_{i=1}^N D_{\rm KL}(p(z|{x_1}_i,{x_2}_i,{x_3}_i) \Vert r(z))\\
-\beta\left( \frac{1}{N}\sum_{i=1}^N\int dz p(z|{x_1}_i,{x_2}_i,{x_3}_i)\ln(q({x_1}_i|z)) + \frac{1}{N}\sum_{i=1}^N\int dz p(z|{x_1}_i,{x_2}_i,{x_3}_i)\ln(q({x_2}_i|z))\right.\\
\left.+\frac{1}{N}\sum_{i=1}^N\int dz p(z|{x_1}_i,{x_2}_i,{x_3}_i)\ln(q({x_3}_i|z))\right).
\end{multline}

\subsubsection{Deep Variational Multimodal Information Bottlenecks}

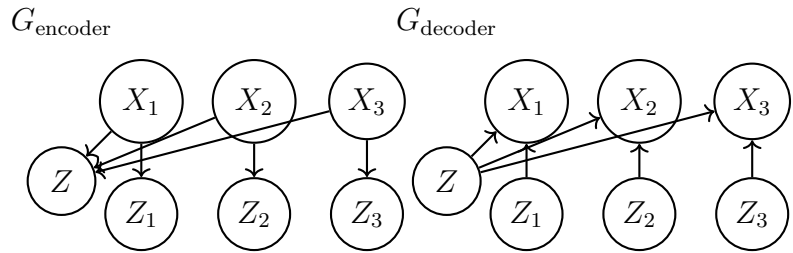
\begin{figure}[ht]
\begin{center}
\begin{tikzpicture}[node distance={15mm}, thick,
main/.style = {draw, circle,minimum size=11mm},
comp/.style = {draw, circle,minimum size=9mm},
labs/.style = {}
] 
\node[labs] (a) {$G_{\text{encoder}}$};
\node[main] (x1) [below right of=a] {$X_1$}; 
\node[main] (x2) [right of=x1] {$X_2$};
\node[comp] (x3) [right of=x2] {$X_3$}; 
\node[comp] (z1) [below of=x1] {$Z_1$};
\node[comp] (z2) [below of=x2] {$Z_2$};
\node[comp] (z3) [below of=x3] {$Z_3$};
\node[comp] (z) [below left of=x1] {$Z$};

\draw[->] (x1) -- (z1);
\draw[->] (x2) -- (z2);
\draw[->] (x3) -- (z3);
\draw[->] (x1) -- (z);
\draw[->] (x2) -- (z);
\draw[->] (x3) -- (z);

\node[labs] (b) [above right of=x3] {$G_{\text{decoder}}$};
\node[main] (mx1) [below right of=b] {$X_1$}; 
\node[main] (mx2) [right of=mx1] {$X_2$};
\node[comp] (mx3) [right of=mx2] {$X_3$}; 
\node[comp] (mz1) [below of=mx1] {$Z_1$};
\node[comp] (mz2) [below of=mx2] {$Z_2$};
\node[comp] (mz3) [below of=mx3] {$Z_3$};
\node[comp] (mz) [below left of=mx1] {$Z$};

\draw[->] (mz) -- (mx1);
\draw[->] (mz) -- (mx2);
\draw[->] (mz) -- (mx3);
\draw[->] (mz1) -- (mx1);
\draw[->] (mz2) -- (mx2);
\draw[->] (mz3) -- (mx3);
\end{tikzpicture}
\end{center}
\caption{Encoder and decoder graphs for Multimodal Information Bottleneck.}
\label{Fig:beta3}
\end{figure}

This example consists of a structure where all the views $X_1$, $X_2$, and $X_3$ are compressed into separate latent views $Z_1$, $Z_2$, and $Z_3$ and one global shared latent variable $Z$. This structure is analogous to DVCCA-private, but it extends to three variables rather than two. It appears in the literature with slightly different variations. In the decoder graph, $X_1$ is reconstructed from both $Z$ and $Z_1$, $X_2$ is reconstructed from both $Z$ and $Z_2$, and $X_3$ is reconstructed from both $Z$ and $Z_3$.
\begin{multline}
L_{\rm DVAE}=\Tilde{I}^{E}((X_1,X_2,X_3);Z)+\Tilde{I}^{E}(X_1;Z_1)+\Tilde{I}^{E}(X_2;Z_2)+\Tilde{I}^{E}(X_3;Z_3)\\
-\beta (\Tilde{I}^{D}(X_1;(Z,Z_1))+\Tilde{I}^{D}(X_2;(Z,Z_2))+\Tilde{I}^{D}(X_3;(Z,Z_3)))
\end{multline}

Using the same library of terms as before, we find:
\begin{multline}
L_{\text{DVAE}}\approx\frac{1}{N}\sum_{i=1}^N D_{\rm KL}(p(z|{x_1}_i,{x_2}_i,{x_3}_i) \Vert r(z))+\sum_{j=1}^3\frac{1}{N}\sum_{i=1}^N D_{\rm KL}(p(z_j|{x_j}_i) \Vert r_j(z))\\
-\beta\left( \frac{1}{N}\sum_{i=1}^N\int dz dz_1 dz_2 dz_3 p(z|{x_1}_i,{x_2}_i,{x_3}_i)p(z_1|{x_1}_i)p(z_2|{x_2}_i)p(z_3|{x_3}_i)\ln(q({x_1}_i|z,z_1))\right.\\
\left.+ \frac{1}{N}\sum_{i=1}^N\int dz dz_1 dz_2 dz_3 p(z|{x_1}_i,{x_2}_i,{x_3}_i)p(z_1|{x_1}_i)p(z_2|{x_2}_i)p(z_3|{x_3}_i)\ln(q({x_2}_i|z,z_2))\right.\\
\left.+\frac{1}{N}\sum_{i=1}^N\int dz dz_1 dz_2 dz_3 p(z|{x_1}_i,{x_2}_i,{x_3}_i)p(z_1|{x_1}_i)p(z_2|{x_2}_i)p(z_3|{x_3}_i)\ln(q({x_3}_i|z,z_3))\right).
\end{multline}

\subsubsection{Discussion}
There exist many other structures that have been explored in the multi-view representation learning literature, including conditional VIB \citep{shi2019, hwang2021}, which is formulated in terms of conditional information. These types of structures are beyond the current scope of our framework. However, they could be represented by an encoder mapping from all independent views $X_\nu$ to $Z$, subtracted from another encoder mapping from the joint view $\vec{X}$ to $Z$. Coupled with this would be a decoder mapping from $Z$ to the independent views $X_\nu$ (or the joint view $\vec{X}$, analogous to the Joint-DVCCA). Similarly, one can use our framework to represent other multi-view approaches, or their approximations \citep{VanderSchaar2021, Hu2021,hwang2021}. This underscores the breadth of methods seeking to address specific questions by exploring known or assumed statistical dependencies within data, and also the generality of our approach, which can re-derive these methods. 

\subsection{Multi-view Information Bottleneck}
\label{App:MVIB}
The multiview information bottleneck (MVIB) \citep{federici2020} attempts to remove redundant information between views $(v_1, v_2)$. This is achieved with the following losses:
\begin{align}
L_1 &= I(z_1;v_1|v_2)-\lambda_1 I(v_2; z_1),\\
L_2 &= I(z_2;v_2|v_1)-\lambda_1 I(v_1; z_2).
\end{align}
These losses are equivalent to two deep variational information bottlenecks performed in parallel. Within our framework, the same algorithm emerges with the encoder graph that compresses $v_1$ into $z_1$ and $v_2$ into $z_2$, while the decoder graph would reconstruct $v_2$ from $z_1$ and $v_1$ from $z_2$.

\cite{federici2020} combines these two losses while enforcing the condition that $z_1$ and $z_2$ are the same. They  bounded the combined loss function to obtain:
\begin{equation}
L_{\rm MVIB}=D_{SKL}(P(z_1|v_1)||P(z_2|v_2))-\beta I(z_1,z_2),
\end{equation}
with  $z_1$ and $z_2$ being the same latent space in this approximation. Here $D_{\rm SKL}$ is the symmetrized KL divergence, $v_i$ corresponds to the two different views, and $z_i$ corresponds to their two latent, compressed representation. (Here we changed the parameter $\beta$ to be in front of $I(z_1,z_2)$, to be consistent with the definition of $\beta$ we use elsewhere in this work.)  While this loss looks similar to the DVSIB loss, it is conceptually different. It attempts to produce latent variables that are as similar to one another as possible (ideally, $z_1=z_2$). In contrast, DVSIB attempts to produce different latent variables that could, in theory, have different units, dimensionalities, and domains, while still being as informative about each other as possible. For example, in the noisy MNIST, $Z_X$ contains information about the labels, the angles, and the scale of images (all needed for reconstructing $X$) and no information about the noise structure. At the same time, $Z_Y$ contains information about the labels and the noise factor only (both needed to reconstruct $Y$). See Appendix \ref{App:2latent} for 2-d latent spaces colored by these variables, illustrating the difference between $Z_X$ and $Z_Y$ in DVSIB. 
Further, in practice, the implementation of MVIB uses the same encoder for both views of the data; this is equivalent to encoding different views using the same function and then trying to force the output to be as close as possible to each other, in contrast to DVSIB.

We evaluate MVIB on the noisy MNIST dataset and include it in Table~\ref{table:SVM-Y}. The performance is similar to that of DVSIB, but slightly worse.

Moreover, MVIB appears to be highly sensitive to parameters and training conditions. Despite employing identical initial conditions and parameters used for training other methods, the approach often experienced collapses during training, resulting in infinities. Interestingly enough, in instances where training persisted for a limited set of parameters (usually low $k_Z$ and high $\beta$), MVIB generated good latent spaces, evidenced by their relatively high classification accuracy.

\subsection{Additional MNIST Results}
\label{App:tables}

In this section, we present supplementary results derived from the methods in Tbl~\ref{table:methods}.

\subsubsection{Additional Results Tables for the Best Parameters}
We report classification accuracy using SVM on data $X$, and using neural networks on both $X$ and $Y$.
\begin{table}[ht]
\caption{Maximum  accuracy from a linear SVM and the optimal $k_Z$ and $\beta$ for variational DR methods on the $X$ dataset. ($^\dag$ fixed values)}
\label{table:SVM-X}
\begin{center}
\begin{tabular}{|l|c|c|c|c|c|c|}
\hline
\textbf{Method} & \textbf{Acc. \%} & \textbf{${k_Z}_\textbf{best}$} & $95\%$ \textbf{${k_Z}_\text{range}$} & \bm{$\beta_\text{best}$} & $95\%$ \textbf{$\beta_\text{range}$} & $\bm{C_\text{best}}$ \\
\hline
Baseline & 57.8 & 784$^\dag$ & - & - & - & 0.01 \\
PCA & 58.0 & 256 & [32,265*] & - & - & 0.1 \\
CCA & 54.4 & 256 & [8,265*] & - & - & 0.032 \\
$\beta$-VAE & 84.4 & 256 & [128,265*] & 4 & [2,8] & 10 \\
DVIB & 87.3 & 128 & [4,265*] & 512 & [8,1024*] & 0.032 \\
DVCCA & 86.1 & 256 & [64,265*] & 1$^\dag$ & - & 31.623 \\
$\beta$-DVCCA & 88.9 & 256 & [128,265*] & 4 & [1,128] & 10 \\
DVCCA-private & 85.3 & 128 & [32,265*] & 1$^\dag$ & - & 31.623 \\
$\beta$-DVCCA-private & 85.3 & 128 & [32,265*] & 1 & [1,8] & 31.623 \\
MVIB & \textbf{93.8} & \textbf{8} & [8,16] & 128 & [128,1024*] & 0.01 \\
DVSIB & \textbf{92.9} & 256 & [64,265*] & 256 & [4,1024*] & 1 \\
DVSIB-private & \textbf{92.6} & 256 & [\textbf{32},265*] & 128 & [8,1024*] & 3.162 \\
\hline

\end{tabular}
\end{center}
\end{table}

\begin{table}[ht]
\caption{Maximum accuracy from a feed forward neural network and the optimal $k_Z$ and $\beta$ for variational DR methods on the $Y$ and the joined $[X,Y]$ datasets. ($^\dag$ fixed values)}
\label{table:NN-Y}
\begin{center}
\begin{tabular}{|l|c|c|c|c|c|}
\hline
\textbf{Method} & \textbf{Acc. \%} & \textbf{${k_Z}_\textbf{best}$} & $95\%$ \textbf{${k_Z}_\text{range}$} & \bm{$\beta_\text{best}$} & $95\%$ \textbf{$\beta_\text{range}$} \\
\hline
Baseline & 92.8 & 784$^\dag$ & - & - & - \\
PCA & 97.6 & 128 & [16,256*] & - & - \\
CCA & 90.2 & 256 & [32,256*] & - & - \\
$\beta$-VAE & \textbf{98.4} & 64 & [8,256*] & 64 & [2,1024*] \\
DVIB & 90.4 & 128 & [8,256*] & 1024 & [8,1024*] \\
DVCCA & 91.3 & 16 & [4,256*] & 1$^\dag$  & - \\
$\beta$-DVCCA & 97.5 & 128 & [8,256*] & 512 & [2,1024*] \\
DVCCA-private & 93.8 & 16 & [2,256*] & 1$^\dag$ & - \\
$\beta$-DVCCA-private & 97.5 & 256 & [\textbf{2},256*] & 32 & [1,1024*] \\
MVIB & 97.5 & 16 & [8,16] & 256 & [128,1024*] \\
DVSIB & \textbf{98.3} & 256 & [\textbf{4},256*] & 32 & [2,1024*] \\
DVSIB-private & \textbf{98.3} & 256 & [\textbf{4},256*] & 32 & [2,1024*] \\
\hline
Baseline-joint & 97.7 & 1568$^\dag$ & - & - & - \\
joint-DVCCA & 93.7 & 256 & [8,256*] & 1$^\dag$ & - \\
$\beta$-joint-DVCCA & \textbf{98.9} & 64 & [8,256*] & 512 & [2,1024*] \\
joint-DVCCA-private & 93.5 & 16 & [4,256*] & 1$^\dag$ & - \\
$\beta$-joint-DVCCA-private & 95.6 & 32 & [4,256*] & 512 & [1,1024*] \\
\hline
\end{tabular}
\end{center}
\end{table}

\begin{table}[ht]
\caption{Maximum  accuracy from a neural network  the optimal $k_Z$ and $\beta$ for variational DR methods on the $X$  dataset. ($^\dag$ fixed values)}
\label{table:NN-X}
\begin{center}
\begin{tabular}{|l|c|c|c|c|c|}
\hline
\textbf{Method} & \textbf{Acc. \%} & \textbf{${k_Z}_\textbf{best}$} & $95\%$ \textbf{${k_Z}_\text{range}$} & \bm{$\beta_\text{best}$} & $95\%$ \textbf{$\beta_\text{range}$} \\
\hline
Baseline & 92.8 & 784$^\dag$ & - & - & - \\
PCA & 91.9 & 64 & [32,256*] & -  & - \\
CCA & 72.6 & 256 & [256,256*] & -  & - \\
$\beta$-VAE & 93.3 & 256 & [16,256*] & 256 & [2,1024*] \\
DVIB & 87.5 & 4 & [2,256*] & 1024 & [4,1024*] \\
DVCCA & 87.5 & 128 & [8,256*] & 1$^\dag$ & - \\
$\beta$-DVCCA & 92.2 & 64 & [8,256*] & 32 & [2,1024*] \\
DVCCA-private & 88.2 & 8 & [8,256*] & 1$^\dag$ & - \\
$\beta$-DVCCA-private & 90.7 & 256 & [4,256*] & 8 & [1,1024*] \\
MVIB & \textbf{93.6} & 8 & [\textbf{8},16] & 256 & [128,1024*] \\
DVSIB & \textbf{93.9} & 128 & [\textbf{8},256*] & 16 & [2,1024*] \\
DVSIB-private & 92.8 & 32 & [8,256*] & 256 & [4,1024*] \\
\hline
\end{tabular}
\end{center}
\end{table}
\newpage

\subsubsection{t-SNE Embeddings at Best Parameters}
\begin{figure}[ht]
\begin{center}
\includegraphics[width=.3\textwidth]{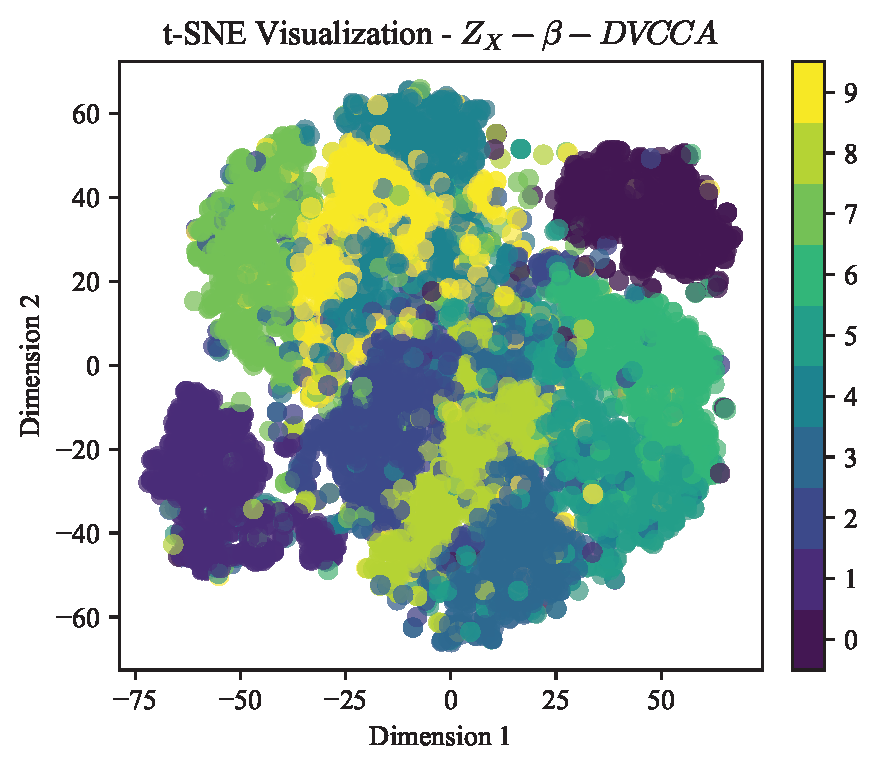}
\includegraphics[width=.3\textwidth]{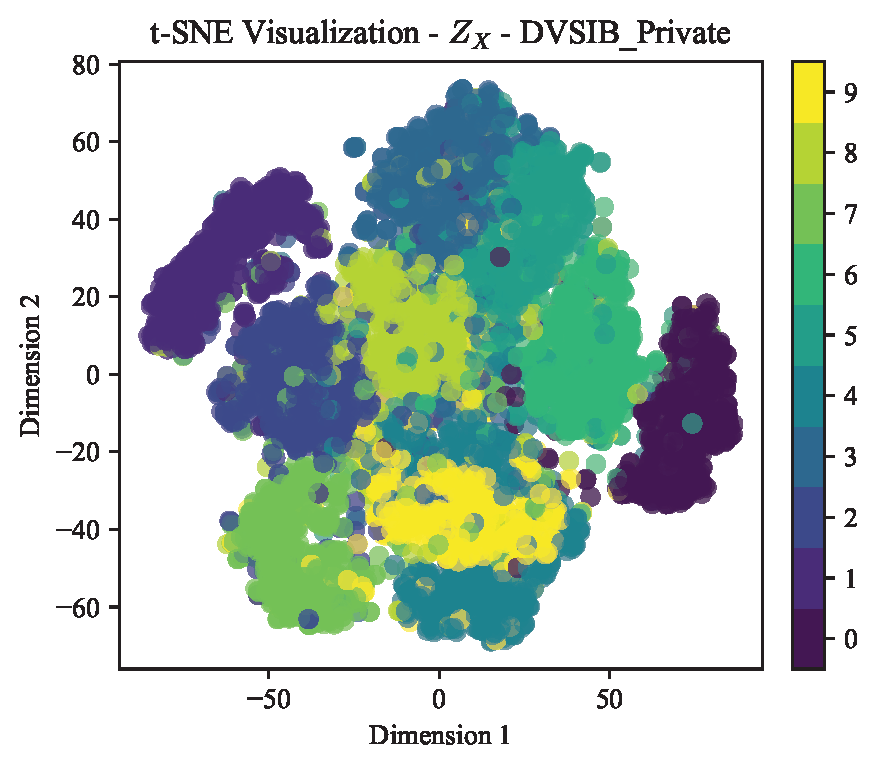}
\includegraphics[width=.3\textwidth]{Figs_Ch3/SI/t-SNE_Visualization-Z_X-beta-DVCCA.png}
\end{center}
\caption{t-SNE X}
\label{Fig:TSNEbestzx}
\end{figure}

\begin{figure}[ht]
\begin{center}
\includegraphics[width=.3\textwidth]{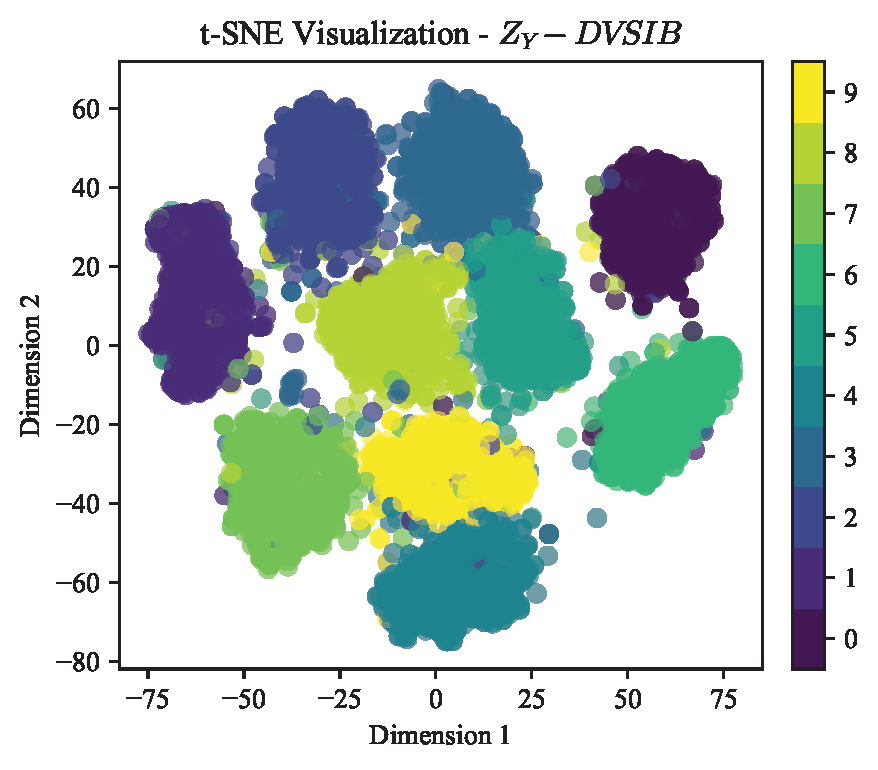}
\includegraphics[width=.3\textwidth]{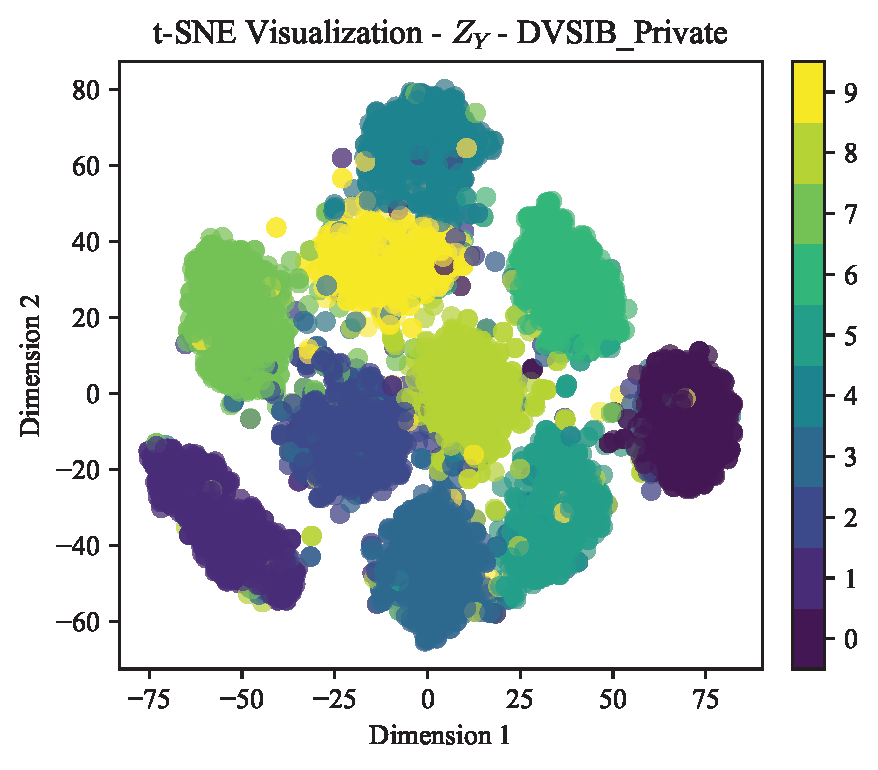}
\includegraphics[width=.3\textwidth]{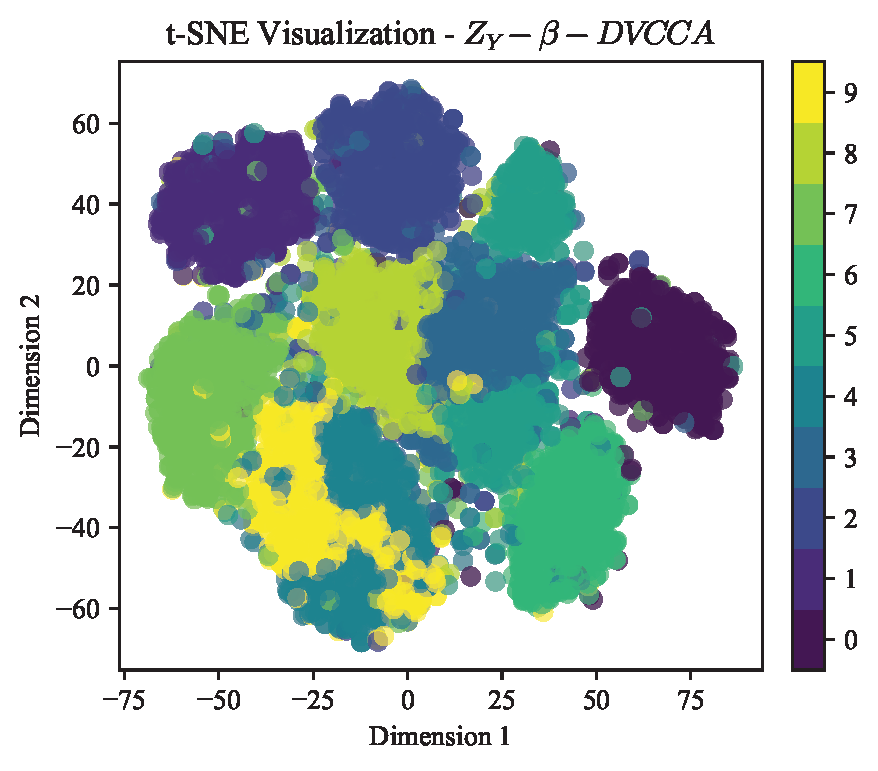}
\end{center}
\caption{t-SNE Y}
\label{Fig:TSNEbestzy}
\end{figure}

Figures \ref{Fig:TSNEbestzx} and \ref{Fig:TSNEbestzy} display 2d t-SNE embeddings for variables $Z_X$ and $Z_Y$ generated by various considered DR methods.

\subsubsection{DVSIB-private Reconstructions for Best Parameters}

\begin{figure}[ht]
\begin{center}
\includegraphics[width=.9\textwidth]{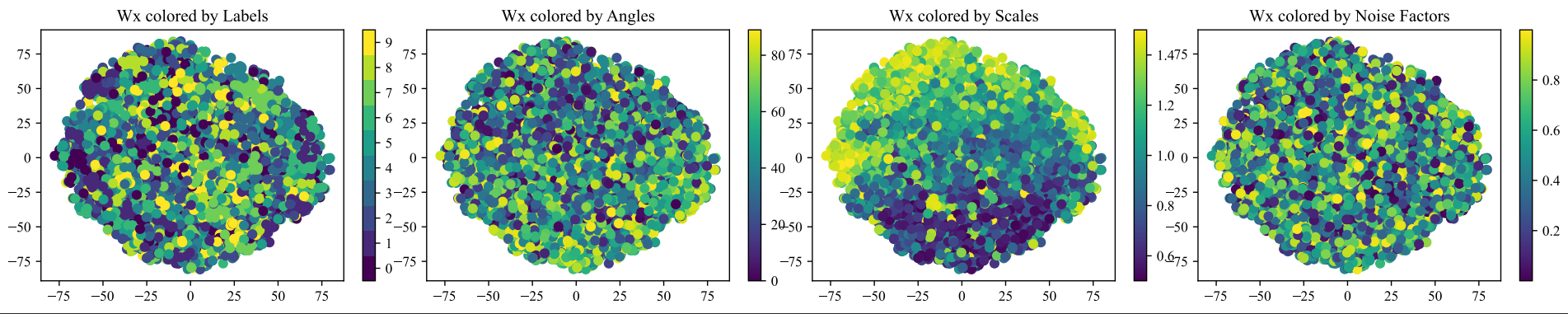}
\includegraphics[width=.9\textwidth]{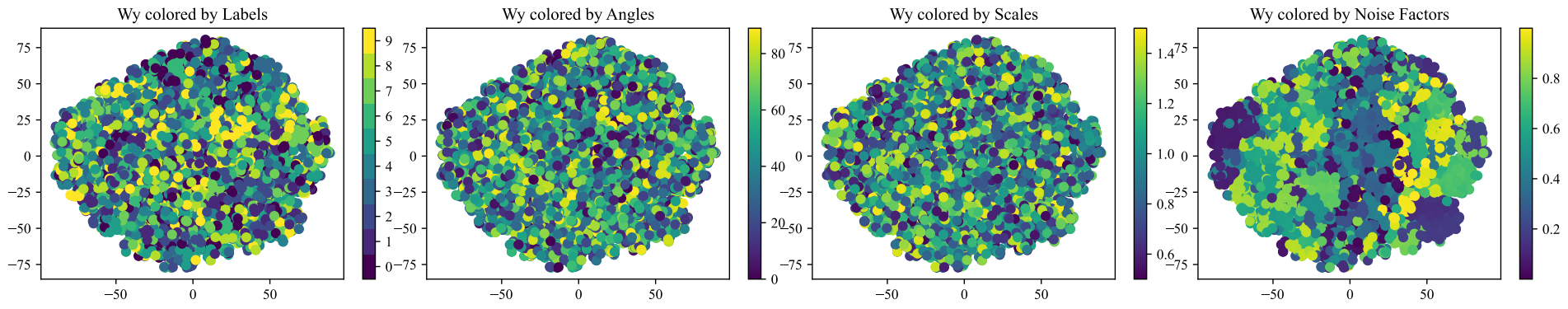}
\includegraphics[width=.9\textwidth]{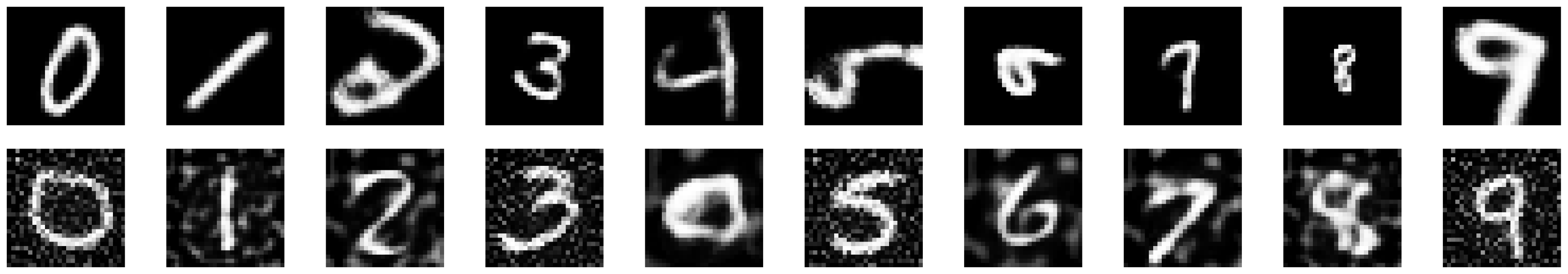}
\end{center}
\caption{Private embeddings of DVSIB-private colored by labels, rotations, scales, and noise factors for $X$ (up), and $Y$ (middle). Reconstructions of the digits using {\em both} shared and private information (bottom) show that the private information allows to produce different backgrounds, scalings, and rotations.
\label{Fig:TSNEbestw}}
\end{figure}

Figure~\ref{Fig:TSNEbestw} shows the t-SNE embeddings of the private latent variables constructed by  DVSIB-private, colored by the digit label. To the extent that the labels do not cluster, private latent variables do not preserve the label information shared between $X$ and $Y$.

% \newpage
\subsubsection{Additional Results at 2 Latent Dimensions}
\label{App:2latent}
\begin{figure}[ht]
\begin{center}
\includegraphics[width=.9\textwidth]{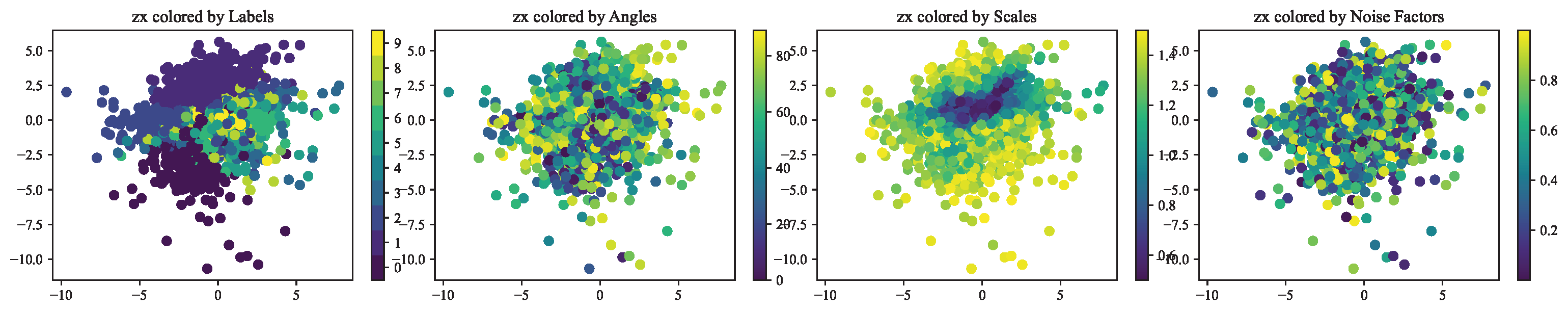}
\includegraphics[width=.9\textwidth]{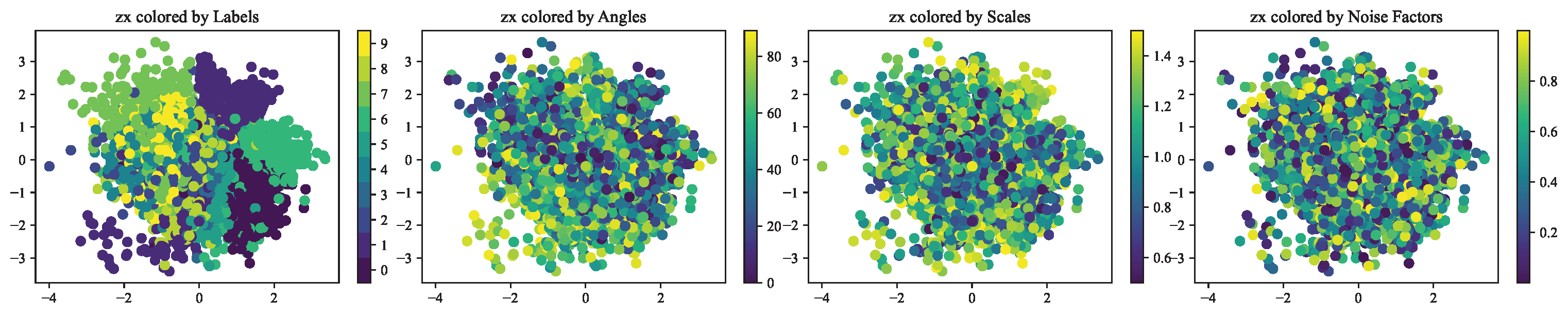}
\includegraphics[width=.9\textwidth]{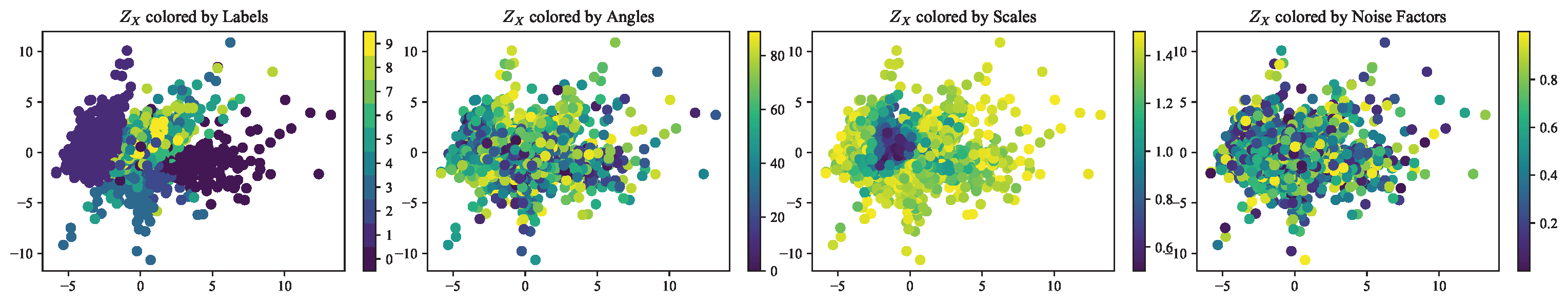}
\caption{Clustering of embeddings when restricting $k_{Z_X}$ to two for DVSIB, DVSIB-private, and $\beta$-DVCCA, results on the $X$ dataset.}
\label{Fig:dz2x}
\end{center}
\end{figure}

\begin{figure}[ht]
\begin{center}
\includegraphics[width=.9\textwidth]{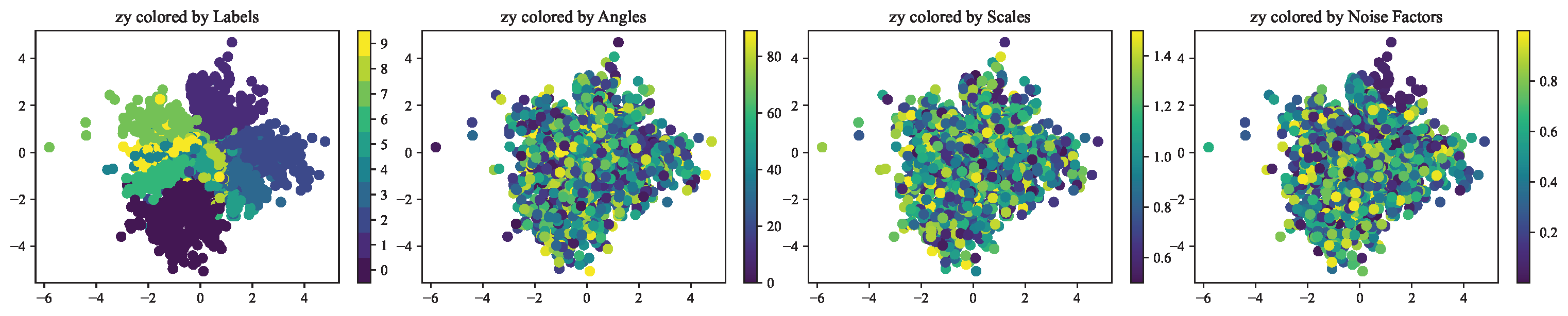}
\includegraphics[width=.9\textwidth]{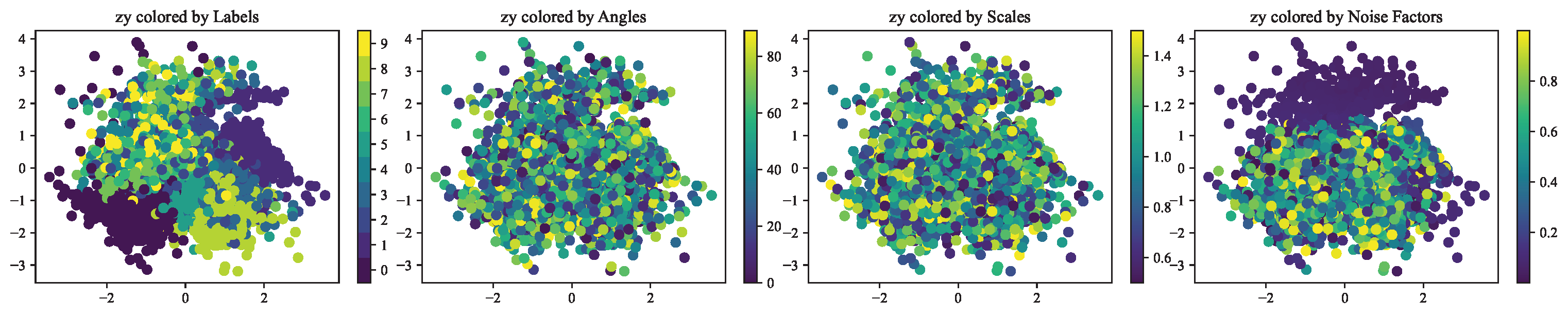}
\includegraphics[width=.9\textwidth]{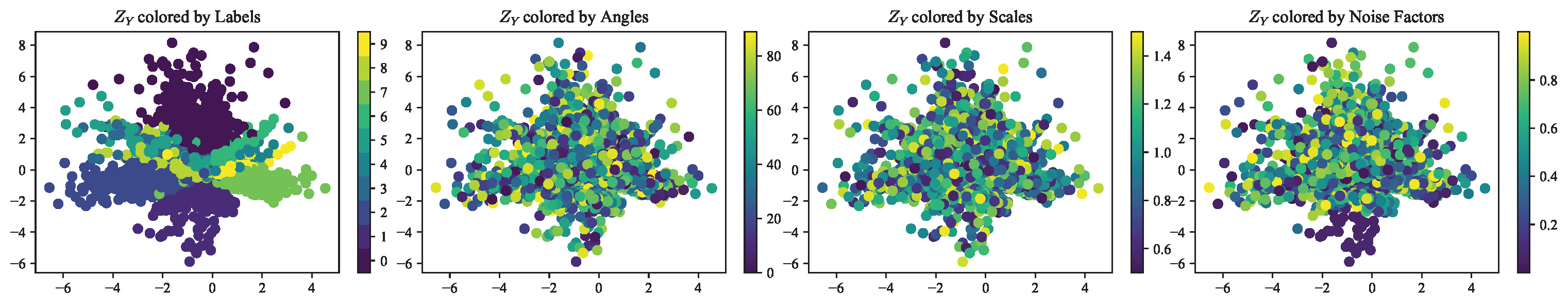}
\caption{Clustering of embeddings when restricting $k_{Z_Y}$ to two for DVSIB, DVSIB-private, and $\beta$-DVCCA, results on the $Y$ dataset.}
\label{Fig:dz2y}
\end{center}
\end{figure}

We now demonstrate how different DR methods behave when the compressed variables are restricted to have not more than 2 dimensions, cf.~Figs.~\ref{Fig:dz2x}, \ref{Fig:dz2y}.

\newpage
\subsection{DVSIB-private Reconstructions at 2 Latent Dimensions}

\begin{figure}[ht]
\begin{center}
\includegraphics[width=.9\textwidth]{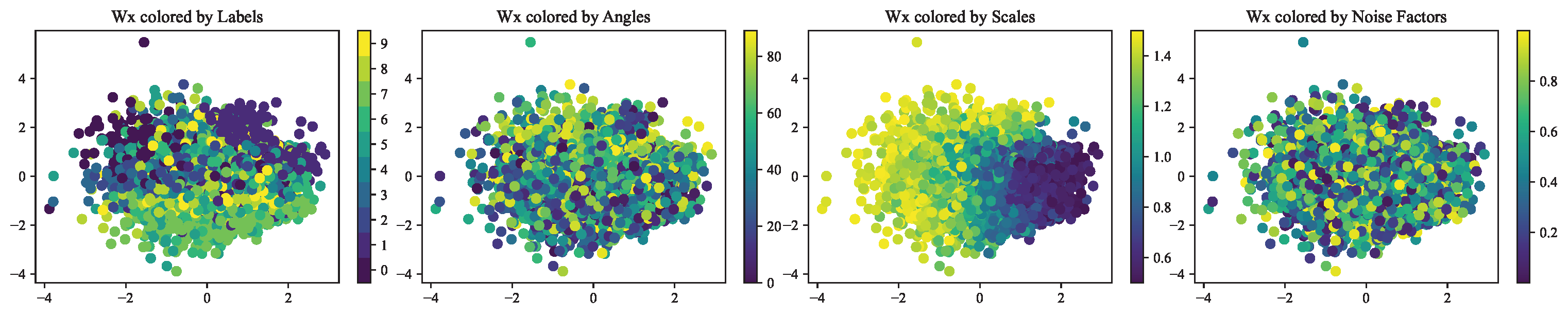}
\includegraphics[width=.9\textwidth]{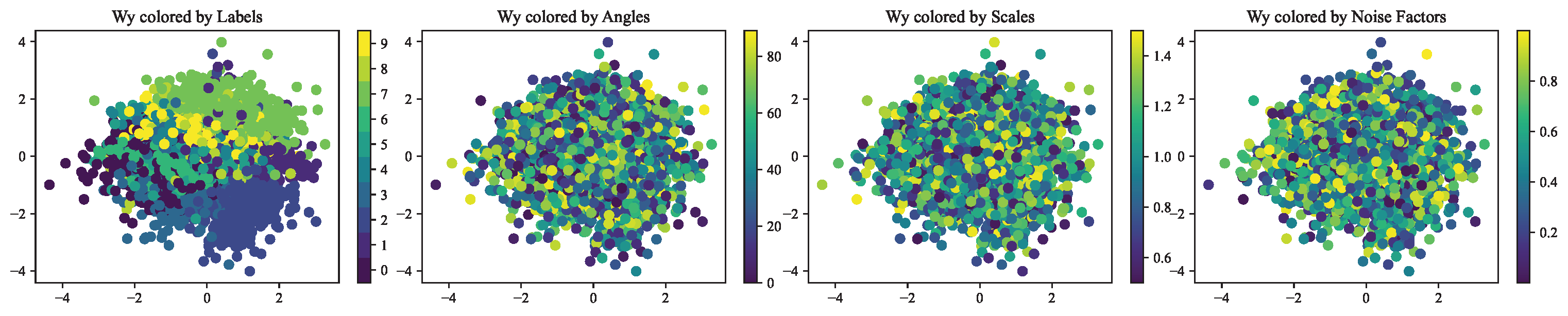}
\end{center}
\caption{Private embeddings of DVSIB-private colored by labels, rotations, scales, and noise factors for $X$ (top), and $Y$ (bottom).
% \caption{Reconstructions of the digits using side and private information}
\label{Fig:dvsibp-dz2w}}
\end{figure}

Figure~\ref{Fig:dvsibp-dz2w} shows the reconstructions of the private latent variables constructed by  DVSIB-private, colored by the digit label, rotations, scales, and noise factors for $X$ (up), and $Y$ (bottom). Private latent variables at 2 latent dimensions preserve a little about the label information shared between $X$ and $Y$, but clearly preserve the scale information for $X$, even at only 2 latent dimensions.

\subsubsection{Testing Training Efficiency}
\label{App:subsamples}
\begin{figure}[ht]
\begin{center}
\includegraphics[width=.9\textwidth]{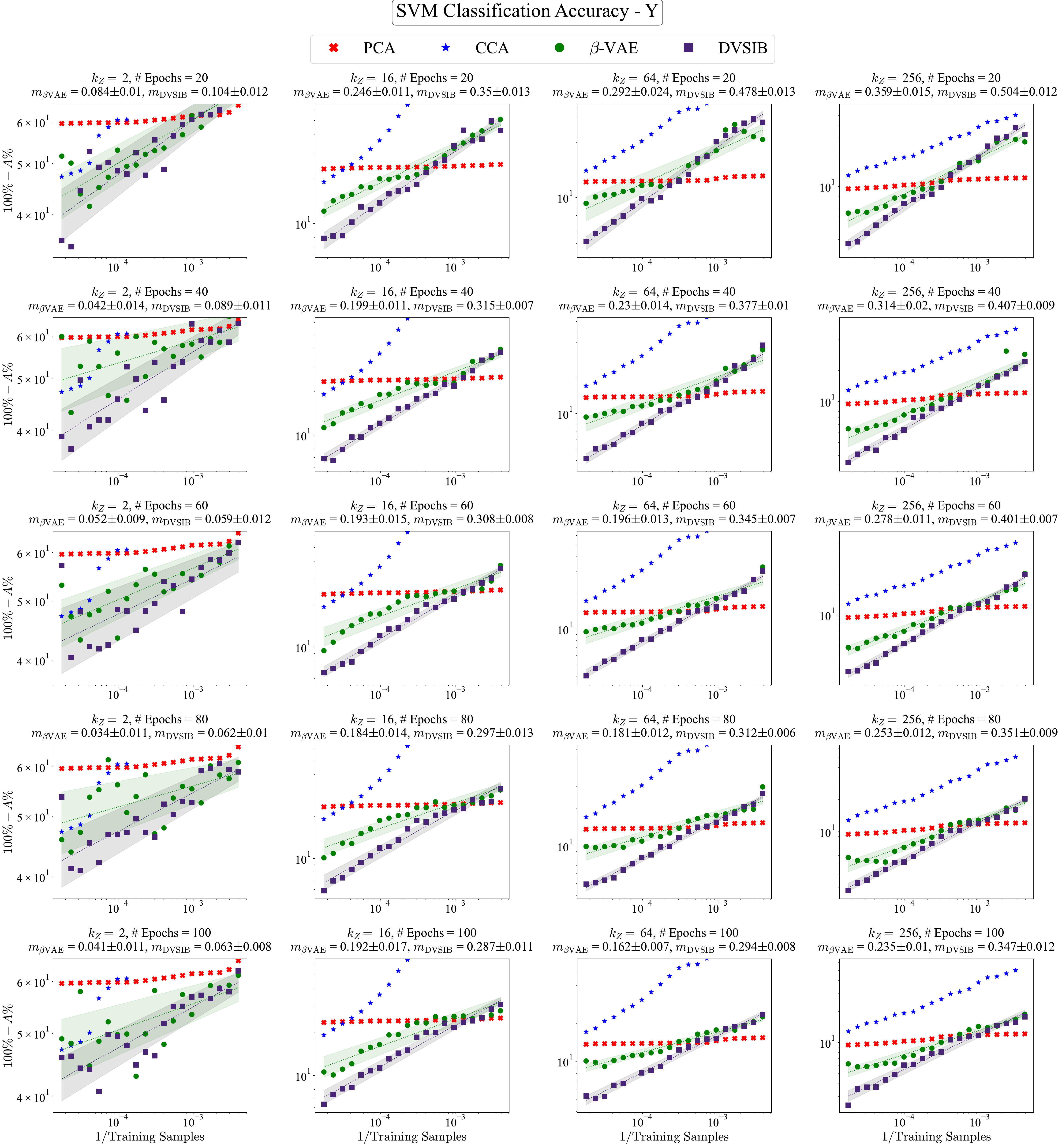}
\end{center}
\caption{Log-log plot of $100 -A$ vs $1/n$. DVSIB has a steeper slope than $\beta$-VAE corresponding to faster convergence with fewer samples for DVSIB. Plots vary $k_Z=2,16,64,256$ and training epochs $20,40,60,80,100$.
\label{Fig:Acc_T_SVM_all}}
\end{figure}

We tested an SVM's classification accuracy for distinguishing digits based on latent subspaces created by DVSIB, $\beta$-VAE, CCA, and PCA trained using different amounts of samples. Figure~\ref{Fig:Acc_T_SVM-Y} in the main text shows the results for 60 epochs of training with latent spaces of dimension $k_{Z_X}=k_{Z_Y}=64$. The DVSIB and $\beta$-VAE were trained with $\beta=1024$. Figure~\ref{Fig:Acc_T_SVM_all} shows the SVM's classification accuracy for a range of latent dimensions (from right to left): $k_{Z_X}=k_{Z_Y}=2, 16, 64, 256$. Additionally, it shows the results for different amounts of training time for the encoders ranging from 20 epochs (top row) to 100 epochs (bottom row). As explained in the main text, we plot a log-log graph of $100-A$ versus $1/n$. Plotted in this way, high accuracy appears at the bottom, and large sample sizes are at the left of the plots. DVSIB, $\beta$-VAE, and CCA often appear linear when plotted this way, implying that they follow the form $A=100-c/n^m$. Steeper slopes $m$ on these plots correspond to a faster increase in the accuracy with the sample size. This parameter sweep shows that the tested methods have not had time to fully converge at low epoch numbers. Additionally, increasing the number of latent dimensions helps the SVMs untangle the non-linearities present in the data and improves the corresponding classifiers.

\chapter{Efficient Estimation of Mutual Information in Very Large Dimensional Data}\label{ch4}

\section{Summary}
\footnote{This section is based on an ongoing work with K.\ Michael Martini and Ilya Nemenman. The development of the idea for this chapter was a collaborative effort among all three authors. The code was written by K.\ Michael Martini and myself. The figures presented in this chapter are produced by myself. The text was jointly written by all authors.
}Mutual information (MI) is a measure of statistical dependencies between two variables. It is a common tool in data analysis in many fields \cite{keys2015application,tang2014millisecond,srivastava2017motor,selimkhanov2014accurate,uda2020application,fairhall2012information}. Thus, accurate estimators of MI from empirical data are needed. However, such estimation is a hard problem, and there are provably no estimators that are universally good for finite datasets \cite{antos2001convergence,paninski2003estimation,holy2002impossibility}. Commonly used estimators perform poorly on high dimensional data, which is a staple of modern experiments. Recently, a series of promising machine learning based MI estimation methods have been introduced. However, it remains unknown how their performance depends on the data set size and on the structure of nonlinearities in the data, as well as on hyperparameters of the estimators, such as the dimensionality of the space used by the neural networks to embed the data and on the duration of training. There are also no accepted tests to signal when the estimators should or should not be trusted. In this Chapter, we systematically explore the dependence of MI estimators on properties of the data sets and on their hyperparameters. We propose and verify a protocol for accurate estimation of MI, with explicit checks for reliability and consistency of the estimators. We show that one can estimate MI reliably from data sets where the number of samples is of the order of the number of dimensions in the data, provided that the statistical dependencies in data can be summarized accurately via embedding in a low dimensional space. This opens opportunities for the use of machine learning based methods for estimating MI from real world datasets.

\section{Introduction}
Mutual information (MI) is a measure of statistical dependence between two variables \cite{shannon1948mathematical}. It is a fundamental quantity in many different disciplines. It captures both linear and nonlinear associations, is reparameterization invariant, and is zero iff the variables are statistically independent. These and other qualities make it a tool of choice for data analysis applications in diverse fields \cite{tang2014millisecond,palmer2015predictive}.
An even wider application of MI as a statistical analysis tool is hampered by the well known difficulty of estimating it from data. Indeed, for continuous variables $X$ and $Y$, MI, measured in {\em nats}, is  
\begin{equation}
    I(X;Y) = \int dx\ dy\, p(x,y) \ln  \frac{p(x,y)}{p(x)p(y)},\label{eq:def}
\end{equation}
where $x$ and $y$ are specific values of the variables, $p(\cdot)$ are the probability density functions of their arguments, and the integration is over the domain of the variables (for discrete $X$ and $Y$, the integrals are replaced by sums, and probability densities by probabilities). Because MI is a nonlinear function of $p$, an unbiased estimate of $p$ plugged into Eq.~(\ref{eq:def}) results in a biased estimate of $I$.  For typical situations, the bias of estimators is a bigger problem than the variance. This was noticed soon after MI was introduced \cite{miller1955note}, and many attempts have been made to design MI estimators that would correct this bias. We now know that, even for discrete variables, no estimator can be unbiased universally, for all underlying distributions, until the number of samples per possible outcome is large \cite{paninski2003estimation}. For continuous variables, the situation is even worse, since MI is reparameterization invariant, and so must be its estimators,  while learning in a reparameterization covariant way is impossible \cite{holy2002impossibility}. 

Nonetheless, significant advances have been made in the field of MI estimation. For continuous variables, which is the focus of this work, the most commonly used estimator is by Kraskov et al.~\cite{kraskov2004estimating}, and its later modifications \cite{holmes2019estimation}. These use the statistics of distances between neighboring data points to estimate the inhomogeneity of $p$ and hence its MI.  While no guarantees of convergence of any estimator to the true value can be made, a common heuristic has been developed \cite{strong1998entropy}. It involves applying the estimator to a subset of the total data, varying the subset size, and verifying that the estimator does not drift statistically significantly as the subset increases towards the full data, thus signifying the absence of the sample-size dependent bias \cite{srivastava2017motor,holmes2019estimation}. Empirically, estimating MI using these approaches is only practical when the dimensionality of both $X$ and $Y$ is, at most, of order 10 \cite{kraskov2004estimating,Nemenman2019}. 

As the number of applications of MI has increased, and the progress of traditional methods has stalled, the need for better methods for estimating MI is now higher than ever. A promising development has been the use of neural networks (NN)  methods to estimate MI via first estimating the deviation of $p(x,y)$ from the product of its marginals using NNs applied to the sampling data \cite{barber2003information, barber2004algorithm, nguyen2010estimating,donsker1983asymptotic,poole2019variational,song2019understanding}. Nominally, these methods can work even for very large dimensional data.  However, in practice, they suffer from multiple drawbacks. First, most of these methods have been tested only on synthetic data with simple multivariate dependence structures and essentially infinite number of samples. Hence their ability to estimate MI in real world scenarios is unknown. Second, since universally good MI estimation for continuous variables is impossible, it is essential for any estimator to have internal consistency checks, which would signal to the user whether the output can or cannot be trusted. Such checks are either at their early stage of development or have not been developed for the NN estimators at all \cite{czyz2024beyond,song2019understanding}. Finally, NN estimators depend on a number of hyperparameters, such as criteria for stopping the training. How to choose these parameters to get an unbiased, low-variance estimate remains unknown.

In this work, we systematically study  NN-based MI estimators. We apply them to synthetic and real world datasets to illustrate their strengths and limitations. We start with multivariate Gaussian data, which is the traditional testing setup. Some methods fail even for these simple cases, especially if the dimensionality of data increases, suggesting that they cannot be trusted for nonlinear, real-world datasets. We observe that the successful methods overcome the curse of dimensionality in MI estimation by first explicitly constructing a low-dimensional embedding of the data and then estimating MI in this lower-dimensional embedding space. We develop a protocol for choosing optimal hyperparameters for NN  estimators and for checking if an estimator is biased. We show that systematically treating NN estimators as a dimensionality reduction problem addresses many challenges inherent in existing approaches. Overall, we show that this approach can estimate MI when the number of samples is as small as of the same order as the number of dimensions in the variables $X$ and $Y$ (not exponential in them!), provided an efficient low-dimensional representation of data can be constructed. Further, it is easy to check if the output of the estimator can be trusted.

\section{Background and Previous Work}

\subsection{Estimation of Mutual Information}
\label{sec:overview}
Estimating MI from finite data is a challenge, as discussed above. The magnitude of the problem can be understood from a simple argument.  Suppose that we estimate MI for continuous variables, with each of the components of the variables bounded in a range $A$. Suppose further that the distribution $p(x,y)$ is smooth, so that the linear size of its smallest feature is $a$. Then the MI will be estimated well when the number of samples is $N\gg (A/a)^K$, where $K$ is the joint dimensionality of $X$ and $Y$, $K=K_X+K_Y$. Even for smooth probability distributions, where $A/a$ is only slightly larger than 1, one needs $N\sim\exp(K)$ samples to estimate MI accurately---the usual curse of dimensionality. Not knowing the parameterization, in which $p$ is smooth, or having unbounded variables makes the problem even harder. While the community has developed many MI estimation methods for continuous variables, none have been able to break this curse and work with dimensions larger than $K\sim 10$ \cite{kraskov2004estimating} (see \cite{holmes2019estimation}, which has pushed this limit). In contrast, most modern data are high-dimensional: for example, images have thousands of pixels, or one can record activity of thousands of neurons.

This inability of traditional methods to deal with the curse of dimensionality gave rise to NN based approaches. Deep NNs can capture complex nonlinear dependencies in large-dimensional data, sometimes from surprisingly few samples \cite{lecun2015deep}. In the context of MI estimation, the class of so called variational deep learning methods has proven to be the most useful \cite{barber2003information, barber2004algorithm, nguyen2010estimating,donsker1983asymptotic}. The basic idea is simple: we have no access to either the joint probability distribution $p(x,y)$ or the marginals $p(x)$, $p(y)$, and we only have samples from them. However, we can transform the problem of estimating MI, that is, of evaluating the integral in Eq.~(\ref{eq:def}), into a problem of evaluating what's known as the {\em critic} and the {\em normalization}. For example, for the MINE estimator of MI \cite{Hjelm2018}, we write  $p(x,y)=p(x)p(y)e^{T(x,y)}/\mathcal{Z}_{\text{norm}}$. Here $T(x,y)$ is the critic, parameterized by a neural network that takes in samples of $x$ and $y$ and returns a single number $T(x,y)$. And  $\mathcal{Z}_{\text{norm}}=\int dx\, dy\, p(x)p(y)e^{T(x,y)}$ is the normalization. While one cannot guarantee that $p(x,y)$ with approximate critic and normalization will normalize properly, one optimizes this approximation over all models implementable by a NN --- hence the variational nature of the approach. With the estimates of $T$ and $\mathcal{Z}_{\text{norm}}$ available, the integral in Eq.~(\ref{eq:def}) is then evaluated by Monte Carlo sampling. 

Other NN methods for MI estimation change the way of parameterizing the normalization factor (such as NWJ \cite{nguyen2010estimating}, Improved MINE \cite{poole2019variational}, clipped MINE -- SMILE \cite{song2019understanding}, etc.) Some change the training protocol, such as by reformulating the problem as a contrastive learning problem (e.~g., InfoNCE \cite{oord2018representation}).

\subsection{Overview of NN information estimators}\label{overview_nn}
Here we briefly introduce NN-based MI estimation methods analyzed in this work. A more holistic review can be found in Refs.~\cite{poole2019variational,song2019understanding}. 

{\bf MINE \& SMILE} \cite{Hjelm2018,song2019understanding} both use the above mentioned critic factorization of $p(x,y)$ resulting in the estimator:
\begin{equation}
I_{\rm MINE}(X,Y)\ge\mathbb{E}_P \left[T(x,y)\right] - \log\left[\mathbb{E}_Q \left(e^{T(x,y)}\right)\right],
\end{equation}
where the first expectation is over the empirical joint probability density, and the second over the product of the empirical marginals. This Monte Carlo sampling leads to biased gradients, which is typically mitigated via a weighted running average over batches \cite{Hjelm2018}.

However, the MINE estimator can have a large variance. To solve this problem, the SMILE estimator clips the joint to marginal density ratio between $e^{-\tau}$ and $e^\tau$, where $\tau$ is some parameter:
\begin{equation}
I_{\rm SMILE}(X,Y)\ge\mathbb{E}_P [T(x,y)] - \log\left[\mathbb{E}_Q \left({\rm clip}(e^{T(x,y)},e^{-\tau},e^\tau)\right)\right].
\end{equation}
Smaller $\tau$ decreases the variance, but at a cost of a larger bias. At $\tau\rightarrow\infty$, SMILE reduces to MINE. In what follows, we use $\tau=5$.

{\bf InfoNCE} \cite{oord2018representation} uses NNs to estimate the conditional distribution $p(y|x)$ as a function of $x$ and $y$, which is the critic $T(y,x)$. The resulting estimate is
\begin{equation}
I_{\rm InfoNCE}(X;Y)\ge\mathbb{E}_P\left[\sum_i^n \log\frac{T(y_i,x_i)}{\frac{1}{K}\sum_{j=1}^n T(y_i,x_j)}\right],
\end{equation}
where the expectation is over the empirical joint distribution, and $n$ is the batch size. This is a form of contrastive predictive coding. The estimator is more biased than the others considered here but also has less variance. The largest MI value  InfoNCE can output is the logarithm of the batch size used for training \cite{poole2019variational,song2019understanding}.

{\bf Different critics}.
We consider two types of critics, separable \cite{bachman2019learning} and concatenated \cite{hjelm2018learning}, which correspond to different factorizations of the critic function $T(x,y)$. Separable critics are of the form $T(x,y)=g(x)\cdot h(y)$, where the functions $g$ and $h$ are implemented via NN embedding into a space of reduced dimension $k$, i.~e. $g:X \rightarrow \mathbb{R}^k$, and $h:Y \rightarrow \mathbb{R}^k$. Concatenated critics use a single NN that takes in the concatenated $x$ and $y$ and produces one output per pair, i.~e. $T:X \times Y \rightarrow \mathbb{R}$. A third combined type of critic, often called a bilinear critic \cite{oord2018representation,henaff2019data,tian2020contrastive}, is of the form $T(x,y) = f(g(x),h(y))$. Similar to the separable critic, the functions $g$ and $h$ are implemented via neural network embeddings into a space of reduced dimension $k$. This is further passed to another concatenated layer(s), similar to the concatenated critic, which produces one output per pair, i.e., $T:X \times Y \rightarrow \mathbb{R}$.

The choice of critic is important \cite{tschannen2019mutual,gosch2020shortcomings}. For example, the separable critic allows for changing the number of embedding dimensions $k$ for both $X$ and $Y$, which plays a significant role. If $k$ is smaller than the intrinsic, latent dimensionality of $K_X$ for $X$ (or $K_Y$ for$Y$), it is likely that the estimator will not capture the full amount of information between the variables. On the other hand, if $k$ is very large, it could lead to what we call ``information washout'', where a small amount of information is distributed among many dimensions, making it indistinguishable from statistical noise in all of them, and hence increasing the variance of the estimator dramatically. Additionally, we will show later that the saturation of information for intermediate values of $ k$ signals the success of the estimator in capturing the mutual information. Another potentially important aspect in choosing a separable vs.\ a concatenated critic is that the former allows the estimator a better chance of capturing the information when variables $X$ and $Y$ have drastically different internal structures (e.~g., different entropies), without variability in one variable drowning that in the other, as could happen for the latter. 
On the other hand, the concatenated critic is more general, allowing for broader relations between the embeddings. For example,  in a concatenated critic, the first dimension of $X$ can be mixed with the second dimension of $Y$, while, in the separable critic,  the first dimension of $g(x)$ interacts only with the first dimension of $h(y)$. However, this mixing comes at the cost of not being able to change the internal dimensionality of the embedding, an important test as we will discuss. Bilinear critics could address this issue by having two separate spaces before allowing them to mix. 

Overall, the choice of the critic depends on the question one is asking and {\em a priori} knowledge about the structure of the data set in question. In this work, we do not aim at a comprehensive evaluation of different critics, and hence focus only on the two basic critics, the separable and the concatenated ones. Similarly, our goal is not a comprehensive comparative analysis of different NN estimators. Thus, even though we explored many NN estimators, including NWJ \cite{nguyen2010estimating}, Jensen-Shannon \cite{nowozin2016f}, Donsker and Varadhan \cite{donsker1983asymptotic}, and TUBA \cite{barber2004algorithm}, we focus on SMILE and InfoNCE, relegating others to SI. We make this choice since (i) these estimators provide examples of high variance/low bias (SMILE) and high bias / low variance (InfoNCE); (ii) none of the other estimators performed uniformly better than these two; and (iii) some of the other methods fail additivity and other consistency checks that MI estimators must obey \cite{song2019understanding}.

\subsection{Problems of NN mutual information estimators}
While NN methods for MI estimation have become popular, the community has not yet addressed serious concerns about them: NN methods are rarely tested on real-world (non-Gaussian) data with only a finite number of samples available. 

To be able to compare the estimates to a known correct answer, a typical (and sometimes the only) test bed is data drawn from relatively low-dimensional ($K_X,K_Y\sim 10$) Gaussian distributions, where the true MI can be calculated analytically.  However, such tests are woefully insufficient. First, simpler, traditional methods \cite{kraskov2004estimating,holmes2019estimation} would be sufficient for $K\sim 10$, for either Gaussian or non-Gaussian data. Testing must involve large dimensional data,  $K\gtrsim 100$, where traditional methods start failing. Yet, only a few NN methods have been systematically tested in this regime.

Further, for Gaussian data, MI and $X-Y$ correlation matrices are related via analytical expressions. Such correlation-based MI estimation sets a natural benchmark for optimal MI estimation  (correlation matrices can be estimated accurately when $K/N\ll1$ \cite{bouchaud2007large}). If a correlation-based approach works, but NN methods do not, the latter are incapable of optimally utilizing the data. This would make it exceedingly unlikely that NN  methods would be able to produce good MI estimates for more complicated, non-Gaussian data at similar sampling ratios. As we show below, most NN methods perform worse---sometimes much worse---than correlation-based estimation, so that sweeping generalizations about their accuracy are hardly warranted.

Finally, NN based MI estimators are typically validated using effectively infinite data \cite{poole2019variational,song2019understanding}. While unlimited data during training is useful for establishing asymptotic consistency of methods, it removes overfitting, unrealistically enhancing the methods' performance. As mentioned above, MI estimators suffer from sample size dependent biases. Thus, success with infinite data does not guarantee success on real-world, finite size datasets.  In practice, one often has $N\sim K$. This renders tests conducted in an infinite data regime not very useful. We will show that it is possible to produce unbiased MI estimators even for some of these severely undersampled, high dimensional cases, provided the data can be accurately embedded into a low dimensional space.

\subsection{MI Estimation as a DR problem}
\label{sec:mi_dr}

As argued in Sec.~\ref{sec:overview}, the accuracy of estimators depends on the dimensionality of the data. Thus, to increase the accuracy, a natural approach is to reduce the dimensionality of $X$ and $Y$ to low dimensional descriptions $Z_X$ and $Z_Y$, respectively, and then estimate $I(Z_X;Z_Y)$ as a proxy for $I(X;Y)$. By the data processing inequality \cite{shannon1948mathematical}, $I(Z_X;Z_Y)- I(X;Y)\equiv \Delta I\le 0$, save for possible statistical fluctuations.  How tight this probabilistic bound is depends on the quality of the dimensionality reduction (DR). Compressing $X$ and $Y$ independently may result in keeping the variation in each of the variables, but not the covariation, resulting in large $|\Delta I|$. The analysis in the preceding Chapters of this dissertation suggests that, to avoid this problem, one needs to compress the variables  {\em simultaneously} \cite{martini2024data,abdelaleem2024simultaneous}, while maximizing $I(Z_X;Z_Y)$, and hence decreasing $|\Delta I|$.

While, to our knowledge, this has not been emphasized in the literature previously, the critics in the NN based estimators reviewed in Sec.~\ref{overview_nn}, indeed, internally perform SDR for the data $X$ and $Y$. This is the easiest to see in a separable critic, for which one trains two networks to perform the following reductions: $Z_X = g(X)$ and $Z_Y = h(Y)$. The critic then is  $T(X,Y) = g(X) \cdot h(Y) =  Z_X \cdot Z_Y$, which is a specific choice that effectively enforces orthogonality of the embeddings: the $i$th component of $g$ is allowed to have information only with the $i$th component of $h$. This approach is similar to other SDR methods we considered in the previous Chapters of this Dissertation, and, specifically, the Deep Variational Symmetric Information Bottleneck (DVSIB) \cite{Abdelaleem2023}. The key difference is that DVSIB uses variational probabilistic encoders, rather than deterministic, feedforward networks as used in the separable critic. Additionally, DVSIB includes two reconstruction decoder networks, $X=X(Z_X)$ and $Y=Y(Z_Y)$, which are not present in the MI estimator (though see Chapter \ref{ch5} for discussion of the removal of the reconstruction from the DVSIB architecture).

The analogy with SDR for the concatenated critic is less clear. However, even here, the NN still maps the combined vector $\{X,Y\}$ into a lower-dimensional vector space $Z$ simultaneously, from which then the critic is evaluated in the output layer of the NN. This approach is a deterministic analog of another SDR method, the Joint Deep Variational Canonical Correlations Analysis (DVCCA) \cite{Abdelaleem2023,Livescu2016}. A clearer analogy can be observed with the bilinear critic, where two networks are used to compress $X$ and $Y$, but they are jointly combined into one variable $Z$. This can be viewed as a deterministic mapping similar to the original Deep Variational Canonical Correlation Analysis (DVCCA) \cite{Livescu2016}.

\section{Results}
We start with the case of infinite training data, which is the common way of testing NN-based MI estimation algorithms. Starting with low-dimensional, Gaussian data and progressing to high-dimensional, nonlinear data, we already observe some of the pitfalls of NN methods. We then show that none of the available algorithms can naively be trusted for the real-world-like, high dimensional regime with limited data. We then propose to use a more careful implementation of the NN based estimators to explicitly change the dimensionality of the low-dimensional embedding space, in which we are maximizing MI between the embeddings of $X$ and $Y$. With additional self-consistency checks, this solves the problem of estimating MI from large-dimensional, undersampled data.

\subsection{Infinite Data}

\subsubsection{Low-dimensional data}
\label{sec:lowd}
\begin{figure}[tbp]
    \centering
    % Only infonce and smile without DVSIB
    \includegraphics[width=\textwidth]{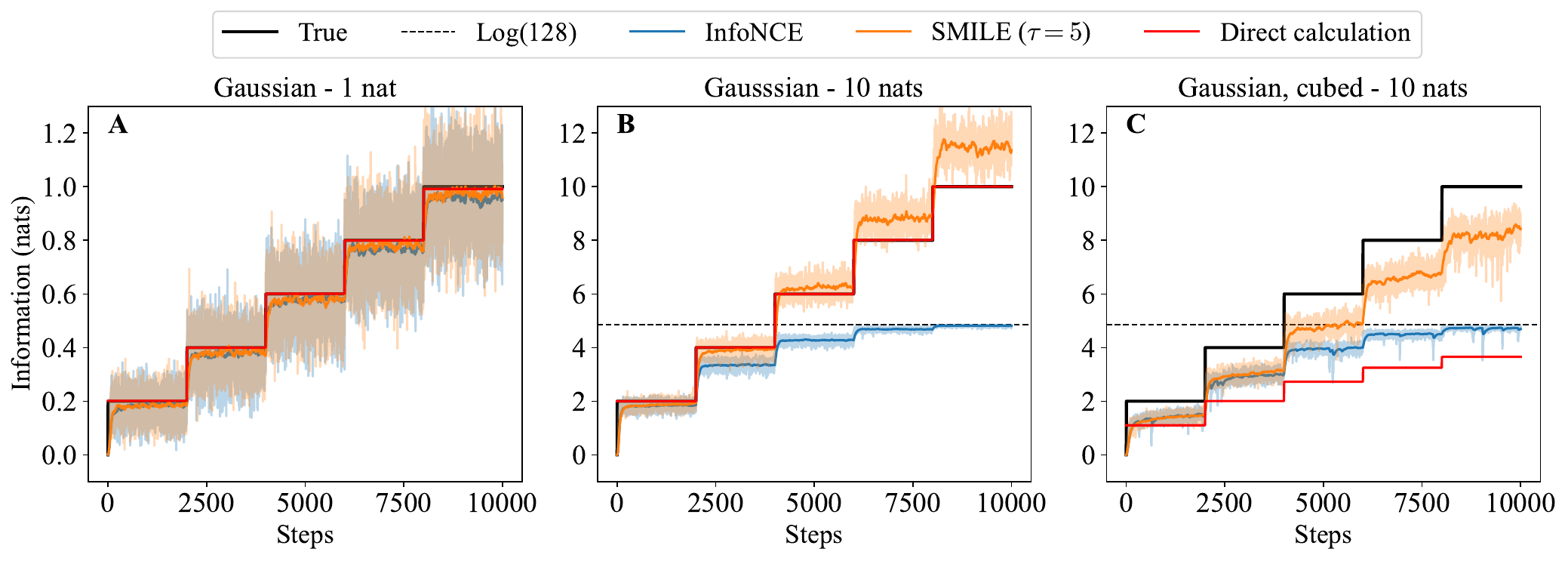}
    \caption{{\bf MI estimation for low-dimensional distributions.} We use a common ``staircase'' protocol for exploring the estimation for different MI values. Here MI jumps after every 2000 steps of training, with every step consisting of a batch of 128 samples. The maximum information is 1 nat in panel (A), and 10 nats in panels (B) and (C). We sample from correlated Gaussian distributions with $K_X=K_Y=10$ and choose the correlation coefficients that result in the needed  MI. All panels show the true information, estimates $I_{\rm InfoNCE}$, $I_{\rm SMILE}$ ($\tau=5$) with a concatenated critic (with $2$ hidden layers, each with $256$ neurons), and the MI estimate from the empirical correlation matrix (denoted as Direct calculation). The correlation estimate uses data from all of the steps preceding the current one for a given MI value. With just one step, the correlation-based MI estimate is hard to distinguish from the true MI. For NN methods, we show their value within each training step (thin lines) and an average smoothed over 100 steps (thick lines). (A) shows that, when the true  MI is small, all methods work well. When MI is high, (B), correlation-based estimation still works, while NN methods degrade: SMILE overestimates and has a large variance, and InfoNCE saturates at $\ln 128$ nats  (logarithm of the batch size). In (C) we add a cubic nonlinearity to the data (see text). Now correlation-based method, non-surprisingly, underestimates MI. However, the effect of the nonlinearity on NN methods is weaker.  }
    \label{fig:10d_gaussian}
\end{figure}

\begin{figure}[htbp]
    \centering
   \includegraphics[width=\textwidth]{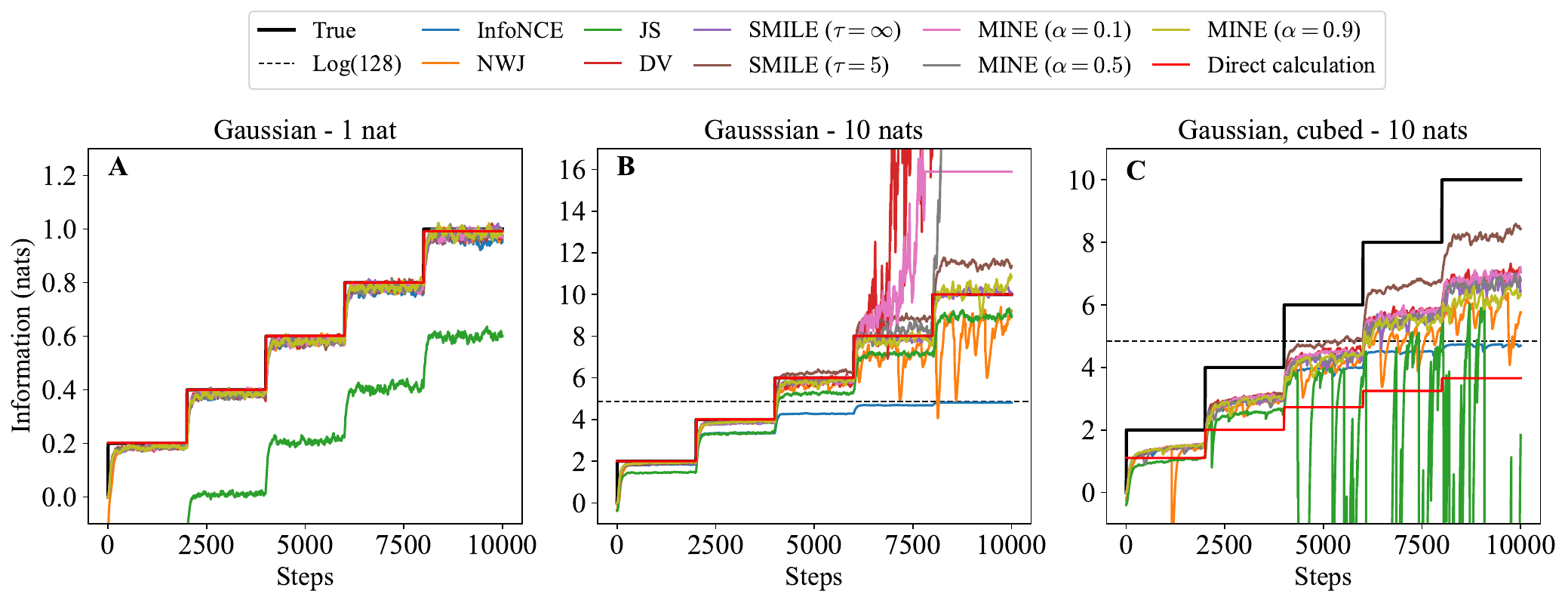}
    \caption{{\bf Performance of NN-based methods for MI estimation.} An extension to Fig.~\ref{fig:10d_gaussian}. Where we used a common ``staircase'' protocol for exploring the estimation for different MI values. Here MI jumps after every 2000 steps of training, with every step consisting of a batch of 128 samples. The maximum information is 1 nat in panel (A), and 10 nats in panels (B) and (C). We sample from correlated Gaussian distributions with $K_X=K_Y=10$ and choose the correlation coefficients that result in the needed  MI. All panels show the true information, estimates $I_{\rm InfoNCE}$, $I_{\rm JS}$ \cite{nowozin2016f}, $I_{\rm NWJ}$ \cite{nguyen2010estimating}, $I_{\rm DV}$ \cite{donsker1983asymptotic}, $I_{\rm SMILE}$ ($\tau=5, \tau=\infty$) \cite{song2019understanding}, $I_{\rm MINE}$ ($\alpha=0.1,\alpha=0.5,\alpha=0.9$) \cite{Hjelm2018} with a concatenated critic (with $2$ hidden layers, each with $256$ neurons), and the MI estimate from the empirical correlation matrix (denoted as Direct calculation). The correlation estimate uses data from all of the steps preceding the current one for a given MI value. With just one step, the correlation-based MI estimate is hard to distinguish from the true MI. For NN methods, we show their value for an average smoothed over 100 steps. (A) shows that, when the true  MI is small, all methods work well (except JS). When MI is high, (B), correlation-based estimation still works, while some NN methods estimates are close to the correct value (MINE ($\alpha=0.9$) and SMILE ($\tau=\infty$)), other methods degrade: DV, MINE ($\alpha=0.1,\alpha=0.5$) overshoots with an extremely large variance. SMILE ($\tau=5)$) overestimates and has a large variance. JS underestimates, and NWJ underestimates with a large variance, and InfoNCE saturates at $\ln 128$ nats  (logarithm of the batch size). In (C) we add a cubic nonlinearity to the data (see text). Now correlation-based method, non-surprisingly, underestimates MI. However, the effect of the nonlinearity on NN methods is weaker (except on JS which degrades greatly).}
    \label{fig:10d_gaussian_all_nn}
\end{figure}

A typical test case for NN based MI estimation uses samples of $X$ and $Y$ from correlated Gaussian distributions with $K_X,K_Y=5\dots20$ dimensions. Typically, there are no correlations among components of $X$ or $Y$, but each $X_i$ is correlated with $Y_i$ with a correlation coefficient $\rho$. A new batch of data, $\mathcal{O}(10^2\dots 10^3)$ samples, is generated at each training step, typically for $\mathcal{O}(10^3\dots10^4)$ steps, resulting in an unrealistic number of samples (typically $N\sim\mathcal{O}(10^6)$). This might seem like a small dataset for modern machine learning tasks. However, for many physical and biological applications, obtaining a dataset of such a size is prohibitively hard. Consequently, $K/N\to0$, and many methods can perform well in these tests (although not all do).  The performance is shown in Fig.~\ref{fig:10d_gaussian}A,B for SMILE and InfoNCE. (This Figure is similar to standard figures shown in \cite{Hjelm2018,song2019understanding}, for example). Both methods work well for small MI values, though SMILE exhibits a large variance, suggesting that averaging MI estimates over multiple steps of training is essential. At larger MI, SMILE starts developing a bias, but InfoNCE completely saturates at $\ln(\text{batch size})$, Fig.~\ref{fig:10d_gaussian}B, confirming that it cannot be trusted in this regime. Additional results for other methods can be found in the SI (Fig.~\ref{fig:10d_gaussian_all_nn})

This relative success of NN methods should come as no surprise. As shown in Fig.~\ref{fig:10d_gaussian}A,B, for jointly Gaussian data, we can calculate MI from the empirical correlation matrix of $X$ and $Y$, $I= \frac{1}{2}\ln \frac{|C|}{|C_{XX}|\, |C_{YY}|}$, where $C$ is the joint correlation matrix, and $C_{XX}$ and $C_{YY}$ are the marginal correlation matrices (numerically, one needs to carefully remove directions with zero correlations before computing the determinants). The error of MI estimation based on the correlations is $\sim K/N$ \cite{bouchaud2007large}, and it is empirically negligible compared to the bias and the variance of NN methods, Fig.~\ref{fig:10d_gaussian}A,B, especially for large MI. While the correlation-based MI is the optimal benchmark, which NNs cannot hope to match, the relatively large discrepancy between the optimum and the NN estimators is a red flag: large variance and bias for these simple data suggest that outputs of NN-based estimators on more complex datasets should be suspect.

{\bf Cubic nonlinearity.} To test how MI estimation methods behave on nonlinear data, we reparameterize the data with an injective continuous nonlinear transformation. Specifically, we keep $X$ unchanged and set $Y\to Y^3$. While the mutual information remains constant under this transformation, the linear correlation changes. This results in the failure of simple linear methods, Fig.~\ref{fig:10d_gaussian}C,  underscoring the need for nonlinear NN approaches. At the same time, NN methods degrade only a little compared to Gaussian data: both SMILE and InfoNCE have similar variances and somewhat larger biases compared to the Gaussian case, Fig.~\ref{fig:10d_gaussian}B.

\begin{figure}[tbp]
    \centering
    % rcca, infonce and smile
    \includegraphics[width=\textwidth]{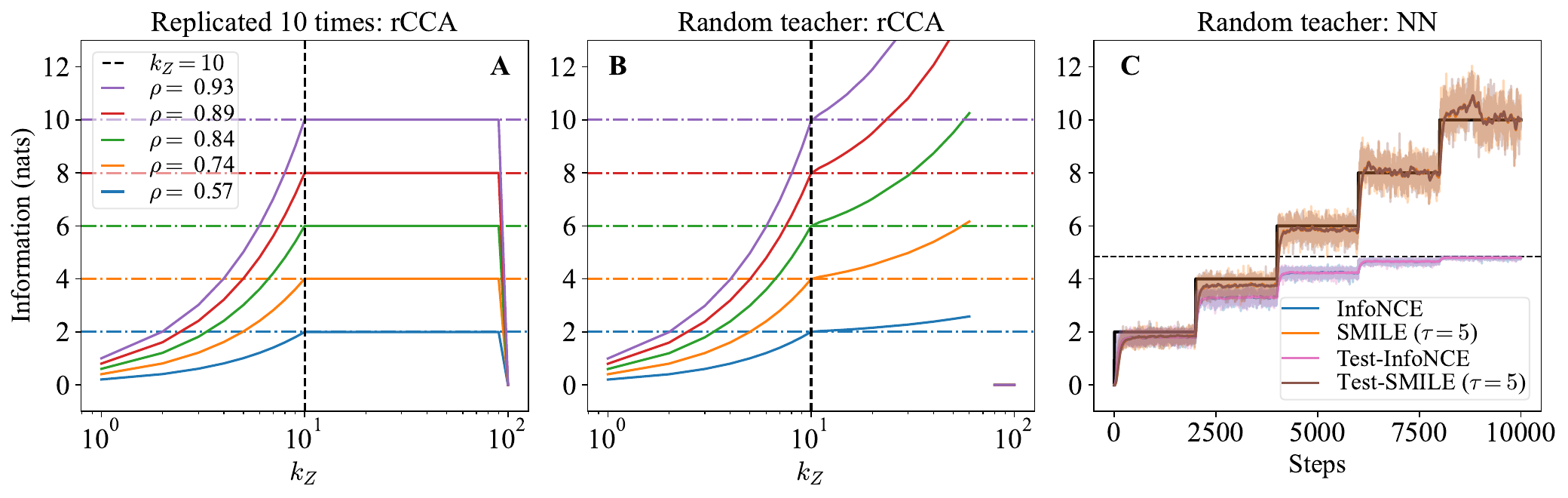}
    \caption{{\bf MI estimation for high-dimensional data.}  (A)  rCCA was applied to the 100 dimensional $X$ and $Y$, each consisting of 10 copies of data from Fig.~\ref{fig:10d_gaussian}B, to produce the reduced descriptions $Z_X$ and $Z_Y$. $I(Z_X;Z_Y)$ estimated via the correlation matrix is plotted. As the dimensionality of the reduced space, $k_Z$ becomes larger than 10 (the number of independent $X$ and $Y$), it is possible to represent all correlations in the $Z_X,Z_Y$ space, and the MI reaches the correct value. (B) Similar analysis for the random features model, produced from 10 independent $X'$ and $Y'$. As the number of linear embedding dimensions increases past 10, there is no obvious saturation, and the MI keeps growing since additional dimensions allow to focus on different parts of the nonlinear relations among the variables. (C) InfoNCE and SMILE with a concatenated critic (with $2$ hidden layers, each with $256$ neurons) on a random features model, with a staircase increase in MI values, as in Fig.~\ref{fig:10d_gaussian}. Here both  $X'$ and $Y'$ are 100 dimensional. Compared to Fig.~\ref{fig:10d_gaussian}C, bias of SMILE nearly disappeared, likely because nonlinearities are weaker, and because SMILE can now use the overcomplete data to estimate correlations more precisely.  Additionally, we show the results of evaluating both methods on a fresh batch of test data, not used in training. The test and the training curves nearly overlap---in this effectively infinite data regime, there is no overfitting.}
    \label{fig:10d_embedded}
\end{figure}

\subsubsection{High dimensional data}
\label{sec:highd}
{\bf Oversampling.} Typical modern experiments have $K\gg 1$, with $K\sim N$. However, the data often can be approximated well with low-dimensional models \cite{Nemenman2022,artemiadis2010emg,williamson2019bridging,Fairhall2015,Fairhall2016,Ganguly2021,Shahbaba2017}. How NN based estimators perform for such high dimensional, but intrinsically simpler data is unknown. We start exploring this with a simple case: starting with 10-dimensional $X'$ and $Y'$, as in Sec.~\ref{sec:lowd}, we replicate the variables ten times each into  $X$ and $Y$, respectively, so that  $K_X=K_Y=100$. Note that $I(X;Y) = I(X';Y')$.  A standard approach for high dimensional data is to perform DR on $X$, $Y$ and then to estimate the correlation matrix and MI $I(Z_X;Z_Y)$ in the reduced space.  For retaining shared information between datasets, Simultaneous DR, such as regularized CCA (rCCA)\footnote{Regularized canonical correlation analysis (rCCA) is a technique that finds linear combinations of two sets of variables that are maximally correlated (cf.~Sec~\ref{rcca}). We fit an rCCA model from \cite{Wang2021} to the equivalent training samples (batch size x number of steps per correlation value in the staircase setup) of data, specifying that the model should yield $k_{Z}$ dimensions. This allows us to calculate the correlation matrix in the reduced space. When calculating the correlation, it is important to be cautious, especially if $k_{Z}>K_Z$ ($K_Z$ being the true latent dimensionality of $X$ and $Y$), as we might have dimensions that contribute very minimally to the correlation. If we calculate the determinant of the correlation matrix as a product of its singular values, these small correlations can lead to numerical instabilities. Therefore, we employ a threshold (typically $10^{-6}\sim 10^{-10}$) to disregard any contribution from singular values below this threshold.} \cite{Hotelling1936,Vinod1976,Strother1998,abdelaleem2024simultaneous} can be used. Figure~\ref{fig:10d_embedded} shows that, indeed rCCA applied to these data results in accurate MI estimation. 

{\bf Random features model}. We generate nonlinear, yet low-dimensional data using teacher NNs. Specifically, we take the 10-dimensional data from  Sec.~\ref{sec:lowd}, pass them through a NN with one fully connected hidden layer with 1024 neurons and a sigmoidal nonlinearity, which then eventually outputs a $K=100$ dimensional $X'$ and $Y'$. Synaptic weights and biases of this teacher NN are initialized with default Pytorch initialization. rCCA, predictably, fails on these data (Fig.~\ref{fig:10d_embedded}B): as we increase the number of embedding dimensions $k_Z$, the reduced variables focus on different parts of the joint probability distribution $p(x,y)$, overestimating MI. In contrast, both SMILE and InfoNCE estimate MI well, Fig.~\ref{fig:10d_embedded}C, even better than in Fig.~\ref{fig:10d_gaussian}C, presumably because the large $K$ allows many ways to detect all statistical structures. As always, for large MI, InfoNCE saturates. Finally,  Fig.~\ref{fig:10d_embedded}C, compares MI values on training and new, test data---in this effectively infinite data regime, there is no overtraining and no difference between the two.

\subsection{Finite data}
\label{sec:FiniteData}
Figure~\ref{fig:10d_embedded} shows that NN methods can estimate MI reliably even for some high-dimensional distributions with nonlinear dependencies. However, Fig.~\ref{fig:10d_gaussian} illustrates that NN methods do not utilize the available data optimally. Since estimating MI is a hard problem {\em precisely} because datasets are finite, it is crucial to understand how the estimator performance depends on the amount and the structure of the data. We explore this on a real and a synthetic dataset, with an even larger dimensionality of $X$ and $Y$,  $K_X=K_Y=784$, and by varying the amount of data $N$ available for training ($N\sim \mathcal{O}(10^5)$).

\begin{figure}[htbp]
    \centering
    \includegraphics[width=\textwidth]{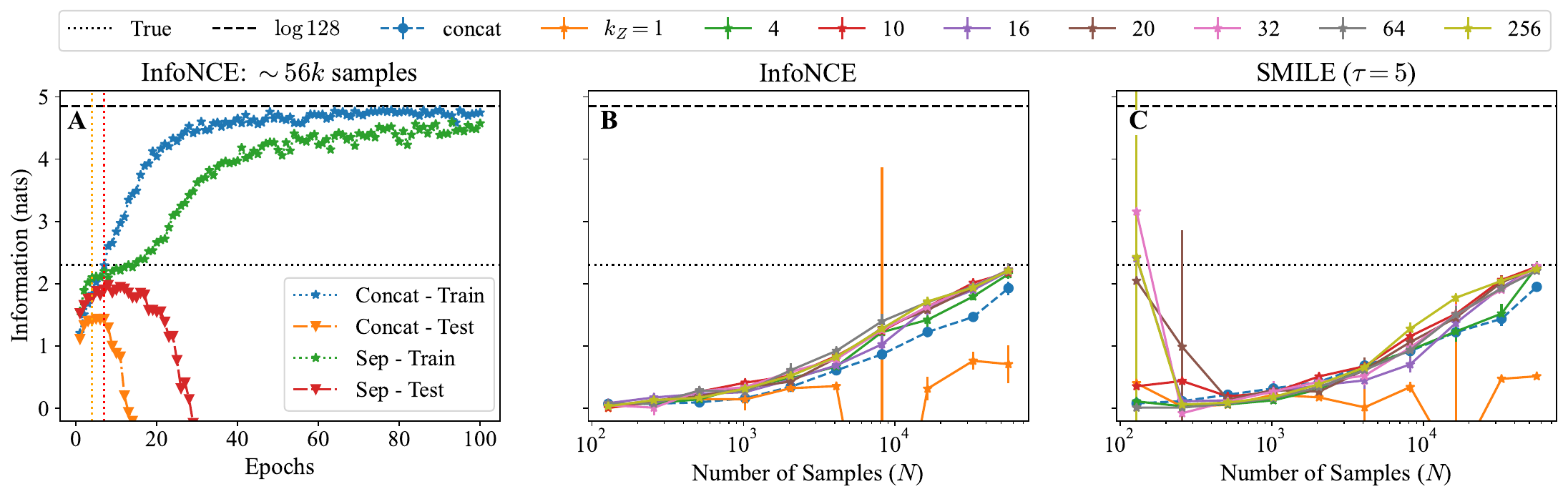}
    \caption{{\bf MI estimation with limited data.} (A) Training and test curves for InfoNCE evaluated on a Noisy MNIST dataset as a function of training time, measured in epochs. The blue and orange curves represent a concatenated critic (with 4 hidden layers, each with 1024 neurons), while the green and red curves represent a separable critic (also with 4 hidden layers, each with 1024 neurons, and an embedding dimension of size 16). The vertical dotted orange and red lines indicate the points where the test information is maximized for each critic, and the corresponding training value is used as the heuristic value for the best estimate of information. The horizontal dotted line represents the theoretical value of the information, $\ln (10)$, and the horizontal dashed line represents the maximum value that InfoNCE can achieve, $\ln ({\rm batch size})$. (B) We use maximum test information heuristic to estimate MI  as a function of sample size with the InfoNCE estimator with different critics. The concatenated critic is shown with blue circles, while the separable critic, which varies by the number of embedding dimensions, is shown with squares in different colors. (C) Same as (B), but for the SMILE estimator $(\tau=5)$. In both (B) and (C), it the information increases with the number of samples $N$ used for training, reaching the true value at large $N$. Additionally, the number of embedding dimensions has minimal effect beyond $4$, suggesting that at least four dimensions are necessary to capture all the information between $X$ and $Y$. We show means and standard deviations for all estimators,  calculated from five independent trials for each data point.} 
    \label{fig:sep_and_concat_mnist}
\end{figure}

\begin{figure}[htbp]
    \centering
    \includegraphics[width=\textwidth]{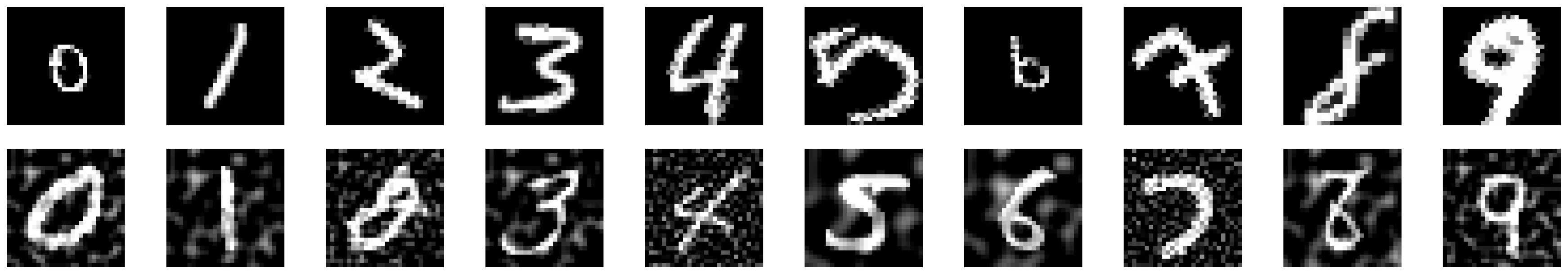}
    \caption{{\bf Samples from  Noisy MNIST data set} \cite{Abdelaleem2023}. The data set contains $\sim 56k$ training and $\sim 7k$ test digit pair, $X$ and $Y$. In each pair, the same digit (but its two different instances) are corrupted by scaling / rotation, $X$, and by background noise, $Y$.}
    \label{SIfig:mnist}
\end{figure}

{\bf Noisy MNIST.} We use the Noisy MNIST dataset, introduced in Ref.~\cite{Abdelaleem2023} as an adaptation of Refs.~\cite{Haffner1998, Bilmes2015, Livescu2016}. These data comprise two distinct views of data, $X$ and $Y$, each of dimensionality $28 \times 28=784$ pixels, as shown in SI Fig.~\ref{SIfig:mnist}. The first view is an image of a digit from the standard MNIST dataset, subjected to a random rotation by an angle uniformly sampled between $0$ and $\frac{\pi}{2}$, and to a scaling by a factor uniformly distributed between $0.5$ and $1.5$. The second view consists of another image with the same digit identity (but different instance) with an additional background layer of Perlin noise \citep{Perlin1985}, with the noise factor uniformly distributed between $0$ and $1$. Both views are normalized to an intensity range of $[0,1)$ and then flattened to form an array of $K_X=K_Y=784$ dimensions. The dataset comprises a total of $55996\sim 56k$ images for training and $\sim 7k$ images for testing. Overall, the only correlations between the two views are via the digit class, and all ten digits are represented nearly uniformly, so that $I(X;Y)\approx\ln 10$. Further, correlations are strongly nonlinear and dimensionality is high, so that the dataset is a natural testbed for MI estimation. A typical training and testing curves for InfoNCE with separable and concatenated critics are shown in Fig.~\ref{fig:sep_and_concat_mnist}. Panel A in Fig~\ref{fig:sep_and_concat_mnist} shows that MI estimates on training and test data diverge during training, well before the estimates reach true MI  value, indicating overfitting. Thus, it is important to stop the training early. We do this by reporting the training value which is obtained when MI estimate on the test data peaks, which we denote as the {\em max test} heuristic\footnote{One might also think of other heuristics. For example, we can consider a {\em zero test} heuristic, in which we choose to report the train value that corresponds to the zero test value. This means that we have no information left between $X$ and $Y$ and essentially cannot trust any train value beyond this point. This approach might be useful for downstream tasks where we want to stop our training once we have captured all possible information. However, we observe that this is not a good heuristic for accurately estimating the information, as it is likely to result in overfitting. We have also found this to be true empirically.}. 

Panels B and C show the mean MI values, averaged over five trials, based on the maximum test heuristic evaluated at different numbers of training samples for InfoNCE and SMILE, respectively. Both critics used for the estimators are built with 4 hidden layers, each containing 1024 neurons. Additionally, the separable critic, which allows for varying the embedding dimension, is tested from embedding dimension of 1 up to 256, for both $X$ and $Y$. Apart from the significant fluctuations observed when the separable critic embedding is restricted to only one dimension or when trained on a very small number of samples, we observe an increase in MI closer towards the true value with more training samples, which is expected. Notably, we observe that varying the size of the embedding for the separable critic results in a saturation of MI after $k_Z = 4$. This indicates that $k_Z\geq 4$ is necessary to capture all the information between $X$ and $Y$ in this specific dataset. When the embedding dimension, $k_Z$, is too small to capture the underlying data structure, MI is underestimated by all methods. When $k_Z\geq 4$, and for as large as $k_Z=256$, information estimates are accurate, as long as $N\gg K$. Note that since $\ln(10)$ is smaller than the logarithm of the batch size, InfoNCE and SMILE would work well, though InfoNCE has a smaller variance, as expected. This observation can be used as a consistency check for the estimators: changing the dimensionality of the embedding space should eventually lead to saturation, reflecting the intrinsic dimensionality of the data, and observing this is one evidence of accuracy of the estimation. Performing this check via adjusting the embedding size is a key advantage of using a separable critic, and aligning the dimensionality of the embedding method with the dimensionality of the data is important for successfully capturing all relevant shared information \cite{abdelaleem2024simultaneous}.

{\bf Large dimensional Gaussians with varying number of signal dimensions.} 
\begin{figure}[tbp]
    \centering
    \includegraphics[width=\textwidth]{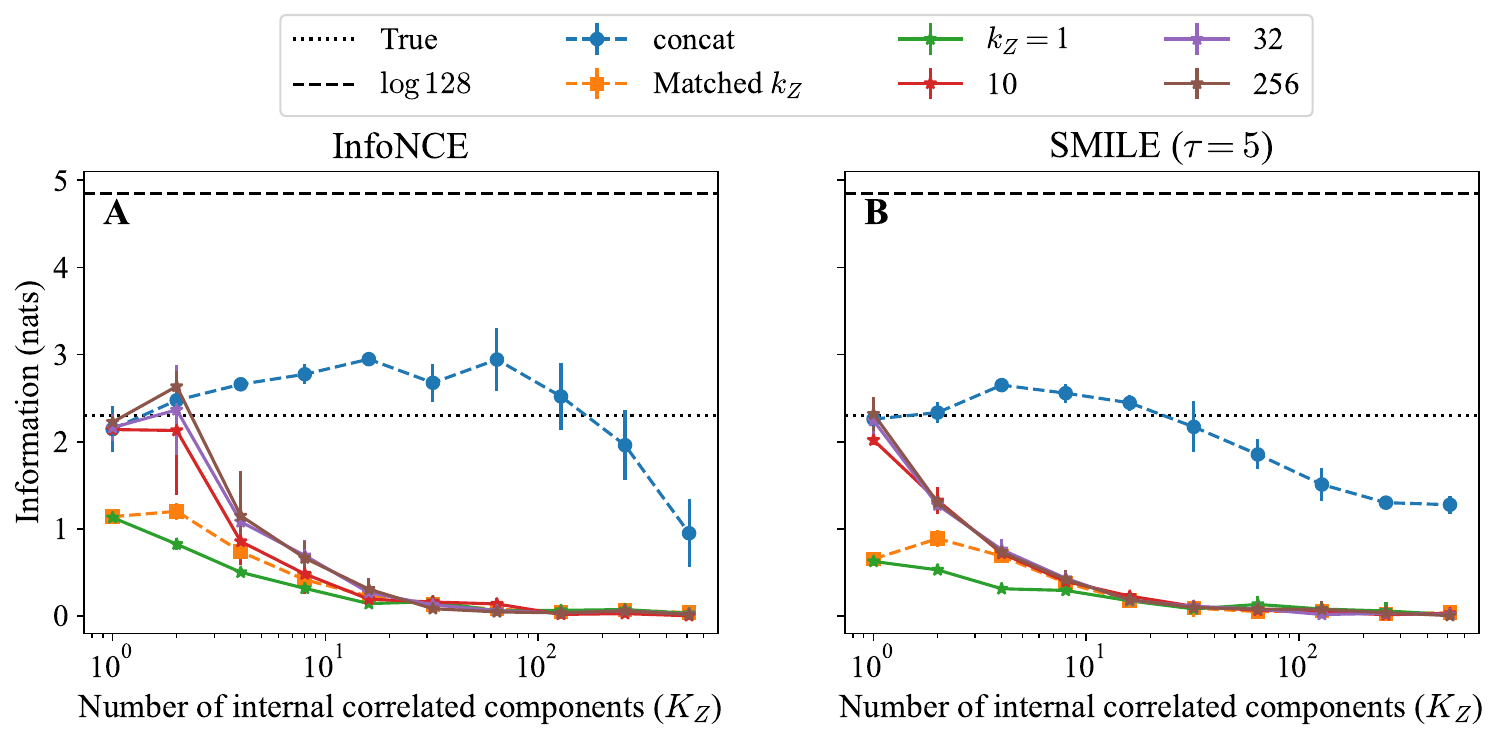}
    \caption{{\bf Large Dimensional Gaussian with Varying Number of Signal Dimensions.} The theoretical value of information, $\log(10)$, is shown as a dotted line, and log the batch size, $\log(128)$, is shown as a dashed line. The plots show the performance of InfoNCE and SMILE $(\tau=5)$ estimators in capturing MI in high-dimensional Gaussian data. We generated two Gaussian variables $X$ and $Y$, each with 784 dimensions, and a fixed amount of information $I(X;Y) = \log{10}$ nats, distributed among a varying number of correlated components $K_Z = 2^0, 2^1, \ldots, 2^9$ on the x-axis. For each setup, we evaluated the estimators using both concatenated and separable critics. The concatenated critic (blue circles) uses a single embedding space, while the separable critic (squares) uses embeddings of different fixed sizes $k_Z = 1, 10, 32, 256$ and a matched size corresponding to the true dimensionality of the data. The plots demonstrate that the InfoNCE estimator effectively captures MI when the information is distributed across fewer than 10 dimensions for the separable critic. However, its performance deteriorates when the information is spread across 32 or more dimensions. Conversely, the concatenated critic shows robustness, with a noticeable decline in performance only at $K_Z \sim 128$ dimensions.}
    \label{fig:sep_and_concat_gaussian}
\end{figure}
The above results demonstrate the success of the estimators in accurately capturing the MI in nonlinear higher-dimensional spaces, for a relatively high number of samplesfootnote{It is hard to clearly conclude in this example if the estimators are unbiased, as we ran out of samples. However, if we are to generate more samples, we should expect asymptotic behavior as $N$ increases.}. We can also see that changing the number of embedding dimensions for the separable critic can be a valuable tool in detecting the intrinsic dimensionality of the system. However, it is not yet clear why these methods work in the first place, given that we are using few samples,  $\mathcal{O}(10^4-10^5)$, for $\mathcal{O}(10^3)$ dimensions. We hypothesize that the reason for the success is that NN MI estimators perform SDR (in two distinct spaces, as for the separable critic, or in one joint space, as for the concatenated critic). We demonstrate that existence of an accurate low-dimensional representation of data is essential for MI estimation by NN methods to work in  Fig.~\ref{fig:sep_and_concat_gaussian}. For this, we generate two Gaussian variables $X$ and $Y$ with 784 dimensions each and a fixed amount of information $I(X;Y)=\log{10}$ nats, to mimic the MNIST dataset. However, this amount of information is distributed among a different number of correlated components in different examples, $K_Z=2^0, 2^1, \ldots, 2^9$. The correlation in each component is chosen such that the total amount of information sums to the same desired $I(X;Y)$. We test this setup at a relatively high number of samples, $N=10^5$, to eliminate the effects of finite sampling and verify whether the estimators can find the same amount of information if distributed among more or fewer intrinsic latent dimensions. Similar to Fig.~\ref{fig:sep_and_concat_mnist}, we evaluate the estimators with concatenated critics, fixed-size embeddings for separable critics $k_Z=1, 10, 32, 256$, and also when matching the embedding size of the critic with the true dimensionality of the data (denoted as Matched, $k_Z=K_Z$). We see that, indeed, for a separable critic, the estimators capture MI effectively if it is distributed in $K_Z < 10$. However, when the information is distributed among 32 dimensions or more, the estimators fail for various embedding sizes. Surprisingly, the concatenated critic starts failing similarly only around $K_Z \sim 128$.  We suspect that this is because the concatenated critic freely mixes the different components of the variables, without imposing a dot product structure in the latent space, as separable critics do, allowing the critic to have more freedom in capturing small amounts of information distributed among a relatively high number of correlated dimensions.

\section{Discussion and MI Estimation Guidelines}
\subsection{Guidelines for MI estimation from high-dimensional data}
\label{guidelines}

The analysis above leads to suggesting the following practical guidelines for estimating MI from high-dimensional data. This is adapted from Ref.~\cite{holmes2019estimation}, using the procedures developed there to test for the self-consistency of the estimators, and hence possible bias.
\begin{enumerate}
    \item Use rCCA; evaluate MI based on the correlation matrix in  $k_Z$-dimensional embedding space; evaluate $I_{\rm EST}(k_Z)$. If saturation is observed at high $k_Z$, report the value. If not, then the data is nonlinear; proceed to the next item.
    \item Partition the data into training/test sets (90\%/10\%). Use a NN estimator of choice for $I_{\rm EST}$. If the amount of information is expected to be low (less than $\ln(\rm{batch size})$, using InfoNCE is preferred. If not, SMILE is usually a good option, but pay attention to the variance. Stop training at maximum of MI on test data.
    \item Vary $k_Z$ and the amount of data analyzed, which can be achieved by subsampling (without replacements, see \cite{holmes2019estimation}) from the full dataset. Perform multiple training runs for each $(k_Z,N)$ pair, varying the subsample and the random seed for training and estimate the standard deviation over the runs, $\sigma_I(k_Z,N)$.
    \item Choose $\hat{k}_Z$ in the range where $I_{\rm EST}(k_Z)$ is stable within $\sigma_I$ (sufficient expressivity, but no undersampling).
    \item If $I_{\rm EST}$ is stable over $k_Z$ and $N$, report $I_{\rm EST}(\hat{k}_Z, N)$ as the MI estimate. Otherwise, no reliable MI estimate has been found.
\end{enumerate}

\subsection{Discussion}
MI is a difficult quantity to estimate, but it is an important quantity to estimate well. New NN estimators are a promising direction, addressing different regimes of the bias/variance tradeoff for the estimation, which were unaccessible with previous methods. However, MI estimation is a hard problem for finite data, and nonlinear data (otherwise use linear methods!), and these estimators have not been tested for these cases. Here we showed that all standard NN based neural estimators fail when the dimensionality of data increases, well within the range of today's experimental datasets. We have shown how one can view the estimators as an SDR technique, reducing the dimensionality of the data and then estimating MI with NN estimators in the reduced space. 
We have also provided heuristics for verifying whether the estimators' output can be trusted. For this, one needs to verify the stability of the output to the hyperparameter (the number of embedding dimensions) and to the amount of data (via varying the number of training samples), at least when the size of the training subsample and the full sample are not drastically different.

Finally, we made the first steps towards explaining how MI estimation---which is provably hard---can be reliable for large-dimensional data, even with relatively small sample sizes. We argue that reliable estimation is only possible when the data admits a low-dimensional latent structure. This is also supported by the argument in Ref.~\cite{martini2024data}  that fluctuations in estimation of mutual information scale in proportion to the size of the latent space, and not the observable space, though this analysis was done for discrete variables only. We hope that additional tests will support this hypothesis, conclusively proving that it is possible to reliably estimate MI in the undersampled regime $ K>N$, as long as the true latent dimensionality of the data is $K_Z\ll N$.

\section{Supplemental Information}
\subsection{Neural networks architecture and Technical details}
The estimators' implementation is adapted from Ref.~\cite{song2019understanding}. The critics are implemented using Multi-Layer Perceptrons (MLPs) with either 2 or 4 hidden layers (in addition to the input and output layers), ReLU activations, and Xavier uniform initialization \cite{glorot2010understanding}. For the separable critic, two networks are trained separately but simultaneously for $X$ and $Y$, with input dimensionalities $K_X$ and $K_Y$, and output dimensionality $k_Z$. For the concatenated critic, a single network is trained with an input dimensionality of $K_X + K_Y$ and an output dimensionality of $1$. The estimators are trained using the Adam optimizer \cite{kingma2014adam} with a learning rate of $5 \times 10^{-4}$. The estimators' implementations are done in PyTorch \cite{paszke2019pytorch} and trained on various GPUs. 
The rCCA models are trained using the library in Ref.~\cite{Wang2021}, where the regularization $c$ factor is fine-tuned to produce the highest test value, and the corresponding train value is reported. Parameters not explicitly mentioned are set to their defaults.

\subsection{Supplemental figures}

\chapter{Discussions}\label{ch5}
Here I aim to summarize the findings of the Dissertation, emphasizing the key insights from each of the preceding Chapters and to outline potential avenues for future research.

The main message of the Dissertation is as follows: In modern scientific data analysis, we often encounter high-dimensional and possibly multimodal datasets, where the number of samples is of the same order of magnitude as the number of dimensions. For such data, the goal is to find an interpretable, modelable low-dimensional representation via application of DR methods. The Dissertation demonstrated that DR is not merely a data preprocessing step, not requiring much thinking. Instead, the structure and the usefulness of the constructed low-dimensional representations of data depend on the DR method used. Thus one should carefully consider which DR method to employ, ensuring that it aligns with the structure of the data and the question we are addressing. For instance, if we seek a low-dimensional representation of a multimodal dataset that captures shared information between modalities (i.~e., covariation), then Simultaneous DR methods will require less data and will produce more useful representations. Even simpler, if the underlying structure of a dataset suggests linearity, linear methods suffice to obtain comprehensive low-dimensional descriptions that capture all relevant information.  Additionally, in this Dissertation, I showed that many DR methods can be systematized within a single generalized framework, the Deep Variational Multivariate Information Bottleneck (DVMIB), which facilitates the translation of dependency graphs for data compression and reconstruction into practical methodology. Furthermore, the development of the new method within this framework, Deep Variational Symmetric Information Bottleneck (DVSIB), holds promise as a valuable tool across various fields. Additionally, the Dissertation demonstrated that low-dimensional representation of data helps in the estimation of important statistics of the data. For example, we showed that we can reliably estimate Mutual Information (MI) in large $K$-dimensional datasets from only $N\sim K$ samples if the data, indeed, has a good low-dimensional latent approximation. This is a major improvement in the field of MI estimation, where, until recently, one expected to need exponentially large datasets to estimate information in high dimensional data.  

The Dissertation opens new venues for subsequent research. Some of these venues are already being explored,  showing promising results. As one of such examples, below I  illustrate the utility of DVSIB  on a simple, but exciting problem: discovering canonical coordinates for a dynamical system from experimental data. Namely, we will use time lapses of images of a physical pendulum to derive the angle and the angular velocity as the two dynamical variables for this system. Secondly, I discuss the utility of DVSIB-like methods in the field of neuroscience, and how they could be a valuable tool in understanding various brain-brain, brain-behavior, and behavior-behavior interactions. With our approaches, we can obtain low-dimensional representations that capture the most covariation between different modalities, allowing for better modeling and understanding of underlying phenomena. 

Below  I summarize the findings of each Chapter and how they relate to each other. Then I follow up with a deeper discussion of potential future research directions. 

\section{On Linear DR Methods}
In Chapter~\ref{ch2}, I presented several key findings regarding linear dimensionality reduction methods and the implications of using different methods for different scenarios. Firstly, I argued that the simple linear model that we studied can capture many realistic features of experimental datasets, along with its flexibility to accommodate more than two modalities of data. We established the superiority of SDR methods over IDR methods when the objective is to capture covariation between modalities rather than mere variation. Supported by empirical evidence and theoretical insights, we strongly advocated for the practical preference for SDR methods, particularly rCCA, in identifying shared signals across different data modalities. This assertion was further reinforced by our exploration of a nonlinear dataset, the noisy MNIST, and corroborated by various sources from the literature, collectively emphasizing the efficiency of SDR methods in extracting covariation. We also addressed the efficacy of SDR techniques in low sampling situations, underscoring the need to align the dimensionality of reduced descriptions with the actual dimensionality of the shared signals. Moreover, we introduced a diagnostic test for differentiating between shared and self signals in data, providing practitioners with a valuable tool for characterizing complex datasets. In summary, Chapter~\ref{ch2} supports what I believe to be a fundamental, albeit often overlooked, principle of data science: SDR methods outperform IDR methods in detecting shared signals. The Chapter also connects to the secondary theme of the entire Dissertation:  in data analysis applications, it is important to match the analysis methods to the underlying structure of the data. Despite certain mentioned limitations, such as linearity of the methods (discussed in Chapter~\ref{ch3}), and linearity of the metric (discussed in Chapter~\ref{ch4}), by focusing on a linear mixing system, Chapter~\ref{ch2} provides an important intuition, which can be extended to nonlinear dimensionality reduction techniques and many practical applications.

\section{On DVMIB}
In Chapter~\ref{ch3}, we introduced a new framework based on MIB principles for deriving variational loss functions tailored for different DR applications. Through this framework, we developed a novel variational method called DVSIB, which compresses variables $X$ and $Y$ into latent variables $Z_X$ and $Z_Y$ respectively, while maximizing the mutual information between $Z_X$ and $Z_Y$. Notably, DVSIB produces two distinct latent spaces, a feature highly desirable in various applications, while achieving superior data efficiency compared to existing methods in terms of classification accuracy. By illustrating the derivation process of variational bounds for terms common to all examined DR methods, we offer a comprehensive library of typical terms in Appendix~\ref{App:Library}, which can serve as a reference for deriving additional DR techniques. Moreover, we (re)-derived several prominent DR methods, including $\beta$-VAE, DVIB, DVCCA, etc., showcasing the versatility and applicability of our framework. Through implementation and evaluation, we demonstrated that methods aligning more closely with the structure of dependencies in the underlying data tend to yield more useful latent spaces, as evidenced by dimensionality and reconstruction accuracy metrics. Notably, SDR methods like DVSIB and DVSIB-private emerged as top performers, offering separate latent spaces for $X$ and $Y$ and enabling the capture of additional information beyond shared labels with a sample efficiency better than in competing methods. This quality -- being more sample efficient \cite{Sarstedt2011,Thompson2006,Thompson2012,Shanechi2021} -- previously mentioned in the literature with no clear understanding and conflicting arguments \cite{Lewis2013} is now confirmed and understood better. Furthermore, our framework extends beyond variational approaches, allowing for the implementation of deterministic models, such as autoencoders. By leveraging specialized encoder and decoder networks, our framework can accommodate various constraints and symmetries, thus potentially serving as a versatile toolkit for a wide range of DR methods. With the availability of tools and code \citep{githubCode}, I aim to facilitate the adoption of this approach across diverse problem domains.

\section{On Efficient Estimation of Mutual Information}
Chapter~\ref{ch4} investigated the estimation of mutual information within the lens of SDR approaches discussed in Chapters~\ref{ch2} and~\ref{ch3}. The MI estimation is challenging, with no universally good estimators for finite data sets, necessitating the use of assumptions. Traditional methods in the literature are limited in their application to modern experimental data, as they cannot effectively handle high dimensionality (they work well up to $\mathcal{O}(10)$ dimensions, whereas many modern experimental setups generate data with  $\mathcal{O}(10^3)$ dimensions or more). Neural Networks (NN) based methods were introduced to mitigate such issues, and they offer promising results. However, the literature lacks rigorous testing and evaluation of NN based estimation methods. For instance, ML methods are typically tested on low-dimensional Gaussian data with effectively infinite samples, a scenario that eliminates overfitting but does not reflect real-world research problems with limited samples and high dimensionality. Nevertheless, we demonstrated that, if the data are truly linear, simple estimation based on empirical correlation matrices suffices; otherwise, more complicated methods are necessary. Specifically, in nonlinear cases, appropriate nonlinear methods are required. If the data is linearly embedded in a high-dimensional space, suitable DR methods (like rCCA, as shown in Chapter~\ref{ch2}) can embed it into a lower-dimensional space, where MI can be estimated, again, based on empirical correlation matrices. However, selecting the number of latent dimensions is crucial in this case to avoid overfitting. In cases where the data is linear but embedded nonlinearly (e.~g., with a frozen neural network), even the best linear DR methods, such as rCCA, are insufficient, as determining the correct dimensionality becomes challenging. Already existing ML methods are suitable for effectively infinite data scenarios without overfitting concerns. However, for more realistic finite data scenarios, addressing overfitting and when to stop training, is necessary. We viewed existing ML methods as SDR approaches, which simultaneously embed large-dimensional data into smaller dimensional latent spaces, where NN based methods can be used to estimate MI with high accuracy.  We developed various consistency checks and heuristics to determine the reliability of the results. This Chapter aimed to demystify such methods and provide a practical guide for their usage in realistic scenarios. It also highlights the utility of SDR approaches in general. As it turns out, successful MI estimators could be viewed as SDR methods as well.

\section{On Discovering Coordinates of Dynamics}
\footnote{This section is in part based on an ongoing work with K.\ Michael Martini and Ilya Nemenman. The development of the idea was a collaborative effort among all three authors. The code was written by K.\ Michael Martini and Myself. The figures presented in this section are produced by myself. The text was written by myself and jointly revised by all authors.}So far,  previous Chapters, largely revolved around static datasets, where the relationship between modalities is captured within each $(X,Y)$ pair. We have not yet considered dynamics, which is where many interesting physics problems lie. In fact,  dynamical data can be cast in a manner consistent with our multiview setup. For example, consider a dynamical system, where $X$ represents the current observations and $Y$ denotes the future ones \cite{Tishby2009}. In this context, past and future could be symmetrically defined as a fixed number of past and future snapshots of the system state \cite{Lipson2022}. Or, similar to a predictive information definition \cite{bialek2001predictability, bialek1999predictive}, we can define the future as consisting of an infinite number of snapshots (in practice, it suffices to use a large number, larger than the autocorrelation time), and the past would then be described by a long sequence of snapshots of the system state. Here then the shared representation between the past and the future is some generalized coordinates of the system's dynamics. 

\subsection{The Setup}
\begin{figure}[h]
    \centering
    \includegraphics[width=\textwidth]{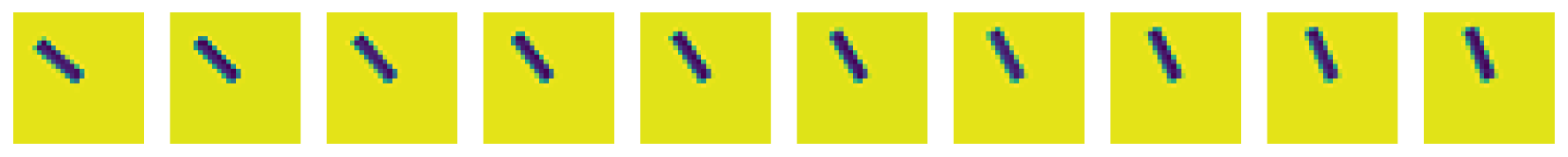}
    \caption{Individual frames from a physical single pendulum's motion. The dataset and more information can be found in \cite{Lipson2022}. Each image represents a frame captured during the pendulum's real motion. The sampling rate is $60$ Hz, and each frame is $28 \times 28$ pixels, with a total of $60$ frames per experiment. Each experiment has different initial conditions, with a total of $1200$ experiments. The pendulum is of mass of $1$ kg, and a length of $0.5$ m.}
    \label{fig:pendulum_frames}
\end{figure}

\begin{figure}[ht]
    \centering
    \includegraphics[width=\textwidth]{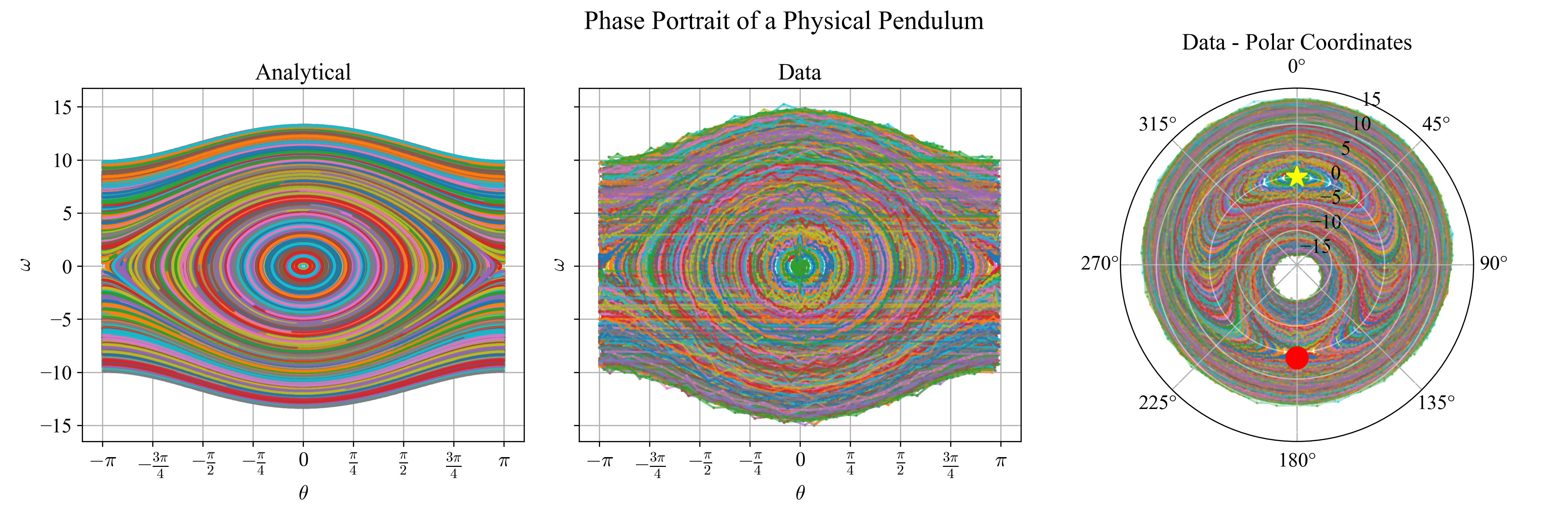}
    \caption{Phase space portraits of a physical pendulum for $\omega$ vs $\theta$. The left subplot shows the exact phase space calculated analytically from the differential equations of a physical pendulum with the same properties as the pendulum in Fig.~\ref{fig:pendulum_frames}, with different initial conditions, then wrapped between $-\pi$ and $\pi$. Each trajectory is for a new experiment with different initial conditions, and the color scheme is arbitrary. The middle subplot is for the actual physical pendulum with data obtained from \cite{Lipson2022} and shown in Fig.~\ref{fig:pendulum_frames}, where the physical quantities $\theta$ and $\omega$ are obtained separately from the videos by different analysis, in which the angles and their derivatives are calculated by tracking the pendulum using computer vision tools, as described in \cite{Lipson2022}. The right subplot shows the same data but in polar coordinates. The yellow star is the stable fixed point, while the red circle is the unstable one.}
    \label{fig:pendulum_phase}
\end{figure}

To illustrate this, let us take a simple physical pendulum as shown in Fig.~\ref{fig:pendulum_frames}. 
Its dynamics are perfectly described by the angle and angular velocity at any given time $t$. The phase portrait of $\omega$ vs $\theta$ is shown in Fig.~\ref{fig:pendulum_phase}\footnote{The direction arrows are suppressed for clarity, in this figure and the subsequent figures as well.}. The left subplot shows the exact phase space calculated analytically from the differential equations of a physical pendulum with the same properties as the pendulum in Fig.~\ref{fig:pendulum_frames}, with different starting conditions and centered at zero to be between $-\pi$ and $\pi$. The middle subplot is for the actual physical pendulum with data obtained from \cite{Lipson2022}. The right subplot shows the same data but in polar coordinates, which will become relevant later. The phase space of a pendulum is equivalent to an infinite cylinder due to the periodic nature of the angle $\theta$ and the unbounded nature of the angular velocity $\omega$. $\theta$ wraps around to form the circumference of the cylinder, while $\omega$ extends along its length. One can project the cylinder onto a plane using polar coordinates $theta$ and $r$ as follows:
\begin{align}
    \theta &\rightarrow \theta \nonumber\\
    \omega + \text{offset (or: offset} -\omega) &\rightarrow r,
\end{align}
with the offset larger than the maximum angular velocity in the experiments, so that the radius $r$ is positive. The offset arises from compressing the negative range of $\omega$ into a small region, creating a hole in the polar plot. Additionally, the system has two fixed points: a stable fixed point at $(\theta, \omega) = (0, 0)$, where trajectories are periodic, and an unstable fixed point at $(\theta, \omega) = (\pi, 0)$ (or ($-\pi, 0$)), where trajectories diverge. A proper embedding from any algorithm should reflect these topological characteristics: the hole due to the cylindrical structure of the phase space, one stable fixed point, and one unstable fixed point.

\subsection{DVSIB For Dynamics}
Using insights from previous chapters, we leverage a DVSIB-like structure to compress the movie data in the hope to discover generalized coordinates of the physical pendulum. In this setup, we choose $X$ to be two consecutive ``past'' frames at time steps $t$ and $t+1$, while $Y$ then represents the ``future'' with frames at time steps $t+2$ and $t+3$. Our goal is to embed these data into  $Z_X$ and $Z_Y$, maximizing the mutual information between the latter. We expect the low-dimensional embeddings to represent interpretable generalized coordinates for this system. Since we do not explicitly need the reconstruction of the movie images from the latent variables in this task, we can turn off the reconstruction term \footnote{Reconstruction adds complexity in terms of compute and data. This is because we are learning distributions in the form of $p(X|Z)$, which has the dimensionality of $X$. Unless needed for generative tasks, or to stabilize compressed variables and avoid representation collapse, reconstruction should be minimized or turned off. Representation collapse occurs when $Z_X$ and $Z_Y$ become the same variable, or independent of the data. This phenomenon, leading to multiple independent trivial solutions, is discussed in \cite{martini2024data}. The washout problem (the degradation of information as it propagates through the layers of a neural network) is also mentioned in \cite{Lipson2022} as a known issue in various ML algorithms.}. Additionally, we want to impose time translation symmetry in our reduction process, in the sense that the algorithm learns we go from observables (the frames formed into $X$ and $Y$) to the latent variables ($Z_X$ and $Z_Y$) in the same way. Thus we employ the same encoder network for both the past and the future frames, as it forces the past encoder to learn the same compression that the future encoder does (i.e, $p(Z_X|X) = p(Z_Y|Y)$). The resultant loss function is: 
\begin{equation}
    \label{sib_dynamics}
    L=I^E(X;Z_X)+I^E(Y;Z_Y)-\beta I^D(Z_X;Z_Y),
\end{equation} where $I^E(X;Z_X)$ and $I^E(Y;Z_Y)$ are approximated with the same neural network.

\subsection{Preliminary Results}
\begin{figure}[ht]
  \centering
  \includegraphics[width=\textwidth]{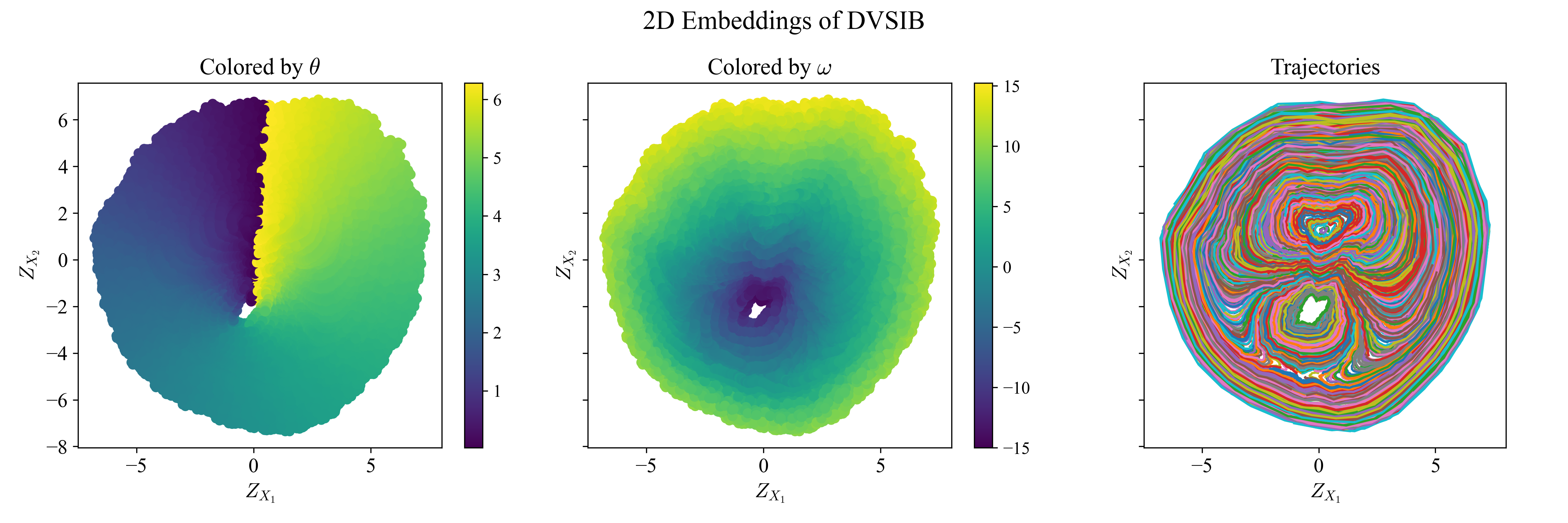}
  \caption{Left: Embeddings $Z_X$ in 2D colored by $\theta$. The gradient of angles within the manifold along the ``angular" direction is clearly visible, with points at 0 and $2\pi$ connected, indicating a continuous representation. Middle: Embeddings $Z_X$ in 2D colored by $\omega$. A gradient along the ``second" dimension of the manifold is observed, indicating recovery of the angular velocity $\omega$. Right: The same embeddings, but colored as individual trajectories. We observe that these embeddings represent the phase space of the pendulum in polar coordinates, as shown in Fig.~\ref{fig:pendulum_phase}, subject to an arbitrary rotation in $\theta$ and a shift in $\omega$. The embeddings are results of training a simple implementation of \ref{sib_dynamics} with frames of experiments of single physical pendulum obtained from \cite{Lipson2022}, the $I(Z_X;Z_Y)$ term is trained as part of DVSIB with SMILE $(\tau=5)$ estimator with a concatenated critic (the networks architecture is described in detail in SI~\ref{app:nn_archs}).}
  \label{fig:ZX_2d}
\end{figure}

\begin{figure}[ht]
  \centering
  \includegraphics[width=\textwidth]{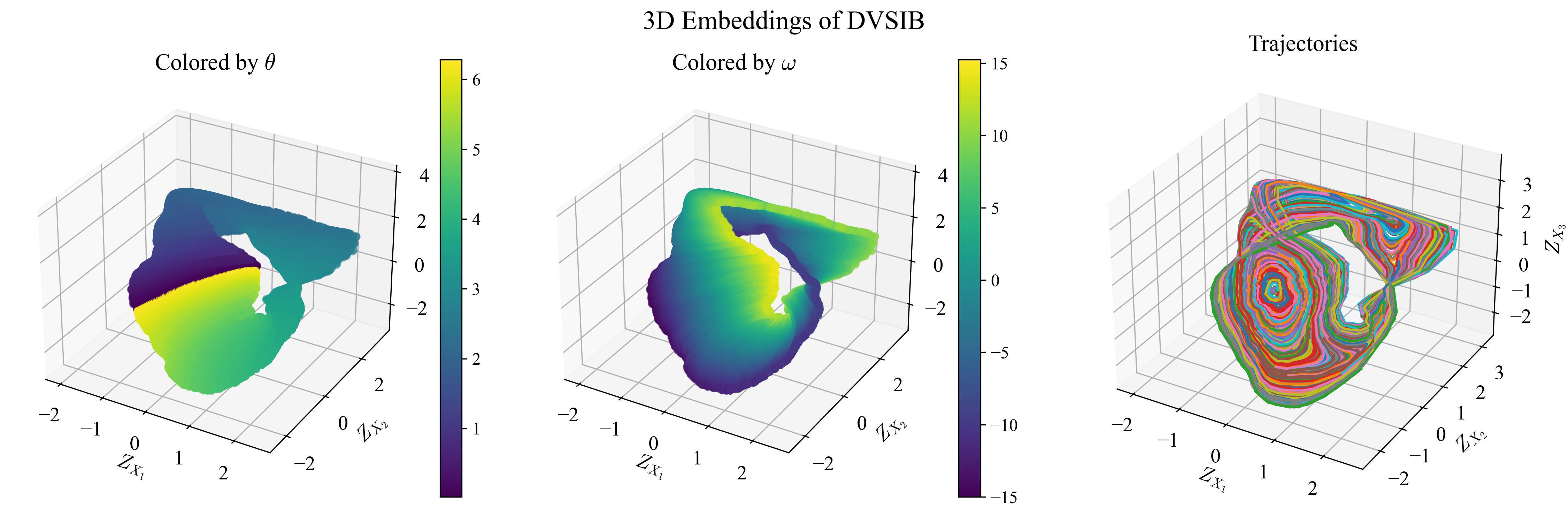}
  \caption{Left: Embeddings $Z_X$ in 3D colored by $\theta$. The gradient of angles within the manifold along the ``angular" direction is visible, with points at 0 and $2\pi$ connected, indicating a continuous representation. Middle: Embeddings $Z_X$ in 3D colored by $\omega$. A gradient along the ``second" dimension of the manifold is observed, indicating recovery of the angular velocity $\omega$, albeit with a twist near the unstable fixed point. Right: The same embeddings, but colored as individual trajectories. We observe that these embeddings represent closed trajectories around the stable fixed point, albeit with an ambiguity near the unstable one. Different training parameters (different number of neurons and hidden layers in the critic) could lead to different embeddings, some of which are shown in the SI Fig.~\ref{app:3d_embeddings}. The embeddings are results of training a simple implementation of Eq.~\ref{sib_dynamics} with frames of experiments of single physical pendulum obtained from \cite{Lipson2022}, the $I(Z_X;Z_Y)$ term is trained as part of DVSIB with SMILE $(\tau=5)$ estimator with a concatenated critic (the networks architecture is described in SI~\ref{app:nn_archs}).}
  \label{fig:ZX_3d}
\end{figure}

Figure~\ref{fig:ZX_2d} illustrates the preliminary results obtained from the training (additional details about the architecture of the networks and training procedure are in the SI~\ref{app:nn_archs}). To illustrate that the method learned the underlying physics, we directly input the past and future frames into our model, and we require our model to give us two dimensional $Z_X$ and two dimensional $Z_Y$ (when we are designing the networks, we choose the size of the low dimensional spaces --among other parameters--, 3d embeddings are shown next in Fig.~\ref{fig:ZX_3d}).Then we plot the resulting embeddings $Z_X$ (or $Z_Y$), colored by the values of various quantities, which we know to be important in this problem. Specifically, we know that the two key variables governing this system are the angle $\theta$ and the angular velocity $\omega$. The values of $\theta$ and $\omega$ are obtained separately through another procedure described in the paper, from which we obtained the datasets \cite{Lipson2022}. The embeddings we obtain are nonlinear manifolds in two dimensions. 

Interestingly, our embeddings successfully capture the dynamically relevant variables within this 2D space as shown in Fig.~\ref{fig:ZX_2d}. When we color the embeddings $Z_X$ (and similarly for $Z_Y$, due to the symmetric compression, $Z_Y$ is the same as $Z_X$ shifted by two frames) based on $\theta$ (left), a clear gradient of angles becomes evident within our manifold along the ``angular'' direction. Moreover, the points at $0$ and $2\pi$ in the angle space are connected, indicating a continuous representation. Additionally, when we color the embeddings by $\omega$ (middle), we observe a gradient along the ``second'' dimension of the manifold, revealing that along this second nonlinear embedding, we recover the angular velocity $\omega$. More importantly, when we color the embeddings as individual trajectories (right), we recover a topologically correct phase space of $\omega$ vs $\theta$ in polar coordinates, which was suggested from the coloring by $\theta$ or $\omega$ individually that $\theta$ is encoded in the angular direction, and $\omega$ is encoded in the radial one. The resemblance between the topology of the results in Fig.~\ref{fig:ZX_2d} and the polar representation of the true phase space in Fig.~\ref{fig:pendulum_phase} is evident. We recover the hole due to the parametrization of $\omega$ in the radial direction, the two fixed points in the right place, and the nearly circular trajectories near the stable fixed points. The model, indeed, learned that the relevant features between the frames of the past and the future are generalized coordinates that resembles the true $\theta$ and $\omega$ of the pendulum.

Even if we increase the dimensionality of $Z_X$ and $Z_Y$ as in Figure~\ref{fig:ZX_3d}, the model continues to learn a deformed 2D manifold (as shown in the last subplot). This is shown in the continuity of the angular direction, the gradient in the radial direction, the fixed points in the phase space, and the hole in it --albeit not as clear as in the 2d situation, probably due to suboptimal training. These observations indicate that although the system is observed in high dimensionality, its latent variables are 2d, even when allowed to occupy 3d. While one may have expected this knowing the second-order nature of Newton's laws and the fact that we only used 2 movie frames to define both the past and the future, the dimensionality of the data is $\sim10^3$, so the fact that the DVSIB for dynamics architecture detects the 2d structure is nontrivial. 

It's important to note that our approach is an ``out-of-the-box'' application of the method, and we do not impose constraints on the embeddings. And while we could end up with a twist or a collapse in our embeddings (as shown in 3d in Fig.~\ref{fig:ZX_3d}, and while other training parameters could lead to the same behaviour in 2d as well--not shown), the `twists' in the manifold appear to be points of ambiguity near the unstable fixed points (as we do not see that along the $\theta$ direction), where the model is unable to resolve the changes from positive to negative $\omega$. Nevertheless, even with this twist, Figure~\ref{fig:ZX_3d} still showcases a clear separation of the relevant dynamical variables that the model has learned directly from the experimental videos, without any preprocessing or further embedding. A possible second step would be learning the dynamics, symbolically, in this low dimensional space, characterized by a handful of dimensions rather than the apparent high dimensionality of the videos. There are multiple methods in the literature one can use for learning such differential equations \cite{brunton2016discovering, daniels2015automated, champion2019data}. 

\subsection{Additional Questions and Remarks}
While the preliminary results for using DVSIB for dynamics are promising, indicating that we can recover the system's latent variables directly from observations with minimal additional constraints, there are still relevant questions that need further exploration.

\begin{figure}[ht]
    \centering
    \includegraphics[width=0.7\textwidth]{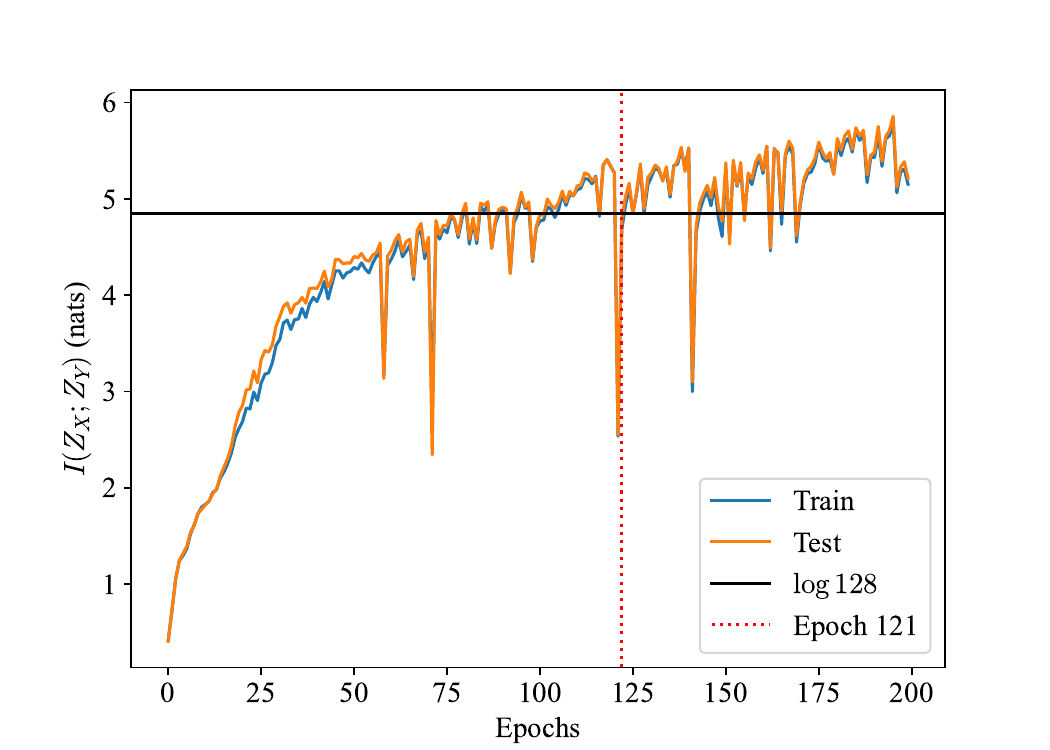}
    \caption{Information between embeddings of the past and the future for the pendulum dataset, $I(Z_X;Z_Y)$, during the training. The training curve evaluates a subset of the frames used during the training (100 experiments out of 1000 used for training, each experiment having 60 frames) and another subset for the test that was not seen during the training (100 other experiments with 60 frames each). The training and test curves are almost on top of each other. We notice fluctuations in the training that decrease the amount of information severely. These fluctuations are often observed in the training embeddings, as seen in Fig.~\ref{training_epochs} for epoch 127, where the embedding changes significantly, likely indicating an optimizer-related phenomenon (often called loss spikes\protect\footnotemark~ \cite{zhu2023catapults,zhang2023loss}) that the training encountered.}
    \label{fig:pendulum_3d_training}
\end{figure}

\footnotetext{Loss spikes are an abrupt increase --or decrease in the case of MI-- in the loss function value, typically occurring due to the network encountering regions in the loss landscape with sharp gradients or suboptimal local minima. These spikes often arise from instabilities in the optimization process, where the training dynamics momentarily lead the model into unfavorable regions, causing the loss to increase before stabilizing again as the model continues to learn and adapt.}

\begin{figure}[ht]
    \centering
    \includegraphics[width=\textwidth, trim={4.7in 0.7in 4in 1in}, clip]{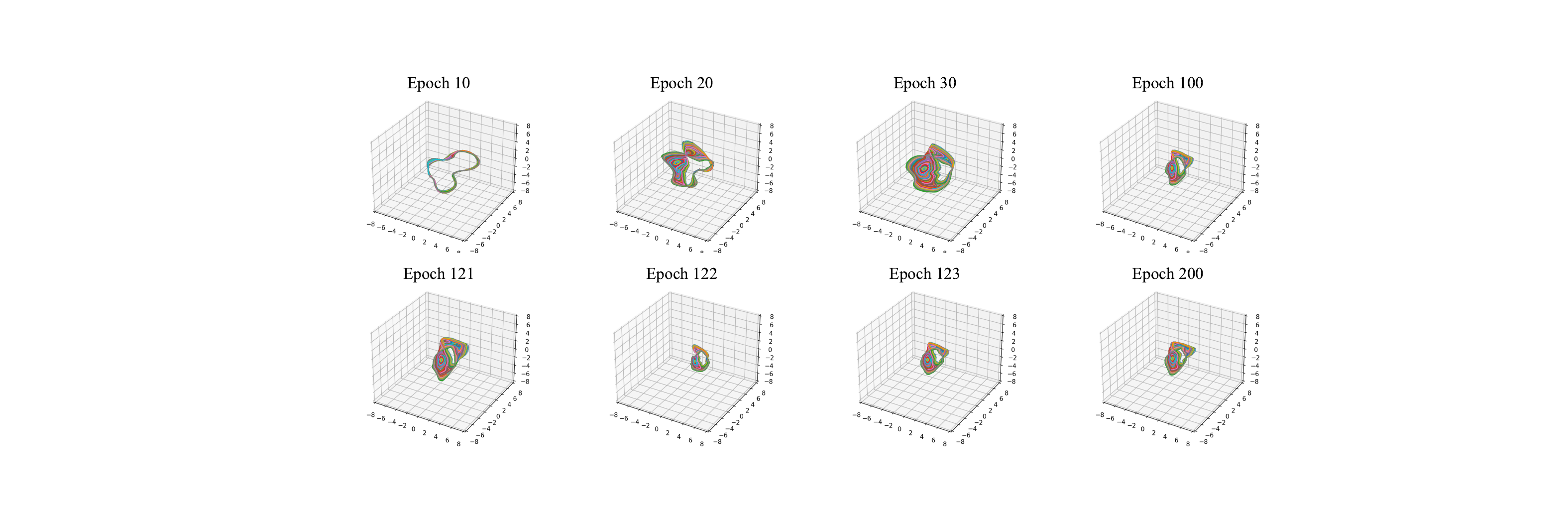}
    \caption{Embeddings of 100 experiments evaluated at specific training epochs. We observe the evolution of the embeddings from a straight line to the final structure obtained at the end. Notably, after 100 epochs, the training becomes relatively stable, as also reflected in Fig.~\ref{fig:pendulum_3d_training}, with fluctuations that can cause the embeddings to change within their space, as shown in epoch 127, for example.}
    \label{training_epochs}
\end{figure}

\textbf{When to stop training?} As discussed in Chapter~\ref{ch4}, the MI estimators (SMILE with $\tau=5$ in this case) can overfit during training, and we have to make sure that we are reporting the true information, not just an arbitrary number produced by the model during its training. However, when we examine the training curves to obtain the embeddings in Fig.~\ref{fig:ZX_3d}, for example, we notice that the model does not overfit. This is evident from the curves of $I(Z_X;Z_Y)$ for the train and test datasets in Fig.~\ref{fig:pendulum_3d_training} being on top of each other\footnote{At some points during the training, the test curve is slightly higher than the training curve, which is primarily due to the finite data set size used in evaluation, which produces fluctuations. During training, the model uses mini-batches of 128 frames each. However, the evaluation of MI is performed on 100 experiments, each with 60 frames, resulting in 6000 frames for both training and testing to be loaded onto the GPU simultaneously. This creates a significant computational overhead. While it is possible to increase this number to increase the accuracy, it would require considerably larger computational resources than we have at our disposal.}, which could be attributed to the simplicity of the image we are trying to compress---it is one body on a static background that moves slightly between each frame and the subsequent one. However, we observe fluctuations in the training curves that could be attributed to getting stuck in local minima, as seen in Fig.~\ref{training_epochs}, where the embeddings at epoch 127 moved in space. Thus, additional investigation is required to explore these fluctuations, and whether a gentler optimization with a smaller (or adaptive) learning rate, for example, would alleviate this issue. We think that the approach to choose the stopping condition for this problem is by considering some metric of consistency and saturation of the training curves. For example, \cite{schneider2023learnable} considers the correlations between different embeddings (for different experiments, for example) as a sign of good training, which could be useful in this problem as well.

\textbf{The interpretation of mutual information:} MI calculated between the past and future embeddings of the system could be a good proxy for the success of the system in recovering the relevant variables. In this problem, an analytical calculation of the MI between the sets of frames in the past and the future would provide a theoretical bound to check if the model actually learned the underlying dynamics. There are known aspects of this problem from physics or experiments, such as the sampling rate (i.~e., the time between frames; if this time interval approaches zero, becoming truly continuous, the mutual information should become infinite), as well as the length and weight of the pendulum, and the governing equations of its motion. However, it is not straightforward to calculate the MI between multiple frames of the past and the future, which will require further work. While analytically calculating the MI might not be possible for scenarios where we do not know the answer ahead of time, it would be beneficial in establishing benchmark problems to assert the utility of DVSIB for inferring dynamics (or identifying its shortcomings). Establishing the utility of MI as a measure of goodness of the training would be valuable not only for the general training of DVSIB for dynamics, but also for addressing additional questions as the following.

\textbf{How many frames to use for $X$ (and $Y$)?} The previous results (Figs~\ref{fig:ZX_2d},~\ref{fig:ZX_3d}) were obtained for two frames for both the past and the future, $X$ and $Y$. Since we know that the dynamics of the pendulum are captured within the variable $\theta$ and its first derivative, $\dot{\theta} = \omega$, and the derivative is calculated from two time points, we should expect that increasing the number of frames would not increase the MI by much if the finite difference between two time points is a good approximation for the derivative. Thus, we would expect to see a saturation in the measurement of MI with the addition of more frames. We should be able to see such saturation in our data, but we leave this analysis for future work.

\textbf{How many dimensions for $Z_X$ (and $Z_Y$)?} We have already explored the cases for 2D and 3D low-dimensional embeddings in Figs.~\ref{fig:ZX_2d} and~\ref{fig:ZX_3d}. We observed that 2D is sufficient to replicate the true phase space of the pendulum. In 3D, the embeddings exhibited similar topological features to the 2D case, though with a twist in the embeddings. With more careful training, it might be possible to achieve a clear 2D embedding within the 3D space as well. This is because we have, in theory, only two latent variables $\theta$ and $\omega$. Thus, adding more possible dimensions would not add more dynamical variables, and we should expect a saturation in the measurement of MI with the addition of more dimensions. However, the extra dimensions might allow different parameterizations for the latent variables. We leave the exploration of the interplay between the extra embedding dimensions, the ease of training, and detecting more useful parameterization of the latent space to future work.

\textbf{Finally:} By customizing DVSIB to match the structure of the problem and to consider explicitly the time translation symmetry, the method was able to learn the proper coordinates for the underlying dynamics directly from observations (movies) without preprocessing or {\em ad hoc} assumptions.  While this approach is still under development, we believe that its potential for use in many future applications is vast.

\subsection{Supplementary Information}
\subsubsection{Neural networks architecture used for dynamical inference}\label{app:nn_archs}

The loss function Eq.~(\ref{sib_dynamics}) is implemented as follows:
\begin{itemize}
    \item $I^E(X;Z_X)$ is implemented as a fully connected feed-forward neural network. The network has an input layer of 1568 neurons (each frame is 28x28 pixels, and we have two frames flattened and stacked together), two hidden layers with 1024 neurons each, and two nodes in the output layer that correspond to the means and variances of a gaussian distribution that we learned. Each node is of the size of the dimensionality of $Z_X$which is 2 for Fig.~\ref{fig:ZX_2d} and 3 for Fig.~\ref{fig:ZX_3d}. $I^E(Y;Z_Y)$ is implemented in a similar manner.
    \item The MI estimator network is a concatenated critic that takes both $Z_X$ and $Z_Y$ embeddings, stacks them together, and then passes them through a fully connected feed-forward neural network with one hidden layer of 32 neurons and an output layer of size 1. The outputs for different inputs in the final layer are calculated based on the SMILE approximation to MI (cf.~Chapter~\ref{ch4}).
    \item All layers in all networks are initialized with Xavier Uniform initialization\footnote{Xavier uniform initialization is a method used to initialize the weights of neural networks by drawing them from a uniform distribution. The weights are sampled from a range of $[-\sqrt{6/(n_{\rm in} + n_{\rm out})}, \sqrt{6/(n_{\rm in} + n_{\rm out})}]$, where $n_{\rm in}$ and $n_{\rm out}$ represent the number of input and output units, respectively. This initialization helps to maintain the variance of activations and gradients across layers, allowing for stable and efficient training.
} \cite{glorot2010understanding}.
    \item The networks are trained with the ADAM optimizer \cite{kingma2014adam} with a learning rate of $5 \times 10^{-5}$, and $\beta = 256$.
\end{itemize}

\subsubsection{Additional embeddings at different parameters }\label{app:3d_embeddings}

\begin{figure}[ht]
    \centering
    \includegraphics[width=\textwidth, trim={4in 0.7in 3.5in 0in}, clip]{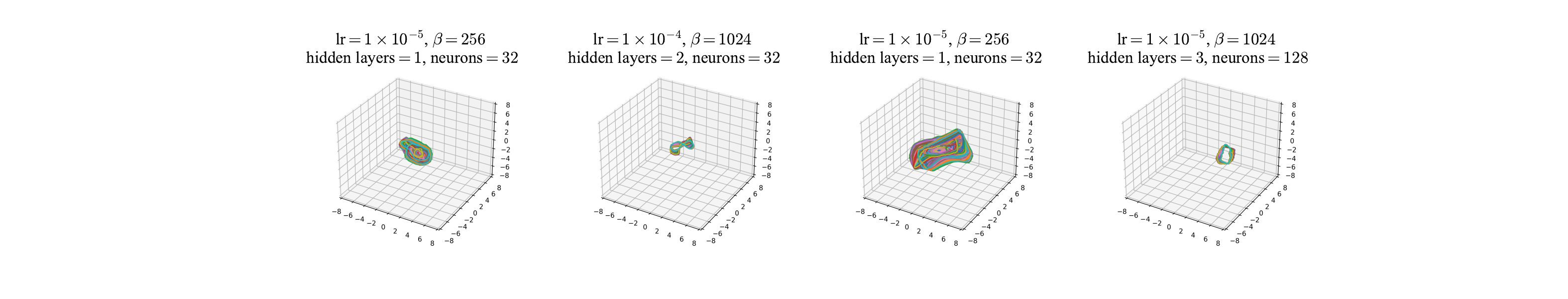}
    \caption{Embeddings of training experiments evaluated at the 200th training epoch for different parameters of learning rate for the optimizer, $\beta$ for the DVSIB loss, number of hidden layers, and number of neurons in those hidden layers. We observe different behaviors, suggesting the need for a careful training. We can see that all of them developed some form of confined manifold. This manifold is regular in some situations and irregular in others.} \label{fig:pendulum_3d_training_epochs_appendix}
\end{figure}

The embeddings change as the parameters ($\beta$, number of hidden layers in the MI critic, number of neurons in these layers, learning rate, etc.) change. The following plots show different embeddings after training for 200 epochs and using the trained encoders to obtain the embeddings of all the training experiments in 3d space. Alternative embeddings evaluated for other parameter values are shown in Fig.~\ref{fig:pendulum_3d_training_epochs_appendix}.

\section{On Neural Activity and Behaviour}
Understanding the mutual interplay between neural activity and behavior is the Holy Grail for deciphering brain-body interactions. Recent technological advancements enable simultaneous recordings of activity from tens of thousands of individual neurons across various brain regions with exceptional temporal and spatial precision \cite{Harris2019, Harris2021, Churchland2022}. Simultaneously, there is a growing effort in theoretical research to develop decoders capable of translating neural signals into behavior \cite{Fairhall2015, Fairhall2016, Shahbaba2017, Ganguly2021, Poeppel2017, pandarinath2018inferring}. These decoders often reveal that a low-dimensional representation of neural data is sufficient for decoding behavior, shedding light on how the brain integrates signals from different regions to execute specific tasks. This alignment with low-dimensional representations suggests the existence of population variables within neural activity that correspond to behavior \cite{perich2020motor, gallego2017neural, williamson2019bridging}. Besides facilitating data interpretation, these representations may lay the groundwork for theoretical frameworks describing the underlying dynamics of both neural activity and resulting behavior, which directly impact how we understand different diseases and how to mitigate their effects with prosthetic devices  \cite{artemiadis2010emg, hu2018decoding}, or drugs \cite{cao2006abnormal, hooten2016chronic, appelbaum2023synaptic}.

However, despite considerable progress, a comprehensive theoretical framework describing the relationship between neural activity and behavior across diverse experimental conditions remains elusive. Different DR methods are at the core of the ongoing research. Traditional DR methods are tailored to specific problem classes, primarily focusing on reducing variability in neural activity without considering its covariation with behavior. We have shown in synthetic and real world examples that such approaches may overlook dimensions significant for describing covariation, rendering them unsuitable for this purpose. 

A more effective approach to address this covariation is by considering the neural activity and behavior simultaneously. There are few examples in the neuroscience literature that consider behavior while reducing neural activity. Such methods typically compress only the recorded neural activity conditioned on the behavior or presented stimuli \cite{Shanechi2021,hurwitz2021targeted,schneider2023learnable}. In the language of Chapter~\ref{ch3}, and with $X$ being the recorded neural activity and $Y$ being the resultant observed behavior, we are dealing within an Information Bottleneck framework, compressing $X$ to $Z$, while $Z$ is maximally informative about $Y$. Such an approach might be useful if we are considering a low-dimensional behavior $Y$. This could be, for example, a left-right motion on a treadmill or a mechanical arm that can be moved in only a few directions \cite{schneider2023learnable,Ganguly2021}. However, once the behavior or stimulus is high-dimensional, (for example, recorded videos of motion or a high-dimensional visual stimulus), a reduction of the behavior $Y$ is needed as well. Such a reduction is usually done separately, by means of extracting useful information based on prior knowledge. For example, \cite{mathis2018deeplabcut} extracts and tracks joint positions when considering recorded motion as the behavior. Similarly, \cite{caron2021emerging} extracts important features from videos often used as visual stimuli. However, such separate independent reduction makes us question the validity of the approach, as it might overlook relevant features of covariation due to their low variation (and vice versa).

However, there are a few exceptions that compress both the behavior and neural activity simultaneously during the reduction. For example, the method in  Ref.~\cite{gondur2023multi}, which can also be cast in Chapter~\ref{ch3} language as Deep Variational CCA with private information, compresses both. In such a setup, $X$, the neural activity is reduced to a shared part, $Z$, and a private part $W_X$, and both $Z$ and $W_X$ are important to recover the full behavior $X$. Similarly, $Y$ is reduced to the shared part $Z$, and a private part $W_Y$. Then, $Z$ is the relevant mixed space of both that can be used for further study. However, while this last example is performing simultaneous reduction, it produces three spaces, one for each of the two modalities that is relevant to them uniquely, and a shared one with mixed units that is relevant for both, making it unclear how to measure and interpret the embeddings as neural activity and/or behaviour in such a mixed latent space. DVSIB, on the other hand, is the first method (to our knowledge) that allows for two distinct latent spaces for the two different modalities $X$ and $Y$, with different properties, yet maximally informative. The new avenues opened by such approaches are interesting and worth studying. For instance, DVSIB and related techniques could be used as supervised dimensionality reduction instead of regular methods like PCA which most practitioners use. Or it can be used to align recorded activity in one part of the brain with other parts of the brain. In this case, a perfect alignment should correspond to maximum mutual information.

\section{Final Thoughts}
In this Dissertation, I  aimed to clarify the concept of dimensionality reduction, illustrating that a specific method chosen for it matters a lot more than a mere preprocessing step before data modeling. I explored the overarching principle that, to address a data-driven research question effectively using DR methods, we must align the structure of our methodology with the essence of the inquiry.

DR requires careful method selection and an understanding of expected outcomes. Our results show that, when we want to find the shared information among multiple data sources, simultaneous DR is the preferred approach. We provided extensive discussions, designed novel tools, introduced new heuristics and consistency checks to support this methodology. Personally, I find a great appeal in the idea of obtaining low-dimensional descriptions of complex systems, recognizing that the underlying simplicity often lies within the apparent complexity. I envision this Dissertation as a step towards providing tools and insights for physicists engaged in the timeless pursuit of uncovering simplicity amidst complexity. While many physicists are rightfully fascinated by the maxim that ``more is different,'' \cite{anderson1972more} perhaps we can similarly embrace the idea that more can also be simple.
% \input{chapters/ch5}
%\input{chapters/ch6}
% \appendix
% \input{chapters/appendix}

% BIBLIOGRAPHY
\backmatter
\addcontentsline{toc}{chapter}{Bibliography}
\bibliographystyle{plainnat}
\bibliography{references}

\end{document}